\PassOptionsToPackage{svgnames}{xcolor}
\documentclass{article}

\usepackage{arxiv}
\usepackage{natbib}

\usepackage[utf8]{inputenc} \usepackage[T1]{fontenc}    \usepackage{hyperref}       \usepackage{url}            \usepackage{booktabs}       \usepackage{multirow}       \usepackage{amsfonts}       \usepackage{amssymb}
\usepackage{amsmath}        \usepackage{nicefrac}       \usepackage{microtype}      \usepackage{graphicx}       \usepackage{placeins}      \usepackage{wrapfig}        \usepackage[svgnames]{xcolor}
\usepackage{framed}
\usepackage[shortlabels]{enumitem} 
\definecolor{shadecolor}{rgb}{0.9765, 0.9882, 0.9961}

\newcommand{\Ttrain}{T_{\mathrm{train}}}
\newcommand{\Tgrad}{T_{\mathrm{grad}}}

\newcommand{\estimateci}[3]{  \shortstack[r]{$#1$\\{\scriptsize $[#2,\,#3]$}}}
\newcommand{\resulttabnote}[1]{  \par\smallskip\begin{minipage}{\linewidth}\footnotesize
  \raggedright #1\end{minipage}}

\title{Steering Recurrent Reasoners at Inference Time with Readout Feedback}

\author{
  \textbf{Shunsuke Kamiya}${}^{*1}$ \quad \textbf{Masanori Koyama}${}^{1}$ \quad \textbf{Seongcheol Jeong}${}^{1}$ \quad \textbf{Fumiya Uchiyama}${}^{1}$ \\[2pt]
  \textbf{Kenji Kubo}${}^{1}$ \quad \textbf{Kohei Hayashi}${}^{1}$ \quad \textbf{Masahiro Suzuki}${}^{1}$ \quad \textbf{Yutaka Matsuo}${}^{1}$ \\[2pt]
  {\normalfont ${}^{1}$Graduate School of Engineering, The University of Tokyo} \\[2pt]
  {\normalfont ${}^{*}$\texttt{shunsuke.kamiya@weblab.t.u-tokyo.ac.jp}} \\
}

\date{}
\renewcommand{\headeright}{Preprint}
\renewcommand{\undertitle}{Preprint}
\hypersetup{pdfauthor={Shunsuke Kamiya, Masanori Koyama, Seongcheol Jeong, Fumiya Uchiyama, Kenji Kubo, Kohei Hayashi, Masahiro Suzuki, Yutaka Matsuo}}
\begin{document}

\maketitle

\begin{abstract}
Recurrent models, which repeatedly update latent states with shared computation blocks, have emerged as powerful architectures for solving complex reasoning tasks. Existing inference-time methods scale computation by running more steps or sampling more trajectories, but ignore information revealed within each trajectory. Here we show that recurrent models can be improved at inference time by using their own readout probabilities to steer latent dynamics without retraining. We introduce Readout Feedback (RoFB), a test-time intervention that converts intermediate predictions into token-wise pairwise coupling forces injected into the latent dynamics. Across three recurrent models (AKOrN, ItrSA++, TRM) on Sudoku and Maze, RoFB yields clear gains in four of six model-task pairs and a small gain in one. The four positive pairs achieve performance unattainable by merely running more steps or selecting from multiple trajectories, at comparable or lower computational cost. These results suggest that closed-loop steering of latent dynamics can serve as a complementary inference-time control mechanism for recurrent reasoning models.
\end{abstract}

\section{Introduction}
\label{sec:intro}
Human experts often solve difficult problems by thinking longer: mathematicians refine intermediate structures, and chess players search and revise candidate lines before committing to a move. 
Modern reasoning systems increasingly exploit an analogous principle \textemdash additional computation at inference time can improve performance. 
In large language models, this idea appears in chain-of-thought prompting \citep{Wei2022-rs}, self-consistency \citep{Wang2022-aj}, and search-based reasoning \citep{Yao2023-iz}, where inference-time computation is used to generate, evaluate, or aggregate candidate reasoning paths. 
These developments have motivated inference-time computation as an additional scaling axis beyond model size and training data.

A complementary line of work studies recurrent reasoning models, which solve problems by repeatedly updating latent states with shared computation blocks. 
Unlike feedforward models with a fixed computational depth, recurrent reasoners expose inference-time computation explicitly through the number of recurrent updates. 
This principle has shown effective in extrapolating learned algorithms by iterating longer \citep{Bansal2022-qw}, as well as in improving reasoning performance in language models \citep{Yang2023-xk, Saunshi2025-qb, Geiping2025-nj}. 
In parallel, compact recurrent reasoners designed for structured reasoning, such as HRM \citep{Wang2025-lx}, TRM \citep{Jolicoeur-Martineau2025-gi}, AKOrN \citep{Miyato2024-zl}, and related iterative transformer and self-attention models \citep{Kubo2026-vz} have shown strong performance on puzzle-style reasoning benchmarks including Sudoku, Maze, and ARC-AGI 1 \& 2 \citep{Chollet2019-ah, Chollet2026-wb}.

Despite this progress, current inference-time improvements for recurrent reasoners mostly exploit their iterative structure in passive ways: running the dynamics for more steps \citep{Wang2025-lx, Geiping2025-nj}, or sampling multiple trajectories and selecting one according to a criterion, such as one with the lowest potential defined on the dynamics \citep{Miyato2024-zl}, or the most confident one \citep{Kubo2026-vz}. 
Although effective, these strategies come with inherent limitations: longer trajectories may saturate or remain trapped in unsuccessful dynamical regimes, while multi-trajectory voting increases compute roughly linearly with the number of candidates. 
This raises a natural question: can we improve a frozen recurrent reasoner by actively steering each latent trajectory during inference, rather than merely running longer or sampling more trajectories?

\begin{figure}[t!]
\begin{center}
\includegraphics[width=0.97\textwidth]{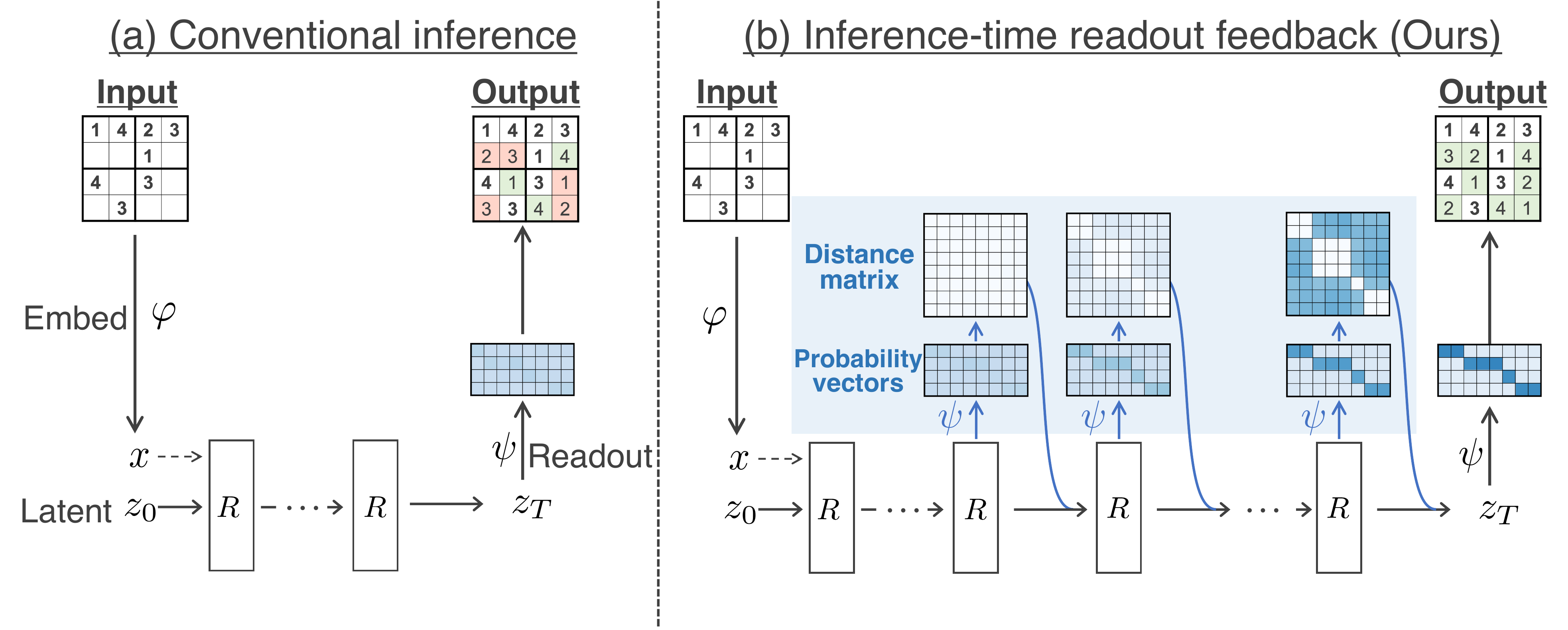}
\caption{\textbf{Conventional inference vs. our method.} (a) Conventional inference without feedback for a recurrent model. (b) Inference with RoFB. RoFB injects a feedback coupling into the latent dynamics based on the distance between readout probabilities, leaving all model weights unchanged.}
\label{fig:comparison_conventional_rofb}
\end{center}
\end{figure}

In this work, we propose a method to steer the latent trajectories using the feedback signal derived from itself. Specifically, we use the current readout probability vectors. We call this approach \textbf{Readout Feedback (RoFB)}, a closed-loop inference-time intervention for recurrent reasoning models (Fig.~\ref{fig:comparison_conventional_rofb}). 

RoFB is motivated by an observation that tokens develop a class-dependent cluster structure in latent space when a trajectory hits a correct solution. That is, inter-class distances become larger and the latent representations become more clustered when the model reaches a correct solution.

RoFB is grounded in this observation and the consequent hypothesis that encouraging such inter-class separation during inference can improve reasoning performance. RoFB, at each inference step, injects a coupling term into the latent dynamics that encourages token inter-class separation based on the distance between tokens' readout probabilities, while leaving all model weights unchanged.

We evaluate RoFB on Sudoku and Maze using three complementary recurrent reasoners with distinct architectural designs: AKOrN \citep{Miyato2024-zl}, an oscillator-inspired model; ItrSA++ \citep{Kubo2026-vz}, an iterative self-attention-based model; and TRM \citep{Jolicoeur-Martineau2025-gi}, a compact hierarchical model. 
Out of the total six model-task pairs, RoFB showed clear performance gain in four and and small but consistent gain in one, improving performance without retraining. The former four instances are those where its gains were not reachable with the conventional strategies of running longer or sampling more with $\times 10$\textendash$100$ efficient compute. 
These results suggest that frozen recurrent reasoners contain underutilized inference-time capability that can be unlocked by closed-loop control methods such as RoFB.
Our contributions are:
\begin{itemize}
\item \textbf{Readout-guided inference control.} We introduce RoFB, a test-time intervention that converts disagreement between intermediate readout probabilities into repulsive interactions between latent tokens. \item \textbf{Improved accuracy–compute trade-offs.} Across three recurrent models on Sudoku and Maze, RoFB provides clear gains in four of six model–task pairs, reaching accuracies beyond the evaluated baseline grid at lower inference cost in these four pairs.
\item \textbf{Mechanistic evidence and limits.} Controlled interventions support a contribution of separation-related feedback directions to the gains. Coupling ablations suggest distinct roles for repulsion in reaching solutions and readout-dependent weighting in retaining them.\end{itemize}

\section{Preliminaries}
\label{sec:preliminaries}

\subsection{Recurrent Models for Reasoning}
We first give the general setups for the recurrent models considered in this paper (Fig.~\ref{fig:comparison_conventional_rofb} (a)). In a nutshell, these models consist of three main components: (I) an \textbf{input embedding map} $\varphi$, (II) a \textbf{recurrent module} $R$, and (III) a \textbf{readout head} $\psi$.

More specifically, the input embedding $\varphi$ maps the raw input tokens $\mathbf{x}_{\text{raw}} = \{ \mathbf{x}^{(i)}_{\text{raw}} \}_{i=1}^N \in \{1, \ldots, V\}^N$ to a fixed embedding $\mathbf{x} = \{ \mathbf{x}^{(i)} = \varphi( \mathbf{x}^{(i)}_{\text{raw}}) \}_{i=1}^N \in \mathbb{R}^{N \times D}$, where $V$ is the vocabulary size, $N$ is the number of tokens, and $D$ is the embedding dimension. The recurrent module $R$ iteratively applies an update to the latent state $ \mathbf{z}_t = \{\mathbf{z}^{(i)}_t\}_{i=1}^N \in (\mathbb{R}^D)^N$ over $T$ steps:
\begin{equation}
  \mathbf{z}^{(i)}_{t+1} = R^{(i)}(\mathbf{z}_t; \mathbf{x}), \quad t = 0, 1, \ldots, T{-}1.
  \label{eq:iterative_update_one_token}
\end{equation}
This stepwise update can be a discretization of a continuous-time update: $\dot{\mathbf{z}}^{(i)}_t = f^{(i)}(\mathbf{z}_t; \mathbf{x}), ~~ t \in [0, T].$ Each latent token $\mathbf{z}^{(i)}_t$ is typically confined to a compact manifold $\mathcal{M} \subset \mathbb{R}^D$ (e.g., the $D$-dimensional hypersphere). The initial value $\mathbf{z}_0 \in \mathcal{M}$ is either randomly sampled from a certain distribution, or learned as a fixed parameter. After $T$ steps, the readout head $\psi: \mathcal{M} \to \mathbb{R}^{C}$, usually a multi-layer perceptron (MLP) of a few layers, produces a per-token probability vector $\mathbf{p}^{(i)} = \operatorname{softmax}(\psi(\mathbf{z}^{(i)}_T))$, where $C$ is the number of output classes.

As concrete models, we selected three representative recurrent models for reasoning tasks in the present study: AKOrN \citep{Miyato2024-zl}, ItrSA++ \citep{Kubo2026-vz}, and TRM \citep{Jolicoeur-Martineau2025-gi}. We made this choice based on their strong performance in reasoning tasks and their diverse architectural designs including hierarchical structures and latent state geometry. The key architectural differences are summarized in Table~\ref{tab:architectural_differences}. In what follows, we describe the specific architectures of the three recurrent models in detail.

\begin{table}[b]
\centering
\caption{\textbf{Architectural differences of the models used in this work.} The table summarizes the key architectural features of the three recurrent models (AKOrN, ItrSA++, and TRM) analyzed in this paper, including hierarchy, latent geometry, initialization, and input injection.
}
\label{tab:architectural_differences}
\begin{tabular}{ccccc}\toprule
Model & Hierarchy & Latent Space & Initial Value & Input Injection \\
\midrule
AKOrN & single-level & $(\mathbb{S}^{m-1})^{D/m}$ &  Random, unlearned & Every step \\
ItrSA++ & two-level & $\mathbb{S}^{D-1}(\sqrt{D})$ & Random, unlearned & Every $L$ steps \\
TRM & two-level & $\mathbb{S}^{D-1}(\sqrt{D})$ & Fixed, learned & Every step \\
\bottomrule
\end{tabular}
\end{table}

\subsection{AKOrN}
\label{sec:akorn}
Artificial Kuramoto Oscillatory Neurons (AKOrN) \citep{Miyato2024-zl} is a recurrent model with strong performance in reasoning tasks, such as Sudoku, Maze, and object recognition.
In AKOrN, each latent token $\mathbf{z}_t^{(i)}$ resides on the manifold $(\mathbb{S}^{m-1})^{D/m}$, where $m$ is the oscillator dimension, and follows a Kuramoto oscillator \citep{Kuramoto1984-im}-like dynamics;
\begin{equation}
\label{eq:original_akorn}
\Delta \mathbf{z}_t^{(i)} =\operatorname{Proj}_{\mathbf{z}_t^{(i)}}\Big(\mathbf{x}^{(i)}+\sum_j \mathbf{M}^{(ij)}_{t} (\mathbf{z}_t) \mathbf{z}_t^{(j)}\Big), \quad 
\mathbf{z}_{t+1}^{(i)} =\Pi\left[\mathbf{z}_t^{(i)}+\gamma \Delta \mathbf{z}_t^{(i)}\right].
\end{equation}
Here, $\mathbf{M}^{(ij)}_{t}$ is a learnable coupling matrix such as the self-attention matrix~\citep{Vaswani2017-ys}. Projection and normalization act independently on each $m$-dimensional oscillator group: for group vectors $\mathbf{z},\mathbf{f},\mathbf{v}\in\mathbb{R}^m$, $\operatorname{Proj}_{\mathbf{z}}(\mathbf{f}) = \mathbf{f} - \langle \mathbf{f}, \mathbf{z} \rangle \mathbf{z}$ and $\Pi(\mathbf{v}) = \mathbf{v}/\|\mathbf{v}\|$. Here, $\mathbf{z}_t = \{ \mathbf{z}^{(i)}_t \}_{i=1}^N$ is the collection of all latent states, and $\gamma$ is a learnable step size.
The coupling matrix $\mathbf{M}_{t}^{(ij)}$ defines the interaction between tokens $i$ and $j$ at time $t$, and is implemented as the self-attention output of the current latent state. 
We deliberately drop the natural-frequency term $\boldsymbol{\Omega}_i$ present in the original formulation, for simplicity. Appendix~\ref{app:akorn_natural_frequencies} reports comparisons with natural frequencies enabled and results on a reproduction of the original Sudoku architecture.

In this work, due to the difficulty of training the original formulation (Eq.~\eqref{eq:original_akorn}) on \emph{Sudoku Extreme} task (Appendix \ref{app:akorn_natural_frequencies}), we explored some design choices and implemented AKOrN in a slightly modified form using a normalization-free transformer:
\begin{equation}
\label{eq:modified_akorn}
\Delta \mathbf{z}^{(i)}_t =\operatorname{Proj}_{\mathbf{z}^{(i)}_t}\left( \mathrm{mlp} \left( \bigl[\mathbf{z}_t + \mathbf{x} + \mathrm{SelfAttn}(\mathbf{z}_t + \mathbf{x} )\bigr]^{(i)} \right) \right),
\end{equation}
where $[\cdot]^{(i)}$ extracts the $i$-th token from the sequence, $\mathrm{mlp}$ is a two-layer feed-forward network with a GELU activation, and $\mathrm{SelfAttn}$ is a multi-head self-attention block with a positional encoding. The results with original AKOrN and the modified version with/without natural frequencies are reported in Appendix \ref{app:akorn_natural_frequencies}.

\subsection{ItrSA++}
\label{sec:itrsa}
ItrSA++ is another recurrent model proposed by \citet{Kubo2026-vz}, which shows even stronger reasoning capabilities in Sudoku and Maze. Unlike AKOrN, ItrSA++ is designed to have a two-level hierarchy, where the low-level process $\mathbf{z}$ is designed to capture fast-changing finer details, while the high-level process $\mathbf{y}$ is expected to capture slower-changing global structure.
In the low-level process $\mathbf{z}$, the dynamics is updated as follows:
\begin{equation}
\mathbf{z}_{t+1}^{(i)}
= \begin{cases}
\operatorname{RN}\!\left( \mathbf{z}_t^{(i)}
+\bigl[\operatorname{SelfAttn}(\mathbf{z}_t)\bigr]^{(i)} \right) & t\not\equiv0\pmod L,\\
\operatorname{RN}\!\left(
  \operatorname{RN}(\mathbf{y}_t^{(i)}) +  \bigl[\operatorname{CrossAttn}\!\left( \operatorname{RN}(\mathbf{y}_t),\operatorname{RN}(\mathbf{x}) \right)\bigr]^{(i)}
\right)
& t\equiv0\pmod L.
\end{cases}
\end{equation}
where $\operatorname{RN}$ denotes the RMS normalization and $\operatorname{CrossAttn}$ is a cross-attention block. The high-level process $\mathbf{y}$ is updated every $L$ steps:
\begin{equation}
\mathbf{y}_{t+1}^{(i)} =
\begin{cases}
\mathbf{z}_{t+1}^{(i)}
+\operatorname{SwiGLU}(\mathbf{z}_{t+1}^{(i)})
& t+1\equiv0\pmod L,\\
\mathbf{y}_t^{(i)}
& \text{otherwise},
\end{cases}
\end{equation}
with $\mathbf{y}_0 = \mathbf{z}_0$. 
The latent state space of ItrSA++ can be interpreted as $\mathbb{S}^{D-1}(\sqrt{D})$ thanks to the RMS normalization.

\subsection{Tiny Recursive Model (TRM)}
\label{sec:trm}
TRM \citep{Jolicoeur-Martineau2025-gi} is another recurrent model with strong performance in reasoning benchmark tasks, such as Sudoku, Maze, and ARC-AGI \citep{Chollet2019-ah, Chollet2026-wb}. TRM simplifies the Hierarchical Reasoning Model (HRM)~\citep{Wang2025-lx} while improving its accuracy with a smaller model. 
TRM has a two-level hidden-state hierarchy, which is designed to capture both global structure and local details in the reasoning process.
The low-level process $\mathbf{z}$, designed to capture local details, is updated as follows: 
\begin{equation}
  \mathbf{z}_{t+1}^{(i)} = \left[ \operatorname{TF}_n \!\bigl(\mathbf{z}_t + \mathbf{y}_t + \mathbf{x}\bigr) \right]^{(i)}. \end{equation}
Here, $\operatorname{TF}_n$ is an $n$-layer transformer block. 
The high-level process $\mathbf{y}$, on the other hand, is updated only every $L$ steps to capture the global structure of the problem:
\begin{equation}
  \mathbf{y}_{t+1}^{(i)} = \begin{cases} 
   \left[\operatorname{TF}_n \!\bigl(\mathbf{z}_t + \mathbf{y}_t\bigr)\right]^{(i)} & \text{if $t \equiv 0 \pmod{L}$,} \\
   \mathbf{y}_t^{(i)} & \text{otherwise.}
  \end{cases}
\end{equation}
Each $\mathbf{z}^{(i)}_t$ lies on $\mathbb{S}^{D-1}(\sqrt{D})$ thanks to an RMS normalization \citep{Zhang2019-nz} at the end of each update. The readout is applied to the high-level state $\mathbf{y}_t$ after a certain number of recursions, producing the final output of the model.

\section{Readout Feedback (RoFB)}
\label{sec:readout_feedback}

\subsection{Geometry Matters: Inter-token Clusterization}
\label{sec:token_alignment}

\begin{figure}[ht!]
\begin{center}
\includegraphics[width=0.95\textwidth]{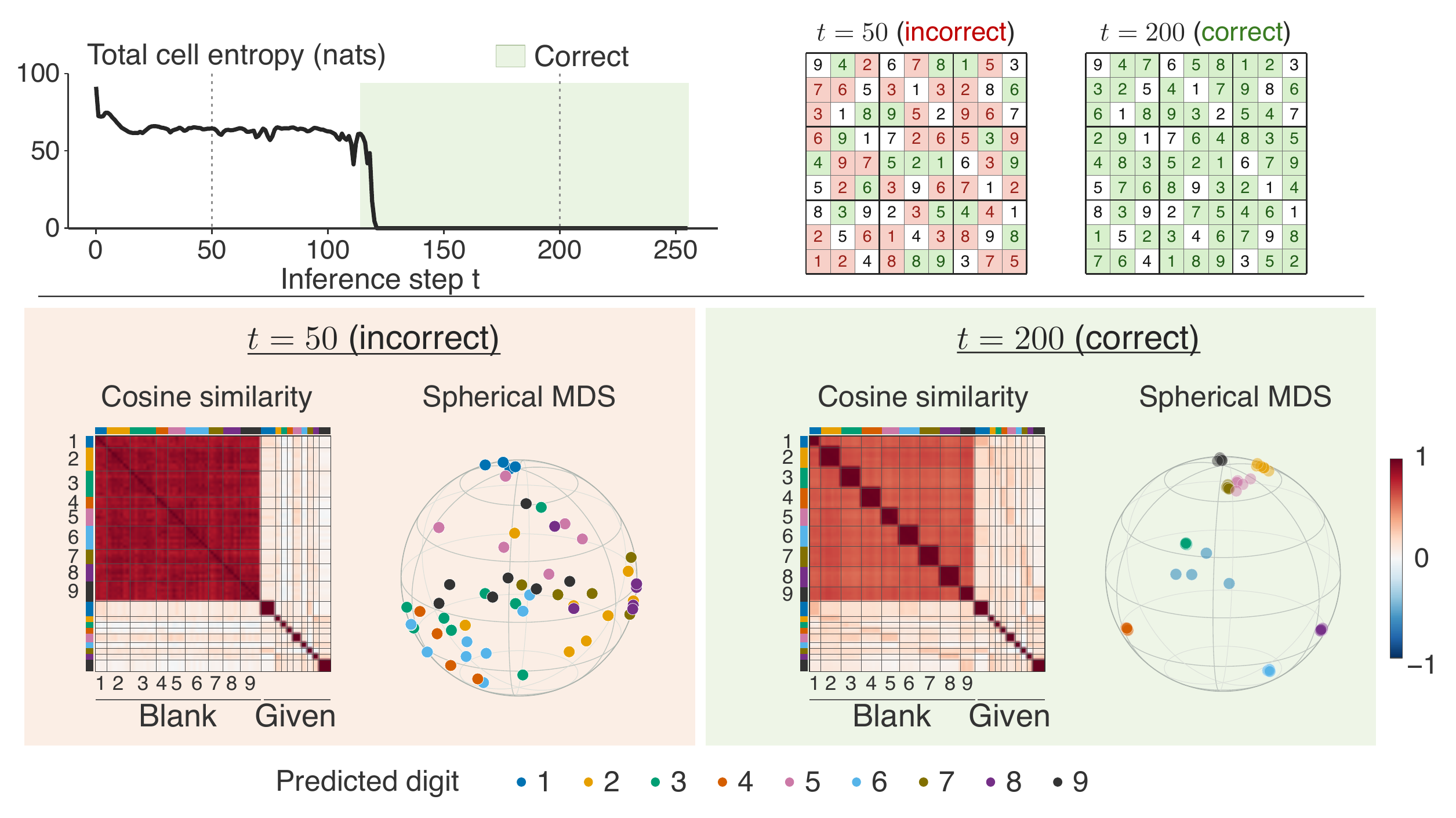}
\caption{
 \textbf{AKOrN inference dynamics on a Sudoku puzzle.} 
(Top left) Total cell entropy, $\sum_i H(\mathbf{p}^{(i)})$, over 256 inference steps. The shaded region indicates steps at which the predicted board is correct.
(Top right) Predicted boards at $t=50$ and $t=200$. \textcolor[RGB]{32,81,25}{Green} and \textcolor[RGB]{142,26,26}{red} indicate \textcolor[RGB]{32,81,25}{correct} and \textcolor[RGB]{142,26,26}{incorrect} predictions on blank cells, respectively.
(Bottom) Token-wise cosine similarity matrices and spherical multidimensional scaling (MDS) visualizations at $t=50$ (left pair) and $t=200$ (right pair).
The matrices include all tokens, grouped by predicted digit, with blank-cell tokens preceding given-cell tokens. The MDS plots show blank-cell tokens only, colored by predicted digit.
}
\label{fig:board_prediction_entropy_cosine_similarity}
\end{center}
\end{figure}

Before describing our method RoFB, we first inspect how tokens in the latent space behave during inference. Fig.~\ref{fig:board_prediction_entropy_cosine_similarity} shows a sample AKOrN trajectory that solves a Sudoku puzzle; Fig.~\ref{fig:maze_inference_dynamics} shows a Maze example.

In both puzzles, the total entropy decreases as the inference proceeds, indicating that the model becomes more confident in its predictions. 
The entropy transition in Sudoku appears abrupt (Fig.~\ref{fig:board_prediction_entropy_cosine_similarity}, Top left), while it is more gradual in Maze (Fig.~\ref{fig:maze_inference_dynamics}, Top left). Once the entropy decreases, the solution stabilizes. In the models we examined, samples that ultimately failed to reach the correct solution tended to have high entropy that did not fully decrease (Appendix \ref{app:clustering_in_correct_solutions}).

We observe that when the model reaches the solution of a problem, the latent tokens clusterize according to the classes that they represent. That is, tokens of the same class (e.g., the same number in Sudoku, solution path in Maze) are closely aligned in the latent space, while tokens of different classes are separated. 
We can see that cluster formation, or inter-class separation, becomes more pronounced once the model reaches the correct answer (\emph{Cosine similarity} and \emph{Spherical MDS} in Fig.~\ref{fig:board_prediction_entropy_cosine_similarity}). 
In Sudoku, almost every blank (i.e., non-input) token pair has a large similarity value $\approx 1.0$ while the puzzle is still unsolved ($t=50$), indicating that the tokens are aligned regardless of the classes to which they belong. 
But once the puzzle is solved ($t=200$), the tokens form class-wise clusters; tokens of the same class are well aligned, having large cosine similarity  $\sim 1.0$, while those representing different classes have lower cosine similarity (Fig.~\ref{fig:board_prediction_entropy_cosine_similarity}, \emph{Cosine similarity}, $t=200$). 
In Maze, although less evident, the inter-class separation becomes more apparent after reaching the correct solution (Fig.~\ref{fig:maze_inference_dynamics}, \emph{Cosine similarity}, $t=200$).
For the models we examined, this latent clusterization was also a good indicator of correctness: see Appendix \ref{app:clustering_in_correct_solutions}.

From these observations, we hypothesized that adding an external force to the latent tokens to promote clustering, or inter-class separation, can improve the model's inference performance.
 We test later that this hypothesis in Section \ref{sec:mechanistic_analysis}, and we propose RoFB, which is described in the following section. 

\subsection{RoFB}
\label{sec:rofb}

To encourage the latent token dispersion, RoFB inserts a feedback coupling term that depends on the distance between readout probability vectors. We designed RoFB so that the dynamics retains the original forward information along with the separation among tokens representing different classes while the model's predictions are not certain. We implement this by a simple addition using readout probability vectors:
\begin{equation}
\label{eq:rofb_discrete_time}
\mathbf{z}^{(i)}_{t+1} = \operatorname{\Pi}_{\mathcal{M}} \left(
R^{(i)}(\mathbf{z}_t, \mathbf{x}) + 
\colorbox[HTML]{F0F8FF}{$\displaystyle
\lambda \cdot g(\mathbf{z}_t) \cdot \text{Proj}_{\mathbf{z}^{(i)}_{t}}\left( \sum_{j \ne i} \tilde{J}^{(ij)}_t \Bigl(\mathbf{p}^{(i)}_t, \mathbf{p}^{(j)}_t \Bigr) \, \mathbf{z}^{(j)}_t \right)
$}
\right).
\end{equation}
Here, $\tilde{J}^{(ij)}_t$ is the normalized coupling term that reflects the distance between the readout probability vectors $\mathbf{p}^{(i)}_t$ and $\mathbf{p}^{(j)}_t$, $g$ a gate function, and $\lambda$ a scaling hyperparameter. The $\operatorname{Proj}$ operator projects the feedback term to the tangent space of the manifold $\mathcal{M}$ at $\mathbf{z}_t^{(i)}$, which ensures that the dynamics stays on the manifold, and $\operatorname{\Pi}_{\mathcal{M}}$ is a projection to the manifold $\mathcal{M}$. We provide more detailed explanation on the coupling term $\tilde{J}^{(ij)}_t$ and the gate function $g$ in the following.

\paragraph{Coupling $\tilde{J}$.} The coupling term $\tilde{J}^{(ij)}_t$ is a scalar defined as
\begin{equation}
\label{eq:coupling_discrete_time}
\tilde{J}^{(ij)}_t = J^{(ij)}_t / \nu^{(i)}, \quad
J^{(ij)}_t \Bigl( \mathbf{p}^{(i)}_t, \mathbf{p}^{(j)}_t \Bigr) = h \left(d \Bigl( \mathbf{p}^{(i)}_t,  \mathbf{p}^{(j)}_t \Bigr) \right), 
\end{equation}
where $h$ is a non-increasing function, $\nu^{(i)}$ is a certain normalization term (e.g., $\nu^{(i)} = \sum_{k \ne i} |J^{(ik)}_t|$: absolute sum), $d$ a bounded distance function such as $d( x, y ) = 1- x^{\!\top} y \in [0,\, 1]$,
$\mathbf{p}^{(i)}_t$ the readout probability vector of the $i$-th token, $\mathbf{p}^{(i)}_t = \operatorname{softmax}\!\left(\psi \Bigl(\mathbf{z}^{(i)}_t \Bigr)\right) \in \Delta^{C-1} := \{\mathbf{p} \in [0,1]^C \mid  p_1 + \cdots + p_C = 1 \}$,
We specifically use $h(d)=-d$, the simplest linear coupling that assigns zero weight to confident same-class predictions and increasingly negative weight as predicted class disagreement grows. We evaluate this design choice against attractive and mixed couplings in our ablations (see \emph{Connectivity} section in Appendix \ref{subsec:ablation_rofb}).

For tasks with pre-given clue tokens (Sudoku, Maze), RoFB is applied only to blank-cell tokens: the update in Eq.~\eqref{eq:rofb_discrete_time} is restricted to $i \in \mathcal{B}$ and the inner sum to $\sum_{j \in \mathcal{B},\, j \ne i}$, where $\mathcal{B} \subseteq \{1, \dots, N\}$ is the set of blank-cell indices and $N_{\text{blank}}=|\mathcal{B}|$. We retain the unrestricted form in Eq.~\eqref{eq:rofb_discrete_time} for notational simplicity.

\paragraph{Gate $g$.} The gate function $g(\mathbf{z}_t)$ is designed to turn on the feedback term only when the model is not confident enough. We can, for instance, define the gate function as follows:
\begin{equation}
\label{eq:rofb_gate}
g(\mathbf{z}_t) := \underbrace{
1[t \geqslant t_{\text{min}}]
}_{\text{Warmup}}
~ \underbrace{ \sigma \Big( \frac{\max_{i} H^{(i)}_t - \alpha H_{\text{unif}}}{\tau} \Big) }_{\text{Confidence check}},
\end{equation}
where $1[\cdot]$ is the indicator function, $\sigma$ is the sigmoid function with temperature $\tau$, $H^{(i)}_t$ is the entropy of the readout probability vector of token $i$ at time $t$, $H_{\text{unif}}$ is the entropy of the uniform distribution, and $\alpha$ is a preset threshold. RoFB thus has four hyperparameters ($\lambda$, $t_{\min}$, $\alpha$, $\tau$), which we select by automated search \citep{Akiba2019-qj} on a validation set (Section~\ref{sec:experiments}).

\section{Related Work}
\label{sec:related_work_v2}

\paragraph{Recurrent Reasoning Models}
Recurrent models that repeatedly apply shared computation blocks, dating back to the Universal Transformer~\citep{Dehghani2018-ug}, have shown strong potential on algorithmic tasks~\citep{Yang2023-xk, Bansal2022-qw}, mathematical reasoning~\citep{Geiping2025-nj, Bae2024-kn, Zhu2025-da, Wang2026-ux}, and puzzles including Sudoku, Maze, and ARC-AGI~\citep{Miyato2024-zl, Wang2025-lx, Jolicoeur-Martineau2025-gi, Gao2025-ja, Kubo2026-vz}, complementing (latent) chain-of-thought in language models~\citep{Wei2022-rs, Hao2024-qf}.
As an inference intervention method, \cite{Sghaier2026-le} proposed an inference-time mechanism that adds on stochasticity to a frozen model and utilizes it to improve predictions. RoFB, on the other hand, leverages the intermediate readout probabilities to steer inference, providing a deterministic and closed-loop approach for inference intervention.

\paragraph{Test-time Intervention}
Inference-time methods for language models mostly generate, compare, or aggregate candidate outputs through prompting, sampling, or compute allocation~\citep{Wei2022-rs,Zhou2022-ni,Wang2022-aj,Yao2023-iz,Zhai2026-xb}. Methods that intervene in the computation itself use readouts to reduce predictive entropy~\citep{Agarwal2025-uk} or to feed probability-weighted embeddings back~\citep{Zhang2025-xc}, steer hidden states with online feedback from a learned verifier or from entropy and logits~\citep{Nguyen2026-ru,Meng2026-sh}, or modify attention through selected heads, query/value spaces, or tokens~\citep{Zhang2023-bs,Zhang2025-uy,Torop2025-ef,Zhang2025-af}. RoFB combines readout feedback with inter-token interaction: it converts readouts into a pairwise coupling within the latent dynamics of recurrent reasoners, which has rarely been explored.

\paragraph{Dynamics and geometry of recurrent inference}
The internal mechanisms of recurrent models are gradually being analyzed \citep{Lu2025-kz, Blayney2026-pz, Ren2026-ql}. Theoretical analyses of the internal dynamics of recurrent Transformers include \citet{Geshkovski2023-ap, Geshkovski2024-dt}, which show the cluster-forming dynamics of recurrent self attention dynamics. While few studies have utilized the internal mechanism in favor of improving inference, RoFB is designed to leverage these characteristics to steer inference, which has been an open question. 

\paragraph{Relational Structure in Output Space.}
Dense CRFs and CRF-RNN refine neural predictions through pairwise label compatibility~\citep{Krahenbuhl2012-cv,Zheng2015-hr}. In source-free domain adaptation, NRC~\citep{Yang2021-lw} and AaD~\citep{Yang2022-rg} exploit neighborhood consistency and attraction--dispersion objectives over predicted distributions to update model parameters. RoFB shares the principle that output distributions contain useful relational structure, but uses the readout probabilities of a recurrent reasoning model at inference time, which makes it novel.

\section{Experiments}
\label{sec:experiments}

We evaluate RoFB on two benchmarks, Sudoku and Maze, with the three models described earlier.

\textbf{Sudoku} is a number-placement puzzle, whose objective is to fill a $9 \times 9$ grid with digits so that each column, row, and box ($3 \times 3$ subgrid) contains all of the digits from 1 to 9. Following the prior works~\citep{Wang2025-lx, Kubo2026-vz}, we used \textit{Sudoku Extreme} dataset, and applied the same preprocessing and augmentation techniques. We tested with 20,000 puzzle boards randomly sampled from all test boards.

\textbf{Maze} is a pathfinding problem where the goal is to find a shortest path from a starting point to a target point in a $30 \times 30$ grid environment with obstacles. Following the prior works~\citep{Wang2025-lx, Kubo2026-vz}, we used \textit{Maze Hard} dataset, and applied the same preprocessing techniques. We tested with all 1,000 puzzle boards.

\paragraph{Training.} 
For AKOrN and ItrSA++, we trained the models on each task using truncated backpropagation through time \citep{Aicher2019-xa} following \citet{Kubo2026-vz}. For TRM, we trained the model using truncated backpropagation through time with deep supervision, following \citet{Jolicoeur-Martineau2025-gi}; see Appendix~\ref{app:setup}.

\paragraph{Evaluation.} For AKOrN and ItrSA++, we evaluated the performance of a model based on the readout after $T_{\text{eval}} \in \{64, 128,256\}$ recurrence steps. We also evaluated model performances using confidence-based voting \citep{Kubo2026-vz} with $K_{\text{vote}} \in \{1,4,16,64\}$, where trajectories are sampled with random initial values and the one minimizing the sum of entropy over all cells is selected as the final output. 

For TRM, we evaluated $N_{\text{block}}\in\{8,12,16\}$ outer iterations, using 16 as the standard horizon following \citet{Jolicoeur-Martineau2025-gi}. We did not apply voting for TRM, as TRM learns a fixed initial value of the latent dynamics, rather than sampling it from a distribution, and thus the voting procedure is not applicable. 

For Sudoku, we report board accuracy, where a prediction is correct only if all cells match the ground-truth board. For Maze, we report shortest-path accuracy, where a prediction is correct only if the predicted path is the shortest path (not necessarily the given ground-truth path).

\paragraph{Hyperparameter Tuning.}
Hyperparameters are tuned on a validation set; see Appendix~\ref{subsec:hp_search}.

\paragraph{Additional Computational Cost of RoFB.}
Since RoFB increases the computational cost per inference step, we evaluate performance based on FLOPs at each $K_{\text{vote}}$ and $T_{\text{eval}}$ value for a fair comparison. The increase in compute per trajectory and per step due to RoFB was kept small, within +20\% (see Appendix Table~\ref{tab:rofb_compute}). 
We also provide results based on wall time, including hyperparameter search, in the Appendix~\ref{subsec:hp_search},~\ref{sec:compute}. Since hyperparameter search is performed only once, the computational efficiency of RoFB improves as $K_{\text{vote}}$ and $T_{\text{eval}}$ increase.

\subsection{Performance}
\begin{figure}[t!]
\begin{center}
\includegraphics[width=0.95\textwidth]{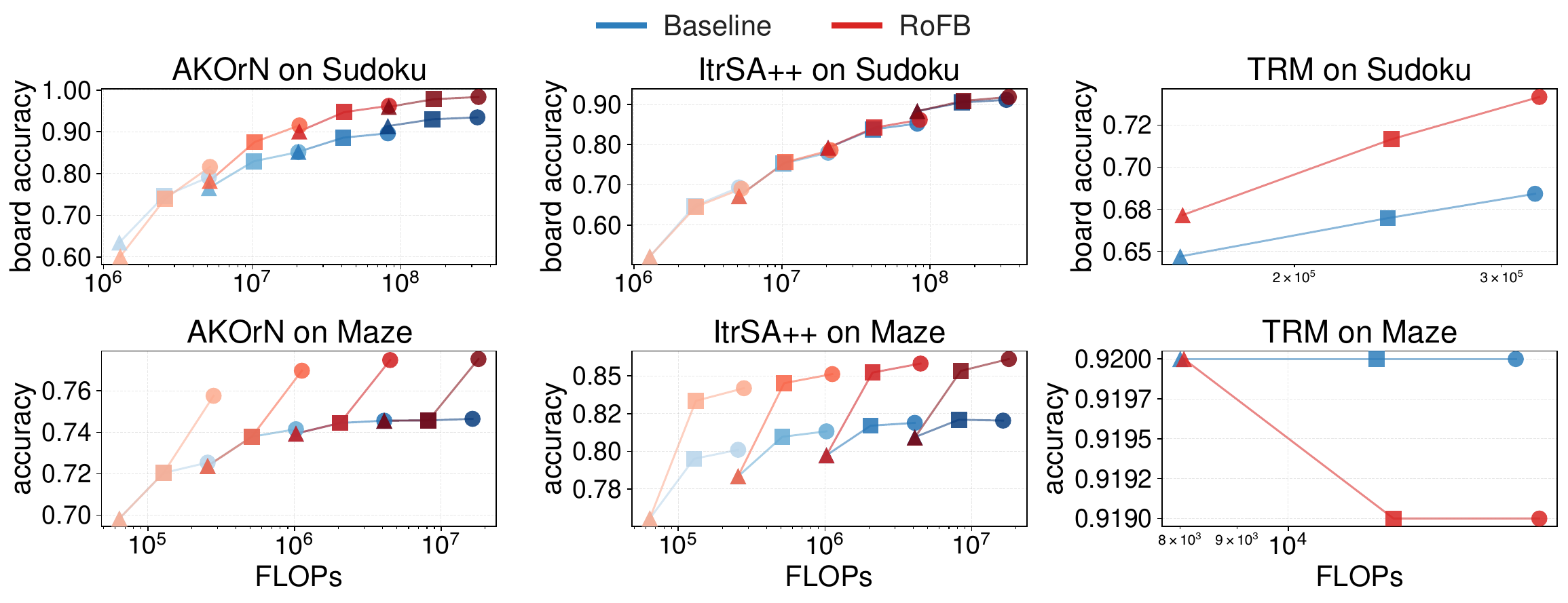}
\caption{
\textbf{Compute-accuracy comparison of baseline vs. RoFB on model-task pairs.} 
    \emph{Color} encodes $K_{\text{vote}} \in \{1, 4, 16, 64\}$ (\textcolor[RGB]{39,117,178}{Baseline \textsf{blue}} / \textcolor[RGB]{210,26,35}{RoFB \textsf{red}}, shade darkens with $K_{\text{vote}}$ from \textcolor[HTML]{B7D4EA}{\textsf{1}}/\textcolor[HTML]{FCAB8F}{\textsf{1}} to \textcolor[HTML]{083C7D}{\textsf{64}}/\textcolor[HTML]{7E0610}{\textsf{64}}).
    \emph{Shape} encodes the inference time steps $T$ ($\blacktriangle, \blacksquare$, {\LARGE $\bullet$} for $T = 64, 128, 256$; $N_{\text{block}} \in \{8, 12, 16\}$ for TRM). Each AKOrN and ItrSA++ point is the mean over 8 inference seeds;  TRM on Maze uses a compressed y-axis due to baseline saturation. The remaining $K_{\text{vote}}$ values and all numerical values are in Appendix~\ref{app:full_grid}.
}
\label{fig:model_task_compute_performance}
\end{center}
\end{figure}

Fig.~\ref{fig:model_task_compute_performance} compares baseline and RoFB across all model-task pairs (TRM has no voting and hence a single shade). Except for TRM on Maze, where both saturate, accuracy generally increases with $K_{\text{vote}}$ and $T_{\text{eval}}$ for both.

\paragraph{Performance Gain of RoFB.}
We observe a clear performance boost with RoFB in four of the six model-task pairs (AKOrN on both tasks, ItrSA++ on Maze, and TRM on Sudoku) at the same or lower computational cost, while the gain was small in ItrSA++ on Sudoku ($+0.6$\%) and absent in TRM on Maze (Fig.~\ref{fig:model_task_compute_performance}). 
Crucially, in the four positive model-task pairs, RoFB reaches operating points that are unattainable by either prolonging $T_{\text{eval}}$ or increasing $K_{\text{vote}}$ in the baseline. 
For instance, on Maze, ItrSA++ with RoFB achieves $83.4\%$
accuracy at
$(K_{\text{vote}}, T_{\text{eval}})=(1,128)$,
exceeding the best baseline accuracy of $82.1\%$ across the
evaluated grid, attained at $(64,128)$, while using approximately
$1/62$ of its inference FLOPs.
For TRM on Sudoku, where confidence-based voting is unavailable, RoFB lifts board accuracy from $68.4\%$ to $74.2\%$ ($+5.8$\%) at $N_{\text{block}}{=}16$, demonstrating that the gain does not depend on the availability of a voting mechanism.

\subsection{Mechanistic Analysis}
\label{sec:mechanistic_analysis}
We next ask how RoFB produces these gains. We find that (i) the gains require the separation-related directions of the feedback, (ii) the feedback plays two roles---repulsion that helps trajectories reach a solution and readout-dependent weighting that helps retain it---and (iii) among the other separation interventions we tested, some also help but none outperforms RoFB. We then characterize when RoFB does not help. Details and statistics are in Appendices~\ref{app:mechanistic_analysis_inter_class_separation}--\ref{app:rofb_confidence}.

\paragraph{Separation-related directions are required.}
As intended, RoFB increases inter-class separation in all six model-task pairs, including instances that remain incorrect under both conditions (Tables~\ref{tab:inter_class_separation} and~\ref{tab:inter_class_separation_both_incorrect}). To test whether this change matters for accuracy, we remove from the RoFB force its projection onto the direction that changes inter-class similarity to first order and compare with RoFB rescaled to the same magnitude. Removal lowers accuracy by 2.3--3.3\% on Sudoku for all three models and by 8.9 and 11.0\% for AKOrN and ItrSA++ on Maze, with no resolved change on TRM on Maze (Table~\ref{tab:separation_component_removal}). 
\paragraph{Two roles: reaching and retaining a solution.}
The coupling $1-\mathbf{p}^{(i)\top}\mathbf{p}^{(j)}$ has two limits: when readouts are uncertain it is nearly constant, so RoFB reduces to uniform repulsion among blank tokens; when readouts are confident it vanishes within a predicted class and remains repulsive across classes. Replacing it by constant repulsion ($h\equiv-1$, same gate and strength) separates these roles (Appendix~\ref{subsec:ablation_rofb}). On Sudoku, constant repulsion matches RoFB (AKOrN 91.6 vs.\ 91.5\%, TRM 74.7 vs.\ 74.2\%, ItrSA++ 77.0 vs.\ 77.0\%; Table~\ref{tab:ablation_coupling}), whereas norm-matched random directions fall 4--11\% short of RoFB in three pairs. Where unsolved states show nearly parallel blank tokens (Fig.~\ref{fig:board_prediction_entropy_cosine_similarity}, $t=50$), repulsion itself thus suffices to reach solutions. On Maze, constant repulsion falls short (AKOrN 73.8 vs.\ 78.8\%, ItrSA++ 84.3 vs.\ 85.3\%).
In longer runs, constant and readout-dependent coupling reach a correct solution in a similar number of trajectories, but among those that do, 11.4\% vs.\ 3.1\% (AKOrN) and 2.0\% vs.\ 0.2\% (ItrSA++) later end incorrectly.
Readout dependence therefore matters for \emph{retaining} a solution, consistent with the vanishing within-class repulsion once predictions are confident. 

\paragraph{Comparison with other separation interventions.}
Inter-class separation could also be promoted in other ways. To examine whether such alternatives are effective as well, we replaced only the direction of the RoFB force with forces derived from three readout-based clustering objectives (margin repulsion, a probability-weighted contrastive loss, and within-class variance plus a centroid margin) and from a readout-free K-means attraction, each with a validation-selected strength (Appendix~\ref{app:mechanistic_analysis_inter_class_separation}). Some of these alternatives also improved accuracy over the no-feedback baseline: margin repulsion, which, like RoFB, repels tokens with dissimilar predictions, did so in four model-task pairs (by 2.2--7.6\%), and the other two readout-based objectives did so on AKOrN on Maze. None of them, however, was significantly more accurate than RoFB in any of the 24 comparisons, whereas 12 were significantly less accurate (Fig.~\ref{fig:alternative_separation_interventions}). In the settings we examined, RoFB therefore performed best among these separation interventions. Separately, a fixed-scale entropy-gradient control changed accuracy by at most 0.7\% in the four pairs where RoFB gained 3.2--7.2\% (Appendix~\ref{app:rofb_confidence}), suggesting that RoFB does more than sharpen predictions.

\paragraph{When does RoFB not help?}
On TRM on Maze, RoFB increased inter-class separation without improving accuracy. 
Two possible reasons for the this non-effectiveness are given. Frist, incorrect predictions were already confident and strongly separated, weakening feedback on confident errors through the entropy gate: 71 of 72 incorrect outputs have maximum normalized entropy at most 0.10, and all 72 have a silhouette score of at least 0.5 (Appendix~\ref{app:clustering_in_correct_solutions}). Second, because the coupling acts on current predicted classes, it strengthens the relations implied by these predictions but does not by itself distinguish wrong assignments from correct ones. 
We also examined TRM on ARC-AGI-1~\citep{Chollet2019-ah}, including harder logical puzzles. The validation-selected RoFB configuration showed no consistent pass@1 improvement on the remaining 300 evaluation puzzles. In a separate diagnostic sample balanced between baseline successes and failures, 45 of 97 confident terminal predictions were incorrect (Appendix~\ref{app:arc}). Thus, confidence and prediction-based separation do not guarantee correctness, although their role in the lack of improvement remains a hypothesis.

\section{Limitations and Discussion}
\label{sec:limitation_discussion}

We do not yet have a validated diagnostic that predicts whether RoFB will improve accuracy on a new model or task. Confidence and class separation can both be high for incorrect predictions, as observed in TRM on Maze, so neither alone establishes when feedback will be beneficial.

Our main evaluation covers two puzzle benchmarks (\textit{Sudoku Extreme}, \textit{Maze Hard}) and three small recurrent architectures, with an additional TRM experiment on ARC-AGI-1. We do not evaluate language models, to which direct transfer is non-trivial. If RoFB is applied directly to LLMs, the large vocabulary size may dilute the inter-token coupling on which RoFB relies. Whether RoFB or such a variant provides additional gains to LLMs is left to future work.

\section{Conclusions}
\label{sec:conclusion}

We proposed RoFB to enhance the inference performance of recurrent reasoning models. RoFB utilizes the readout probabilities during inference to provide feedback to the latent states. 
Across three architectures on Sudoku and Maze, RoFB yielded clear gains in four of six model--task pairs and a small gain in one, reaching accuracies unattainable by longer rollouts or confidence-based voting at lower inference compute. 
Our analysis indicates two roles of the feedback, repulsion that helps trajectories leave unsolved states and readout-dependent weighting that retains solutions once reached.
Our results provide evidence that readout-based closed-loop steering can complement existing inference-time strategies such as longer rollouts and multi-trajectory voting. While the present evidence is limited to puzzle-style benchmarks and compact recurrent reasoners, it suggests a promising direction for using intermediate predictions to control frozen recurrent dynamics at test time.

\subsection*{AI use statement}
We used generative AI tools, including Codex and Claude Code, to assist with experimental design and refinement, implementation and debugging of evaluation and analysis code, interpretation of results, preparation of figures and tables, and drafting and revision of the manuscript. Checks of AI-assisted code and numerical outputs included execution tests and comparisons with saved experimental artifacts. The authors take responsibility for the final content of this work, including its text, claims, code, and results.

\subsection*{Ethics statement}
This work studies inference-time interventions on Sudoku and maze-solving benchmarks and does not involve human participants or personal data. The experiments concern small-scale recurrent reasoning models; they do not establish the reliability or safety of the method in real-world applications. Any such application would require a separate assessment of failure modes and potential harms in its intended setting.

\subsection*{Reproducibility statement}
Section~\ref{sec:rofb} defines RoFB, and Appendix~\ref{sec:architecture_specific_details} describes its model-specific implementation. Section~\ref{sec:experiments} and Appendix~\ref{app:setup} describe the datasets and evaluation protocol; Appendix~\ref{app:hparams} lists training hyperparameters. Appendix~\ref{subsec:hp_search} specifies the validation-based hyperparameter search and selected settings, while Appendix~\ref{sec:compute} details computational-cost accounting. Appendix~\ref{app:full_grid} reports the full evaluation grid and inference-seed replication, and Appendices~\ref{app:mechanistic_analysis_inter_class_separation}--\ref{app:rofb_confidence} describe the mechanistic interventions and controls.

\section*{Acknowledgments}
Authors SK, MK, SJ, FU, KK, and KH are supported by computational resources of the TSUBAME4.0 supercomputer provided by Institute of Science Tokyo through the HPCI System Research Project (Project ID: hp260232).

\bibliographystyle{plainnat}
\bibliography{references}

\appendix

\appendix
\label{sec:appendix}

\section{Sample of Correct Inference Dynamics of AKOrN on Maze}
\label{sec:maze_overview}
\begin{figure}[ht!]
\begin{center}
\includegraphics[width=0.95\textwidth]{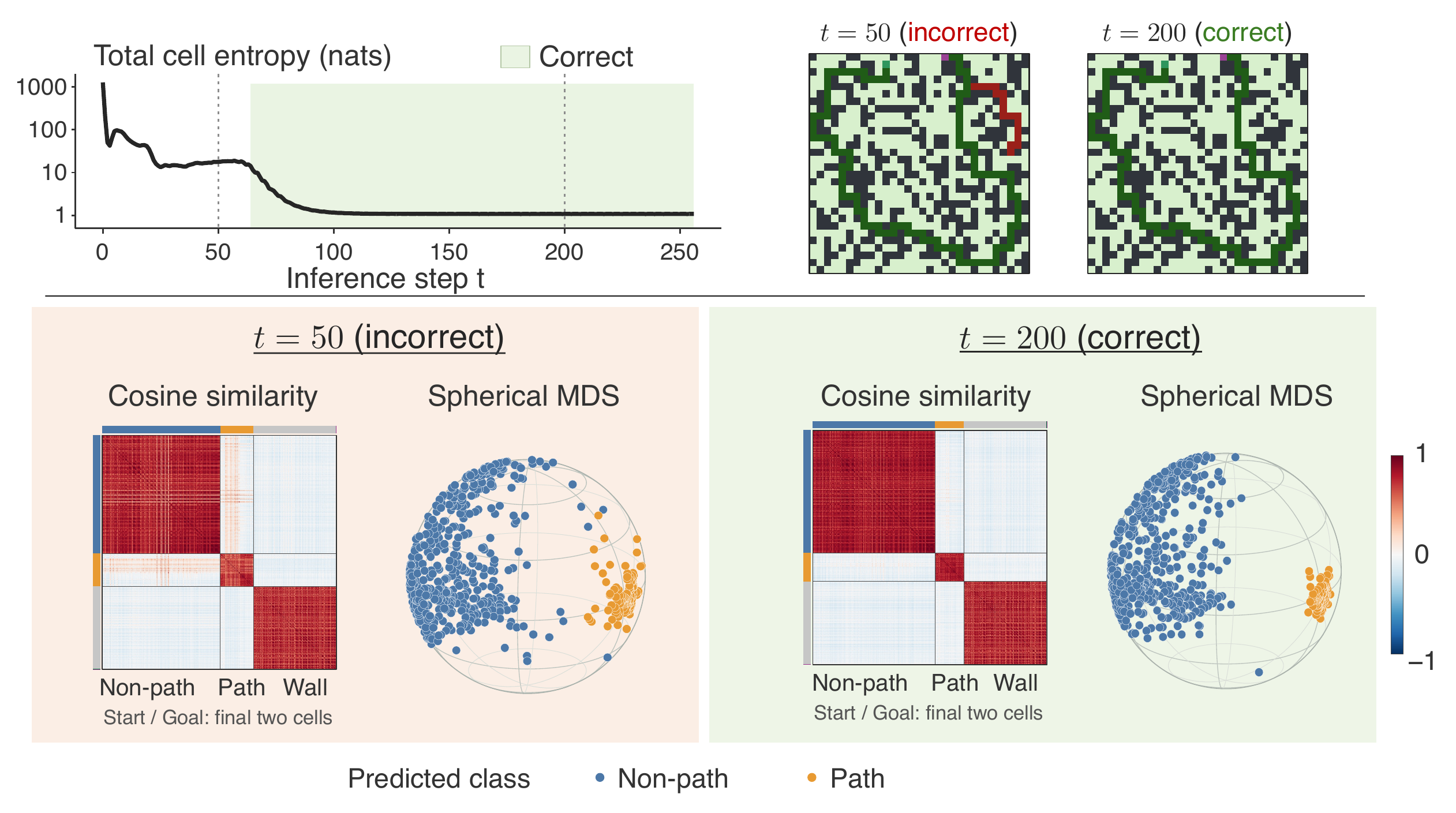}
\caption{
\textbf{AKOrN inference dynamics on a Maze puzzle.}
(Top left) Total cell entropy, $\sum_i H(\mathbf{p}^{(i)})$, over 256 inference steps, shown on a logarithmic vertical axis. The shaded region indicates steps at which the predicted board is correct.
(Top right) Predicted boards at $t=50$ and $t=200$. \textcolor[RGB]{32,81,25}{Green} and \textcolor[RGB]{142,26,26}{red} indicate \textcolor[RGB]{32,81,25}{correct} and \textcolor[RGB]{142,26,26}{incorrect} predictions on passage cells relative to the reference path, respectively. Darker shades indicate cells predicted to lie on the path.
(Bottom) Token-wise cosine similarity matrices and spherical multidimensional scaling (MDS) visualizations at $t=50$ (left pair) and $t=200$ (right pair).
The matrices include all tokens, grouped by predicted class in the order non-path, path, wall, start, and goal. The MDS plots show passage-cell tokens only, excluding the start and goal, colored by predicted class.
}
\label{fig:maze_inference_dynamics}
\end{center}
\end{figure}

Figure~\ref{fig:maze_inference_dynamics} illustrates the inference dynamics of a correct Maze puzzle solved by AKOrN, highlighting the evolution of cell entropy, readout boards, cosine similarity matrices, and the three-dimensional visualization of latent tokens throughout the inference process.

\section{Confidence, Correctness, and Clusterization}
\label{app:clustering_in_correct_solutions}
In this section, we show the relationships between confidence, correctness, and clustering behavior of latent tokens in the three models and the Sudoku and Maze tasks. In a nutshell, we show

\begin{itemize}
\item Correct solutions tend to exhibit strong class-aligned clustering of latent tokens and high confidence.
\item In the Sudoku task, strong clusterization and high confidence are strong indicators of correctness.
\item In the Maze task, strong clusterization and high confidence do not necessarily imply correctness. Both discriminate correct from incorrect outputs more effectively in AKOrN and ItrSA++ than in TRM.
\end{itemize}

We first describe the evaluation setting and define the clustering and confidence metrics. We then examine their relationships with correctness and with each other, before discussing the distinctive pattern observed in TRM on Maze.

\paragraph{Evaluation setting.} We analyze terminal readout states from ordinary inference without RoFB on 2,000 Sudoku and 900 Maze test boards per model, with one frozen checkpoint and one candidate per board. The board identities, checkpoints, initialization procedures, and inference paths match the no-feedback reference of the component-intervention analysis. AKOrN runs for 256 recurrent updates, ItrSA++ for 64 outer blocks containing 256 self-attention updates, and TRM for 16 outer iterations. Geometry is measured in AKOrN's terminal readout state, ItrSA++'s post-MLP readout state, and TRM's high-level readout state $\mathbf z_H$. Correctness requires agreement with the reference solution in Sudoku and a valid shortest path in Maze, including alternative shortest paths.

\paragraph{Clustering metrics.} We use cosine distance and terminal predicted class labels, restricting the analysis to mutable cells: blank cells in Sudoku and passage cells excluding the start and goal in Maze. Let $\mathcal B$ denote this set of cells.

\paragraph{$\Delta_{\cos}$.} $\Delta_{\cos}$ is the difference between the average inter-class and intra-class distances, quantifying the separation among classes. Let $\hat{\mathbf z}^{(i)}=\mathbf z^{(i)}/\|\mathbf z^{(i)}\|$, and let $\mathcal S$ and $\mathcal D$ denote pairs of distinct mutable cells with the same and different predicted labels, respectively. We define
\begin{equation}
C_{\mathrm{same}}=\frac{1}{|\mathcal S|}\sum_{(i,j)\in\mathcal S}\hat{\mathbf z}^{(i)\top}\hat{\mathbf z}^{(j)},\qquad
C_{\mathrm{diff}}=\frac{1}{|\mathcal D|}\sum_{(i,j)\in\mathcal D}\hat{\mathbf z}^{(i)\top}\hat{\mathbf z}^{(j)},
\end{equation}
and
\begin{equation}
\Delta_{\cos}=(1-C_{\mathrm{diff}})-(1-C_{\mathrm{same}})
=C_{\mathrm{same}}-C_{\mathrm{diff}}.
\end{equation}

\paragraph{Silhouette Score.} The silhouette score is the mean of $(b_i-a_i)/\max(a_i,b_i)$ over mutable cells, where $a_i$ is the mean cosine distance to other cells of the same class and $b_i$ is the smallest mean distance to another class. More explicitly, let $c_i$ denote the terminal predicted class of cell $i$. Writing $\mathcal B_c=\{j\in\mathcal B:c_j=c\}$ for the cells of predicted class $c$ and $\delta_{ij}=1-\hat{\mathbf z}^{(i)\top}\hat{\mathbf z}^{(j)}$ for cosine distance,
\begin{equation}
a_i=\frac{1}{|\mathcal B_{c_i}|-1}
\sum_{j\in\mathcal B_{c_i}\setminus\{i\}}\delta_{ij},\qquad
b_i=\min_{\substack{c\ne c_i\\|\mathcal B_c|>0}}
\frac{1}{|\mathcal B_c|}\sum_{j\in\mathcal B_c}\delta_{ij}.
\end{equation}
Thus $b_i$ compares cell $i$ with its nearest other class in mean distance, rather than with all other classes pooled together. If $\mathcal B_{\mathrm{sil}}$ contains the cells whose own class has at least two members and for which another nonempty class exists, the reported score is
\begin{equation}
s=\frac{1}{|\mathcal B|}
\sum_{i\in\mathcal B_{\mathrm{sil}}}
\frac{b_i-a_i}{\max(a_i,b_i)}.
\end{equation}

Singleton-class cells contribute zero and remain included in the averaging denominator $|\mathcal B|$. If $a_i=b_i=0$, the cell's contribution is set to zero; a one-class partition is undefined. Larger values of either index indicate stronger class separation.

\paragraph{Confidence metrics.} Predictive uncertainty is measured by the maximum normalized cell entropy,
\begin{equation}
h_{\max}=\max_{i\in\mathcal B}
\left[-\frac{1}{\log C}\sum_{c=1}^{C}p^{(i)}_c\log p^{(i)}_c\right],
\end{equation}
where $p^{(i)}_c$ is the terminal readout probability of output class $c$ at cell $i$, computed over all $C$ outputs of the model's readout head without restriction or renormalization to task classes. For AKOrN, ItrSA++, and TRM, respectively, $C$ is 9, 9, and 11 for Sudoku and 6, 5, and 6 for Maze. The maximum is sensitive to uncertainty at even a single mutable cell; larger $h_{\max}$ indicates more uncertainty. The metric is computed per trajectory before taking the correct- and incorrect-group medians. Entropy is computed from the saved logits in double precision without clipping probabilities to a positive floor.

We quantify the ability of each metric to distinguish correct from incorrect outputs using the area under the receiver operating characteristic curve (AUROC). Correct outputs are the positive class; larger clustering scores and lower entropy indicate correctness, so the confidence score is $-h_{\max}$. AUROC is the fraction of correct--incorrect pairs ranked in this direction, counting ties as one half; 0.500 indicates chance-level ranking and an exact AUROC of 1.000 indicates perfect ranking in the evaluated sample. The reported AUROCs are descriptive point estimates for these boards and frozen checkpoints, rounded to three decimal places; a rounded value of 1.000 need not indicate perfect ranking.

\subsection{Clusterization and Correctness}

\begin{figure}[!htbp]
\centering
\includegraphics[width=\linewidth]{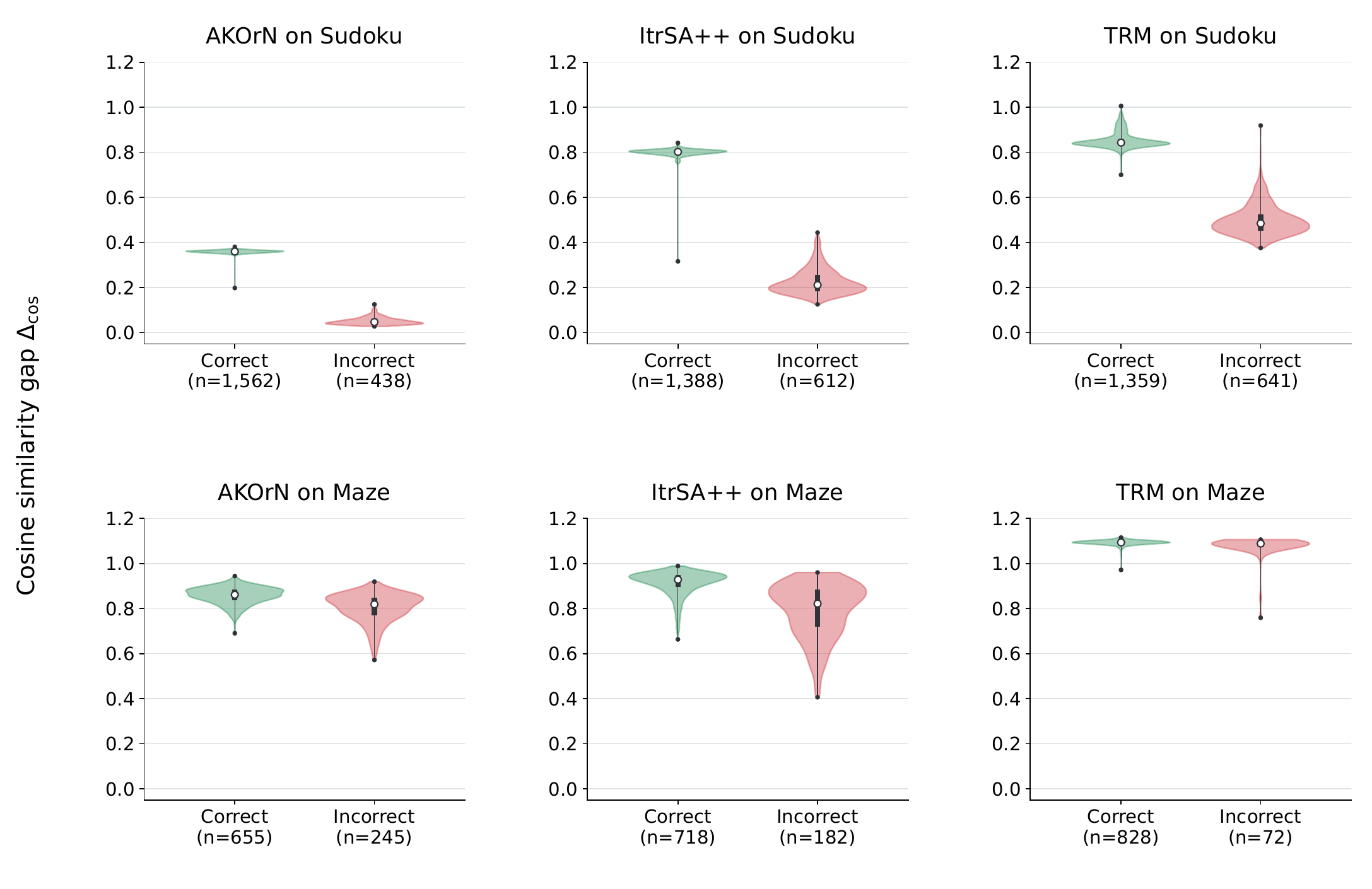}
\caption{\textbf{Cosine gap by solution correctness.} Cells are grouped by terminal predicted labels. White dots show medians, thick bars the interquartile range, and thin bars the observed range; violin widths are normalized separately.}
\label{fig:confidence_correctness_cosine_gap}
\end{figure}

\begin{figure}[!htbp]
\centering
\includegraphics[width=\linewidth]{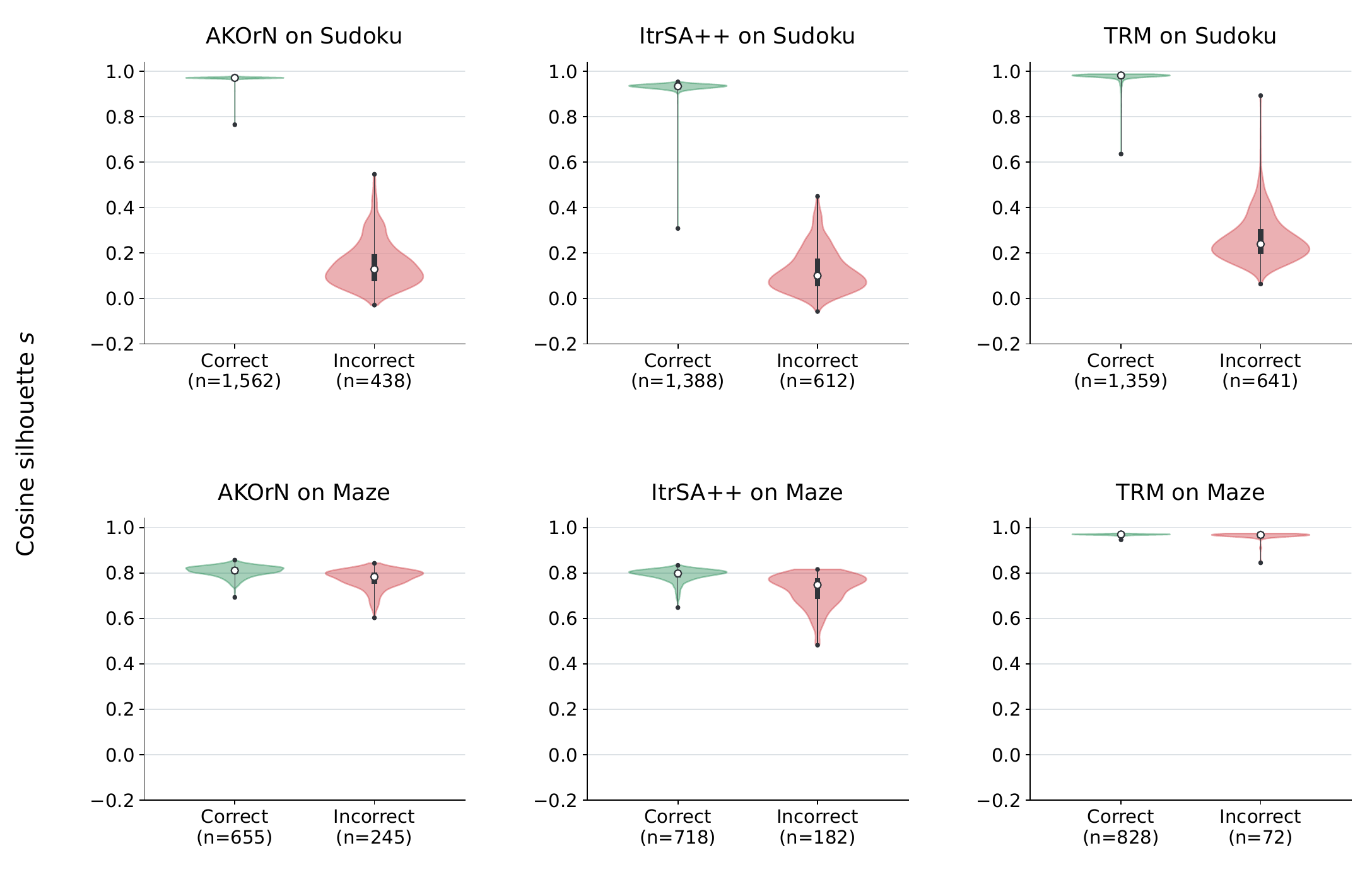}
\caption{\textbf{Silhouette score by solution correctness.} Grouping and plot conventions follow Figure~\ref{fig:confidence_correctness_cosine_gap}. Higher scores indicate stronger separation of latent tokens according to their predicted classes.}
\label{fig:confidence_correctness_silhouette}
\end{figure}

Across the six model--task pairs, correct solutions tend to exhibit strong clustering, but the converse does not necessarily hold, particularly in Maze. Figures~\ref{fig:confidence_correctness_cosine_gap} and~\ref{fig:confidence_correctness_silhouette} show the distributions of the two clustering metrics for correct and incorrect outputs. In Sudoku, the distributions are clearly separated: the cosine-gap AUROCs are 1.000, 1.000, and 0.997 for AKOrN, ItrSA++, and TRM, respectively, while the silhouette AUROCs are 1.000, 1.000, and 1.000. Thus, clustering is a strong indicator of correctness in the evaluated Sudoku outputs, although the small distributional overlap in ItrSA++ and TRM precludes a strict implication.

In Maze, the distributions overlap substantially more. The cosine-gap AUROCs are 0.774, 0.846, and 0.675, and the silhouette AUROCs are 0.779, 0.839, and 0.687, in the same model order. AKOrN and ItrSA++ retain greater discriminative ability than TRM, whose correct and incorrect outputs both exhibit strong clustering. Consequently, clustering according to predicted classes alone does not reliably distinguish correct from incorrect Maze solutions, especially for TRM.

\subsection{Confidence and Correctness}
\begin{figure}[!tbp]
\centering
\includegraphics[width=\linewidth]{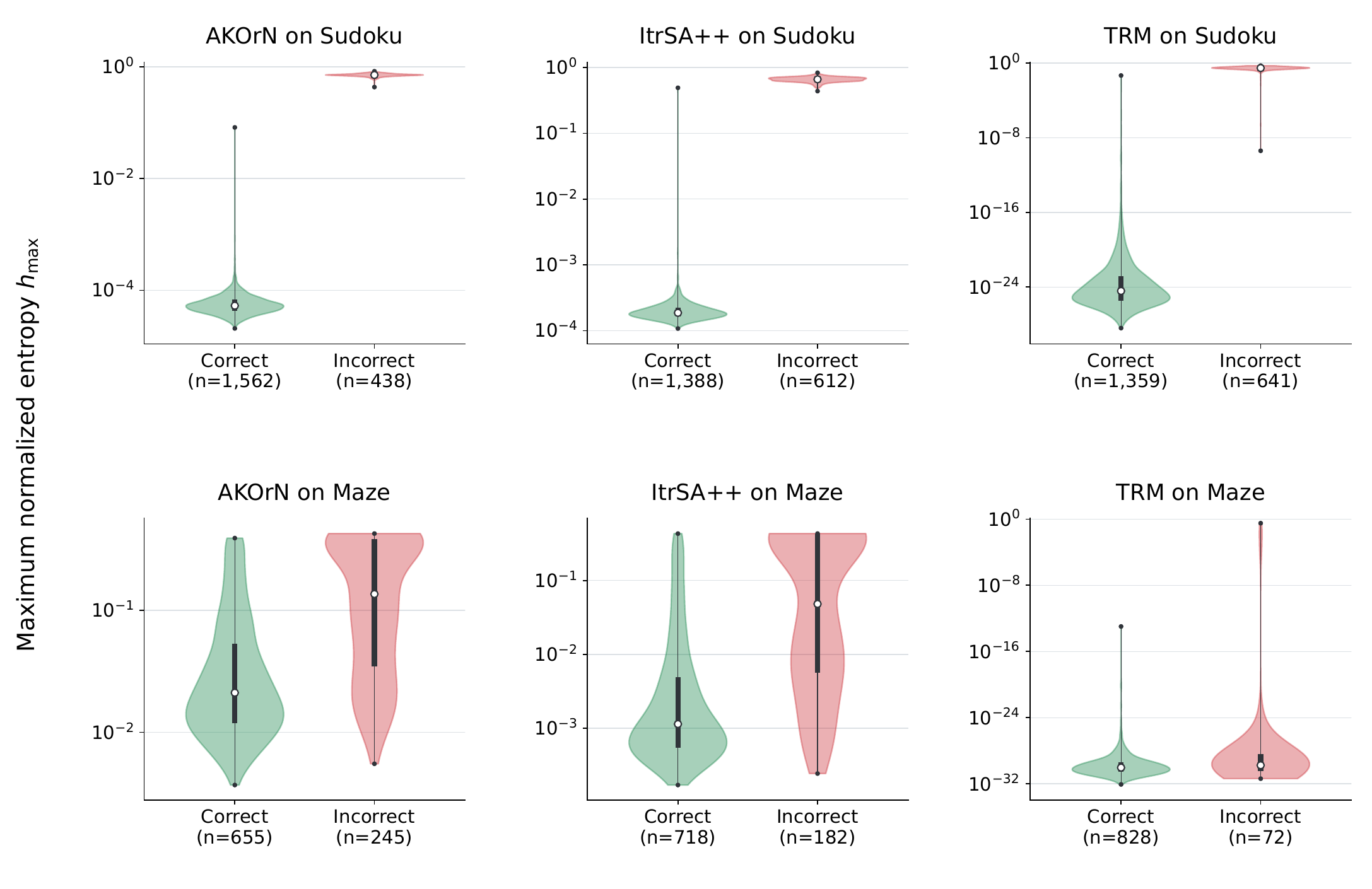}
\caption{\textbf{Maximum normalized cell entropy by solution correctness.} Lower values indicate higher confidence. Entropy is displayed on logarithmic axes with panel-specific ranges; densities are estimated in log space. Other plot conventions follow Figure~\ref{fig:confidence_correctness_cosine_gap}.}
\label{fig:confidence_correctness_entropy}
\end{figure}

The relationship between confidence and correctness follows a similar pattern. Correct solutions tend to have high confidence, but high confidence does not necessarily imply correctness. In Sudoku, Figure~\ref{fig:confidence_correctness_entropy} shows a clear separation between the correct- and incorrect-output distributions. The AUROC of $-h_{\max}$ rounds to 1.000 for all three models. Confidence therefore almost perfectly ranks correct above incorrect outputs in this sample, while still allowing exceptions.

In Maze, the correct- and incorrect-output distributions overlap more, especially for TRM. The corresponding AUROCs are 0.788, 0.822, and 0.607 for AKOrN, ItrSA++, and TRM, respectively. These results show that the models can be confident in incorrect Maze solutions; for TRM, confidence offers only limited discrimination between correct and incorrect outputs.

\subsection{Clusterization and Confidence}

\begin{figure}[!htbp]
\centering
\includegraphics[width=\linewidth]{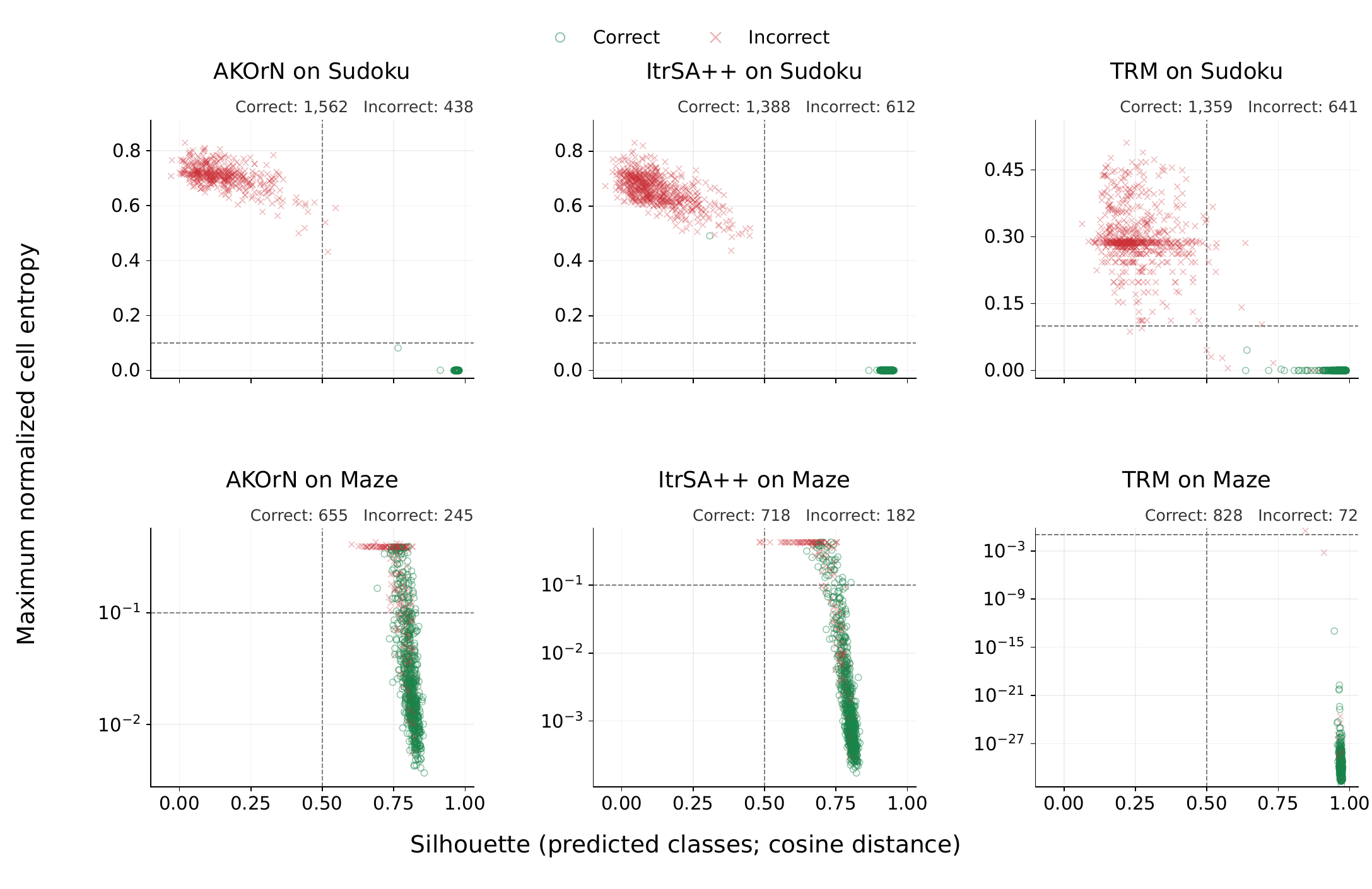}
\caption{\textbf{Clustering and confidence in terminal readout states.} Each point is one board; green circles denote correct outputs and red crosses incorrect outputs. Entropy axes are linear for Sudoku and logarithmic for Maze, with panel-specific ranges. Dashed lines mark $s=0.500$ and $h_{\max}=0.100$ as visual reference levels, not optimized decision boundaries.}
\label{fig:confidence_clustering_scatter}
\end{figure}

\paragraph{Sudoku.} Strong clustering and high confidence generally occur together. In Figure~\ref{fig:confidence_clustering_scatter}, correct outputs concentrate at high silhouette scores and low maximum entropy, whereas most incorrect outputs have lower silhouette scores and higher entropy. This joint pattern is consistent with the strong discrimination observed for each metric separately. Nevertheless, the overlap, including confidently wrong outputs in TRM, means that neither the individual metrics nor their visual combination establishes a guarantee of correctness.

\paragraph{Maze.} Strong clustering also occurs in incorrect outputs, and its relationship with confidence does not separate the two correctness groups as clearly as in Sudoku. In AKOrN and ItrSA++, larger silhouette scores tend to accompany lower entropy, but correct and incorrect outputs overlap along this trend. In TRM, both groups concentrate at high silhouette scores and extremely low entropy. Thus, latent tokens can be well separated according to their predicted classes while the predicted path remains incorrect.

\subsection{TRM on Maze}

TRM on Maze illustrates the limits of both clustering and confidence as indicators of correctness. The median silhouette scores are 0.970 and 0.968 for correct and incorrect outputs, respectively, while the corresponding median maximum normalized entropies are $9.247\times10^{-31}$ and $1.767\times10^{-30}$. Of the 72 incorrect outputs, 71 have $h_{\max}\le0.10$ and all 72 have a silhouette score of at least 0.5. Incorrect predictions can therefore be both strongly clustered and highly confident. The AUROCs above indicate a difference in ranking, but not a clear separation of the two groups.

This pattern provides a possible explanation for the limited benefit of RoFB on TRM on Maze. RoFB promotes separation according to predicted classes, so increasing separation need not repair an already well-separated but incorrect assignment. Its entropy gate also reduces feedback when predictions are confident; if confident errors arise during the intervention window, uncertainty alone may fail to identify them as requiring intervention. These observations are consistent with the finding that increased inter-class separation does not yield an accuracy gain on TRM on Maze. However, the present analysis measures ordinary-inference terminal states, so it does not establish when these errors become confident or whether gate suppression causes the lack of improvement.
\FloatBarrier

\section{Architecture-Specific Details of RoFB}
\label{sec:architecture_specific_details}
In this section, we provide architecture-specific details for the implementation of RoFB. Although the general RoFB design stays the same, as provided in Eq.~\ref{eq:rofb_delta_z_app}, the specific form of the coupling and how it is applied varies depending on the architecture. We discuss each architecture separately below.

\paragraph{Coupling pipeline (recap of \S\ref{sec:rofb}).} For self-contained readability we recall the coupling formula here. Given per-cell readout probabilities $\mathbf{p}^{(i)}_t = \operatorname{softmax}(\psi(\mathbf{z}^{(i)}_t)) \in \Delta^{C-1}$ at iteration $t$, the unnormalized coupling is $J^{(ij)}_t = h(d(\mathbf{p}^{(i)}_t, \mathbf{p}^{(j)}_t)) = -\bigl(1 - (\mathbf{p}^{(i)}_t)^{\!\top}\mathbf{p}^{(j)}_t\bigr)$. We row-normalize $J$ by its absolute sum,
\begin{equation}
\tilde J^{(ij)}_t \;=\; \frac{J^{(ij)}_t}{\sum_{k \ne i}\bigl\lvert J^{(ik)}_t\bigr\rvert + \varepsilon},
\qquad \varepsilon = 10^{-6}, 
\label{eq:rofb_J_tilde_app}
\end{equation}
which guarantees $\sum_{j \ne i}\bigl\lvert\tilde J^{(ij)}_t\bigr\rvert \le 1$ and is sign-preserving when the row's $J^{(ik)}_t$ values share a sign (the case for the repulsive kernel above, where all $J \le 0$).
When a kernel produces sign-mixed rows, Eq.~\ref{eq:rofb_J_tilde_app} can be replaced by a uniform $1/N$ normalization, where $N$ is the number of feedback-receiving tokens. 
The coupling vector for token $i$ is then
\begin{equation}
\delta \mathbf{z}^{(i)}_t \;=\; \lambda \cdot g(\mathbf{z}_t) \cdot \operatorname{Proj}_{\mathbf{z}^{(i)}_t}\!\left(\sum_{j \ne i} \tilde J^{(ij)}_t\, \mathbf{m}^{(j)}_t\right),
\label{eq:rofb_delta_z_app}
\end{equation}
where $\lambda$ is the coupling strength, $g(\mathbf{z}_t)$ is the entropy-driven gate of \S\ref{sec:rofb}, $\operatorname{Proj}_{\mathbf{z}^{(i)}_t}$ is the tangent-space projection at $\mathbf{z}^{(i)}_t$, and the \emph{coupling material} $\mathbf{m}^{(j)}_t$ is the only architecture-dependent quantity, specified below.

\paragraph{AKOrN.} The latent state $\mathbf{z}^{(i)}_t \in (\mathbb{S}^{m-1})^{D/m}$ lies on a product sphere (\S\ref{sec:akorn}). The readout $\psi$ is a $1{\times}1$ convolution after a GELU activation, and the coupling material is the latent itself, $\mathbf{m}^{(j)}_t = \mathbf{z}^{(j)}_t$. We add $\delta \mathbf{z}^{(i)}_t$ to the dynamics velocity as:
\begin{equation}
\mathbf{z}^{(i)}_{t+1} = \Pi\!\left[\,\mathbf{z}^{(i)}_t + \gamma\bigl(\Delta \mathbf{z}^{(i)}_t + \colorbox[HTML]{F0F8FF}{$\displaystyle \delta \mathbf{z}^{(i)}_t $}\bigr)\right],
\label{eq:akorn_rofb}
\end{equation}
where $\Delta \mathbf{z}^{(i)}_t$ is the baseline velocity defined in \S\ref{sec:akorn}, $\gamma$ is the learnable step size, and $\Pi$ is Euclidean norm normalization within each $m$-dimensional oscillator group. Both the tangent projection and normalization are applied groupwise. The radial component of $\delta \mathbf{z}^{(i)}_t$ is automatically removed by the projection embedded in $\Delta \mathbf{z}^{(i)}_t$'s tangent definition, so RoFB acts as a tangential velocity correction on the per-group sphere.

\paragraph{ItrSA++.} We use the ItrSA++ with all RMSNorm gains removed ($\mathtt{rmsnorm\_affine}{=}\mathrm{False}$, see Table~\ref{tab:hp_itrsa}), which is almost equivalent to the original ItrSA++. This forces every RMSNorm output to land exactly on the sphere $\mathbb{S}^{D-1}(\sqrt{D})$.

The original dynamics, for the sake of clarity, are written as follows:
\begin{equation}
\mathbf{z}_{t+1}^{(i)}
= \begin{cases}
\operatorname{RN}\!\left( \mathbf{z}_t^{(i)}
+\bigl[\operatorname{SelfAttn}(\mathbf{z}_t)\bigr]^{(i)} \right) & t\not\equiv0\pmod L,\\
\operatorname{RN}\!\left(
  \operatorname{RN}(\mathbf{y}_t^{(i)}) +  \bigl[\operatorname{CrossAttn}\!\left( \operatorname{RN}(\mathbf{y}_t),\operatorname{RN}(\mathbf{x}) \right)\bigr]^{(i)}
\right)
& t\equiv0\pmod L,
\end{cases}
\end{equation}
where $\mathbf{y}_0 = \mathbf{z}_0$ and
\begin{equation}
\mathbf{y}_{t+1}^{(i)} =
\begin{cases}
\mathbf{z}_{t+1}^{(i)}
+\operatorname{SwiGLU}(\mathbf{z}_{t+1}^{(i)})
& t+1\equiv0\pmod L,\\
\mathbf{y}_t^{(i)}
& \text{otherwise}.
\end{cases}
\end{equation}

We apply RoFB at each self-attention update. The readout probabilities are computed from the pre-update state $\mathbf{z}_t$ through the usual SwiGLU residual transformation:
\begin{equation*}
\mathbf{p}_t^{(i)} = \operatorname{softmax}\!\left(\psi\!\left(\mathbf{z}_t^{(i)} + \operatorname{SwiGLU}(\mathbf{z}_t^{(i)})\right)\right).
\end{equation*}
These probabilities determine both the coupling weights and the entropy gate. The coupling material is $\mathbf{m}_t^{(j)}=\mathbf{z}_t^{(j)}$, and the tangent projection is also evaluated at the pre-update state. The update is
\begin{equation}
\begin{aligned}
\bar{\mathbf{z}}_{t+1}^{(i)} &\;=\; \operatorname{RN}\!\left(\mathbf{z}_t^{(i)} + \bigl[\operatorname{SelfAttn}(\mathbf{z}_t)\bigr]^{(i)}\right), \\
\delta \mathbf{z}^{(i)}_t &\;=\; \lambda\,g(\mathbf{z}_t)\,\operatorname{Proj}_{\mathbf{z}^{(i)}_t}\!\left(\sum_{j\ne i}\tilde J_t^{(ij)}\mathbf{z}_t^{(j)}\right), \\
\mathbf{z}^{(i)}_{t+1} &\;=\; \operatorname{RN}\!\left(\bar{\mathbf{z}}_{t+1}^{(i)} + \colorbox[HTML]{F0F8FF}{$\displaystyle \delta \mathbf{z}^{(i)}_t$}\right).
\end{aligned}
\label{eq:itrsa_rofb}
\end{equation}
Here, $\bar{\mathbf{z}}_{t+1}$ denotes the state after the ordinary self-attention update and before feedback is added. We add the feedback to this state and apply RMS normalization again. The cross-attention and block-output SwiGLU operations remain unchanged.

\paragraph{TRM.} TRM maintains two-level states, low-level $\mathbf{z}^{(i)}_t$ and high-level $\mathbf{y}^{(i)}_t$ (\S\ref{sec:trm}). We apply RoFB on TRM such that the readout is taken from the high-level state $\mathbf{y}_t$, while the feedback target is the low-level state $\mathbf{z}_t$: $\mathbf{p}^{(i)}_t = \operatorname{softmax}(\psi(\mathbf{y}^{(i)}_t))$ and the coupling material is $\mathbf{m}^{(j)}_t = \mathbf{z}^{(j)}_t$. At each low-level update,
\begin{equation}
\mathbf{z}^{(i)}_{t+1} \;=\; \operatorname{RN}\!\left( \left[ \operatorname{TF}_n\bigl(\mathbf{z}_t + \mathbf{y}_t + \mathbf{x}\bigr) \right]^{(i)} \;+\; \colorbox[HTML]{F0F8FF}{$\displaystyle \delta \mathbf{z}^{(i)}_t$}\right).
\label{eq:trm_rofb}
\end{equation}

\section{Experimental Setup}
\label{app:setup}
\paragraph{Tasks and datasets.}
We evaluate three iterative reasoners --- AKOrN~\citep{Miyato2024-zl}, ItrSA++~\citep{Kubo2026-vz}, and TRM~\citep{Jolicoeur-Martineau2025-gi} --- on two combinatorial reasoning tasks: \emph{Sudoku Extreme} (9$\times$9 grid, 1{,}000 base puzzles augmented to ${\sim}1{,}001{,}000$ training samples; 422{,}786 test puzzles) and \emph{Maze Hard} (30$\times$30 grid, 1{,}000 train / 1{,}000 test). Both datasets are shared across all three models --- the same training and test splits, with model-specific token vocabularies --- so that any across-model gap reflects architecture and training choices, not data. 

\paragraph{Training protocol.}
All three models are trained from scratch with AdamW~\citep{Loshchilov2017-yv}, truncated backpropagation through time (TBPTT) \citep{Aicher2019-xa, Kubo2026-vz}, where the backpropagation gradients are computed only on the last fixed number of steps. The parameters were renewed with an exponential moving average (EMA). We follow each model's published recipe for the model-specific hyperparameters. Full hyperparameters are in Appendix~\ref{app:hparams} (Tables~\ref{tab:hp_akorn}--\ref{tab:hp_trm}).

\section{Hyperparameter Search for RoFB}
\label{subsec:hp_search}
For each model-task pair, we perform a single hyperparameter sweep over a scaling parameter $\lambda$ and three RoFB gate parameters $\alpha, t_{\min}, \tau$ defined in Eqs.~\eqref{eq:rofb_discrete_time} and \eqref{eq:rofb_gate}.
Beyond these four parameters, we fixed the coupling function to $h(d)=-d$, representing \textit{repulsive} feedback, which is held constant across the tasks and models.

We used Optuna~\citep{Akiba2019-qj} with the TPE sampler and additionally enable the Successive Halving Pruner~\citep{Li2018-sh} (an ASHA-style pruner with reduction factor $\eta=3$ and $\min_{\text{resource}}=1$) for all model-task pairs, running $n_{\text{trial}}=30$ trials per pair.
The pruner reports intermediate validation accuracy after each of $n_{\text{chunks}}$ disjoint chunks of the validation set: $n_{\text{chunks}}=4$ for Sudoku tasks ($N_{\text{val}}=500$, i.e.\ $125$ boards per chunk) and $n_{\text{chunks}}=2$ for Maze tasks ($N_{\text{val}}=100$, i.e.\ $50$ boards per chunk).
Each trial evaluates a single $(\lambda, \alpha, t_{\min}, \tau)$ tuple by running inference with RoFB on the full validation set at a single operating point ($K_{\text{vote}}=1$ and $T_{\text{eval}}$ matching the standard evaluation horizon per model: $256$ Kuramoto steps for AKOrN, $256$ self-attention steps ($=64$ outer blocks) for ItrSA++, $16$ outer steps for TRM), and computes the aggregated accuracy over all validation boards (board accuracy for Sudoku; shortest path accuracy for Maze).
The validation set $\mathcal{V}_{\text{aug}}$ consists of the first $N_{\text{val}}$ boards of the training set, with one task-specific symmetry-group augmentation applied per board. The best tuple is selected by the aggregated validation accuracy and is then \emph{reused at every $(K, T)$ operating point} in Fig.~\ref{fig:model_task_compute_performance}, treating the sweep as a one-time per-cell compute cost that we account for separately in Section~\ref{sec:compute}.

\begin{table}[b]
  \centering \small
  \caption{Search space and Optuna configuration of the four-axis RoFB hyperparameter sweep.
  $\lambda$ and $\tau$ are sampled log-uniformly; $\alpha$ uniformly; $t_{\min}$ as integers
  in steps of 8 (Kuramoto updates for AKOrN, self-attention updates for ItrSA++, and $L$-level updates for TRM). The Successive Halving Pruner uses $\eta=3$, $\min_{\text{resource}}=1$ for both tasks.}
  \label{tab:hypara_settings}
  \begin{tabular}{lccccccc}
    \toprule
    Task & $\lambda$ & $\alpha$ & $t_{\min}$ & $\tau$ & $N_{\text{val}}$ & $n_{\text{trial}}$ & $n_{\text{chunks}}$ \\
    \midrule
    Sudoku & $[0.01,\,2.0]$  & $[0.01,\,0.50]$ & $\{0,8,\dots,128\}$ & $[0.005,\,2.0]$ & 500 & 30 & 4 \\
    Maze   & $[0.005,\,0.5]$ & $[0.01,\,0.50]$ & $\{0,8,\dots,128\}$ & $[0.005,\,2.0]$ & 100 & 30 & 2 \\
    \bottomrule
  \end{tabular}
\end{table}

$t_{\min}$ is reported in the sweep-internal time unit (Kuramoto steps for AKOrN; self-attention steps for ItrSA++; L-level steps for TRM, where one TRM outer step = 18 L-level steps on Sudoku and 12 on Maze).

\begin{table}[b]
  \centering
  \caption{\textbf{Selected RoFB hyperparameters and validation accuracy.} Aggregated validation accuracies are reported in \%; entries are point estimates.}
  \label{tab:hypara_best_values}
  \begin{tabular}{lcccccc}
    \toprule
    Task $\times$ Model & $\lambda$ & $\alpha$ & $t_{\min}$ & $\tau$ & best val acc & metric \\
    \midrule
    AKOrN $\times$ Sudoku   & 1.949 & 0.281 & 16  & 1.552  & 100.0 & board accuracy \\
    AKOrN $\times$ Maze     & 0.394 & 0.195 & 128 & 0.141  & 75.0 & shortest path accuracy \\
    ItrSA++ $\times$ Sudoku & 0.028 & 0.012 & 64  & 0.461  & 70.6 & board accuracy \\
    ItrSA++ $\times$ Maze   & 0.165 & 0.327 & 104 & 0.028  & 85.0 & shortest path accuracy \\
    TRM $\times$ Sudoku     & 0.829 & 0.182 & 88  & 1.462  & 74.8 & board accuracy \\
    TRM $\times$ Maze       & 0.167 & 0.216 & 88  & 0.939  & 91.0 & shortest path accuracy \\
    \bottomrule
  \end{tabular}
\end{table}

\section{Computational cost of RoFB}
\label{sec:compute}

Here we explain the additional compute introduced by RoFB at test time, relative to a baseline forward pass. The two sources of overhead are (i) the one-time hyperparameter sweep and (ii) the per-step RoFB block inserted into every test forward pass. The former is a fixed cost across all $(K,T)$ operating points, while the latter scales with the baseline compute and shifts the RoFB curve to the right by a constant factor on a log-compute axis. 

We measure the test-time cost of one $(K_{\text{vote}}, T)$ operating point as the total number of forward steps, $(\text{boards}) \times (\text{trajectories per board}) \times (\text{steps per trajectory})$, normalized so that the baseline operating point $(K_{\text{vote}}{=}1, T{=}T_{\max})$ without RoFB equals~$1$. Under this normalization, the cost decomposes into a per-inference forward term and a one-time-per-cell hyperparameter-sweep term:
\begin{align}
\widetilde{\mathcal{C}}^{\text{base}}(K, T)     &= K \cdot \frac{T}{T_{\max}}, \\
\widetilde{\mathcal{C}}^{\text{RoFB-fwd}}(K, T) &= K \cdot \frac{T}{T_{\max}} \cdot r_{\text{tot}}(T), \\
\widetilde{\mathcal{C}}^{\text{RoFB-tot}}(K, T) &= \widetilde{\mathcal{C}}^{\text{RoFB-fwd}}(K, T) + \widetilde{\mathcal{C}}_{\text{sweep}},
\label{eq:rofb_total_compute}
\end{align}
where $K$ is the vote count, $T$ is the inference horizon (in each model's canonical step unit, see Table~\ref{tab:rofb_compute}), $T_{\max}$ is the standard evaluation horizon, $r_{\text{tot}}(T) \geq 1$ is the per-step overhead of an RoFB-augmented forward pass over the baseline (defined in \textbf{\nameref{para:rofb_per_step}}), and $\widetilde{\mathcal{C}}_{\text{sweep}}$ is the hyperparameter-sweep cost amortized over a single test evaluation (defined in \textbf{\nameref{para:rofb_sweep_cost}}).

\begin{figure}[ht]
  \centering
  \includegraphics[width=1.0\textwidth]{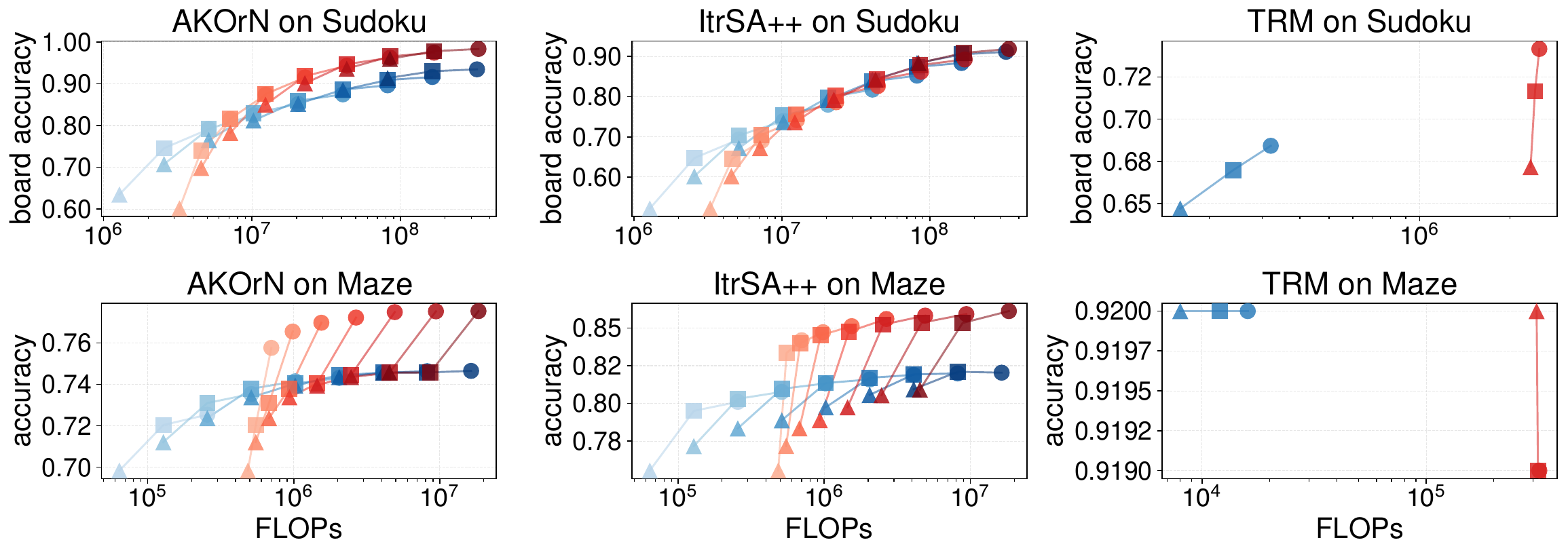}
  \caption{Same operating points as Fig.~\ref{fig:appendix_full_grid}, with the
  per-cell hyperparameter-sweep cost ($\widetilde{\mathcal{C}}_{\text{sweep}}$ in
  Table~\ref{tab:rofb_compute}) added to the RoFB $x$-coordinate. Sweep cost
  shifts the RoFB scatter rigidly to the right by a constant per cell; the
  Pareto improvement is preserved across all six cells.}
  \label{fig:appendix_compute_with_sweep}
\end{figure}

The main result figure (Fig.~\ref{fig:model_task_compute_performance}) plots the per-inference forward cost only --- baseline against $\widetilde{\mathcal{C}}^{\text{RoFB-fwd}}$ --- because the sweep cost is incurred once per cell and is small relative to the K-curve test evaluation suite reported in the figure ($<2\%$ for the four AKOrN/ItrSA++ cells; see \textbf{\nameref{para:rofb_sweep_cost}}). For completeness, the appendix figure Fig.~\ref{fig:appendix_compute_with_sweep} re-plots the same operating points with $\widetilde{\mathcal{C}}^{\text{RoFB-tot}}$ on the $x$-axis, confirming that the Pareto improvement is preserved when the sweep cost is included. We additionally report the same comparison on a wall-clock axis in Fig.~\ref{fig:appendix_compute_walltime_b}.

\paragraph{Hyperparameter-sweep cost.}
\label{para:rofb_sweep_cost}
We pay the sweep cost once per cell, regardless of $(K, T)$. With the lighter sweep protocol of Appendix~\ref{subsec:hp_search} ($n_{\text{trial}}{=}30$ Optuna TPE trials with a SuccessiveHalving pruner, each evaluated on $N_{\text{val}}$ validation puzzles for $T_{\text{sweep}}$ inference steps with $K_{\text{vote}}{=}1$), the FLOP cost of one sweep, expressed in baseline forward steps, is $\mathcal{C}_{\text{sweep}}^{\text{board} \cdot \text{step}} = n_{\text{trial}} \cdot T_{\text{sweep}} \cdot N_{\text{val}} \cdot r_{\text{tot}}(T_{\text{sweep}})$, which we divide by the baseline cost $T_{\max} \cdot N_{\text{test}}$ at $(K_{\text{vote}}{=}1, T{=}T_{\max})$ to obtain the dimensionless $\widetilde{\mathcal{C}}_{\text{sweep}}$ shown in Eq.~\eqref{eq:rofb_total_compute}. The numerical values in Table~\ref{tab:rofb_compute} additionally include a factor $\eta_{\text{ASHA}}{=}0.5$ to reflect that the SuccessiveHalving pruner removes, on average, half of the trials before they reach $T_{\text{sweep}}$ depth.

Across the four AKOrN/ItrSA++ cells, $\widetilde{\mathcal{C}}_{\text{sweep}}$ ranges from $0.38$ to $1.65$ (in units of one baseline test evaluation at $K_{\text{vote}}{=}1$, $T{=}T_{\max}$), which is $<1.7\%$ of the K-curve test evaluation suite ($\sum_{K \in \{1, 2, 4, 8, 16, 32, 64\}} K = 127$ baseline test evaluations) --- the reason the main figure omits this term. For the two TRM cells the same ratio is larger (TRM cells use $K_{\text{vote}}{=}1$ only, so $\sum K = 1$, inflating the ratio to $6.8\times$ on Sudoku and $18.9\times$ on Maze), but the absolute sweep cost is in fact the smallest of any cell ($2.18$M and $0.30$M board $\cdot$ step respectively, see Table~\ref{tab:rofb_compute}); the inflated ratio reflects only the cheap denominator (TRM uses \texttt{bf16} forward and a single trajectory).

\paragraph{Per-step RoFB block cost.}
\label{para:rofb_per_step}
At each inference step where the gate $g(\mathbf{z}_t)$ is non-zero (i.e., $t \geq t_{\min}$, controlled by the warm-up parameter selected by the sweep), the RoFB block adds one $(B, N, N) \times (B, N, D)$ batched matrix multiplication on top of the baseline forward step --- the dominant additional FLOP ($\approx N^2 \cdot D$ MACs per board, where $N$ is the number of cells and $D$ the latent dimension; the readout, similarity, and normalization steps account for less than $15\%$). Writing the active fraction as $\text{active}(T) = \max\bigl(0, \, (T - t_{\min})/T\bigr)$ and the per-step overhead ratio as $r_{\text{step}}$, the per-step total is $r_{\text{tot}}(T) = 1 + r_{\text{step}} \cdot \text{active}(T)$. Because $r_{\text{step}}$ scales as $N^2$ per cell and the baseline forward step scales sub-quadratically in $N$, $r_{\text{step}}$ ranges from $1.2\%$ on TRM~$\times$~Sudoku ($N{=}81$) to $19.5\%$ on AKOrN~$\times$~Maze ($N{=}900$), as shown in Table~\ref{tab:rofb_compute}.

\paragraph{Per-model-task-pair numerical values.}
Table~\ref{tab:rofb_compute} lists the model-task-pair-specific values needed to evaluate Eq.~\eqref{eq:rofb_total_compute}. At the $K_{\text{vote}}{=}1$, $T{=}T_{\max}$ operating point, the per-inference forward overhead $\widetilde{\mathcal{C}}^{\text{RoFB-fwd}}/ \widetilde{\mathcal{C}}^{\text{base}} = r_{\text{tot}}(T_{\max})$ ranges from $1.008$ on TRM~$\times$~Sudoku to $1.098$ on AKOrN~$\times$~Maze. At the largest operating point reported in the main figure ($K{=}64$ for AKOrN/ItrSA++, $K{=}1$ for TRM), the same ratio is essentially unchanged because $r_{\text{tot}}$ does not depend on $K$.

\begin{table}[b]
\centering\small
\caption{
Per-task-pair parameters for the RoFB compute model of Eq.~\eqref{eq:rofb_total_compute}. $T_{\max}$ is each model's standard evaluation horizon (Kuramoto step for AKOrN, self-attention step for ItrSA++, outer block for TRM); $T_{\text{sweep}}$ is the same horizon expressed in the unit used internally for the $t_{\min}$ search (TRM: one outer block $= H_{\mathrm{cycles}}\cdot L_{\mathrm{cycles}}$ $L$-level steps $= 18$ on Sudoku and $12$ on Maze, hence $16\times18 = 288$ and $16\times12 = 192$). $r_{\text{step}}$ is the FLOP fraction of the batched matrix multiplication (\texttt{torch.bmm}) of shape $(B,N,N)\times(B,N,D)$ relative to one baseline forward step. $r_{\text{tot}}^{(T_{\max})}$ is $r_{\text{tot}}(T_{\max})$ evaluated at the cell's $t_{\min}$. $\widetilde{\mathcal{C}}_{\text{sweep}}$ is the amortized sweep cost of Eq.~\eqref{eq:rofb_total_compute}, computed with $n_{\text{trial}}{=}30$, $\eta_{\text{ASHA}}{=}0.5$, and the $(N_{\text{val}}, N_{\text{test}})$ shown. The $\widetilde{\mathcal{C}}_{\text{sweep}}$ column is plotted only on the appendix figures (Figs.~\ref{fig:appendix_compute_with_sweep},~\ref{fig:appendix_compute_walltime_b}); the main figure (Fig.~\ref{fig:model_task_compute_performance}) shows only the per-inference forward overhead.
}
  \label{tab:rofb_compute}
  \begin{tabular}{lrrrrrrrr}
    \toprule
    Model-task pair & $T_{\max}$ & $T_{\text{sweep}}$ & $r_{\text{step}}$ & $t_{\min}$
         & $r_{\text{tot}}^{(T_{\max})}$ & $N_{\text{val}}$ & $N_{\text{test}}$
         & $\widetilde{\mathcal{C}}_{\text{sweep}}$ \\
    \midrule
    AKOrN $\times$ Sudoku   & 256 & 256 & $0.018$ &  16 & $1.017$ & 500 & 20{,}000 & $0.381$ \\
    AKOrN $\times$ Maze     & 256 & 256 & $0.195$ & 128 & $1.098$ & 100 &  1{,}000 & $1.646$ \\
    ItrSA++ $\times$ Sudoku & 256 & 256 & $0.052$ &  64 & $1.039$ & 500 & 20{,}000 & $0.390$ \\
    ItrSA++ $\times$ Maze   & 256 & 256 & $0.160$ & 104 & $1.095$ & 100 &  1{,}000 & $1.643$ \\
    TRM $\times$ Sudoku     &  16 & 288 & $0.012$ &  88 & $1.008$ & 500 & 20{,}000 & $6.806$ \\
    TRM $\times$ Maze       &  16 & 192 & $0.092$ &  88 & $1.050$ & 100 &  1{,}000 & $18.879$ \\
    \bottomrule
  \end{tabular}
\end{table}

\paragraph{Including the sweep cost (appendix figure).}
Fig.~\ref{fig:appendix_compute_with_sweep} re-plots the operating points of
Fig.~\ref{fig:model_task_compute_performance} after adding $\widetilde{\mathcal{C}}_{\text{sweep}}$ to
the RoFB $x$-coordinate (Eq.~\eqref{eq:rofb_total_compute}, third term). The RoFB curve shifts to the right by a constant offset per cell (the values listed in the last column of Table~\ref{tab:rofb_compute}) and the Pareto improvement of RoFB over baseline is preserved on all six cells.

\begin{figure}[h]
  \centering
  \includegraphics[width=1.0\textwidth]{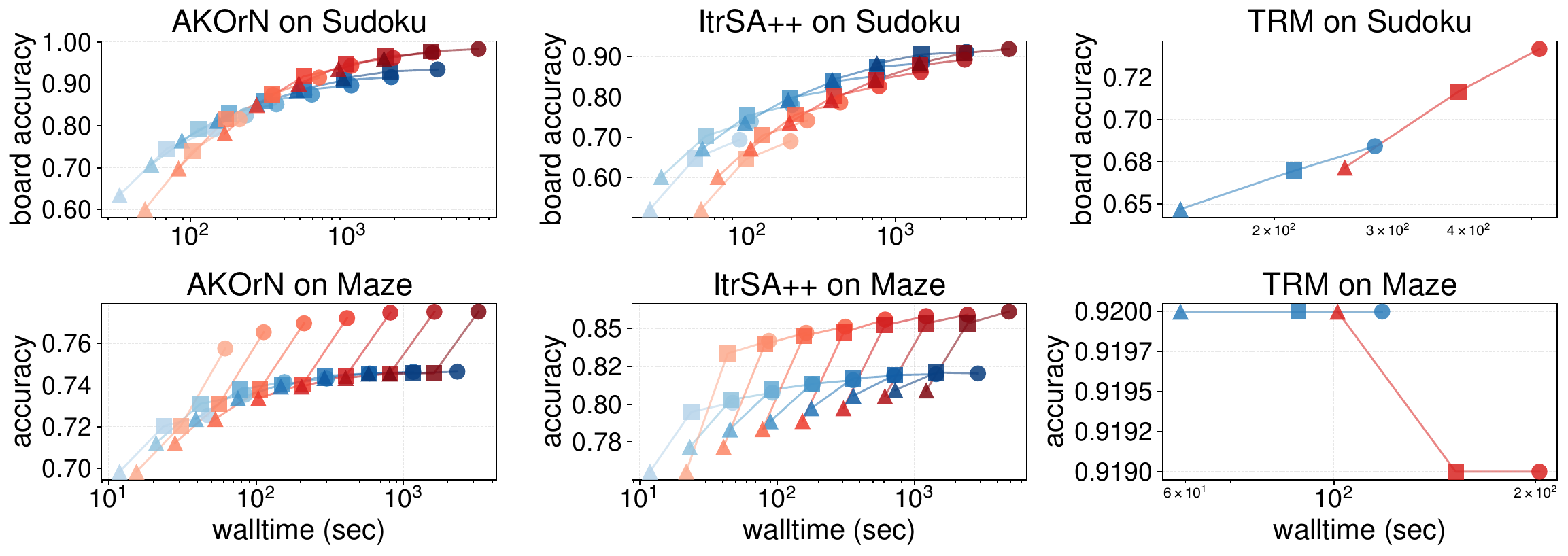}
  \caption{Same operating points as Fig.~\ref{fig:appendix_full_grid}, with the
  $x$-axis replaced by measured wall-clock time on a single NVIDIA GH200
  (forward only, no sweep cost; wall-clock measured in the seed-42 run, accuracies as in Fig.~\ref{fig:appendix_full_grid}). Within-cell baseline-vs-RoFB comparisons are
  dtype-matched (fp32 for AKOrN/ItrSA++; bf16 for TRM); cross-cell wall
  comparisons should be read together with this dtype split.}
  \label{fig:appendix_compute_walltime_a}
\end{figure}

\begin{figure}[h]
  \centering
  \includegraphics[width=1.0\textwidth]{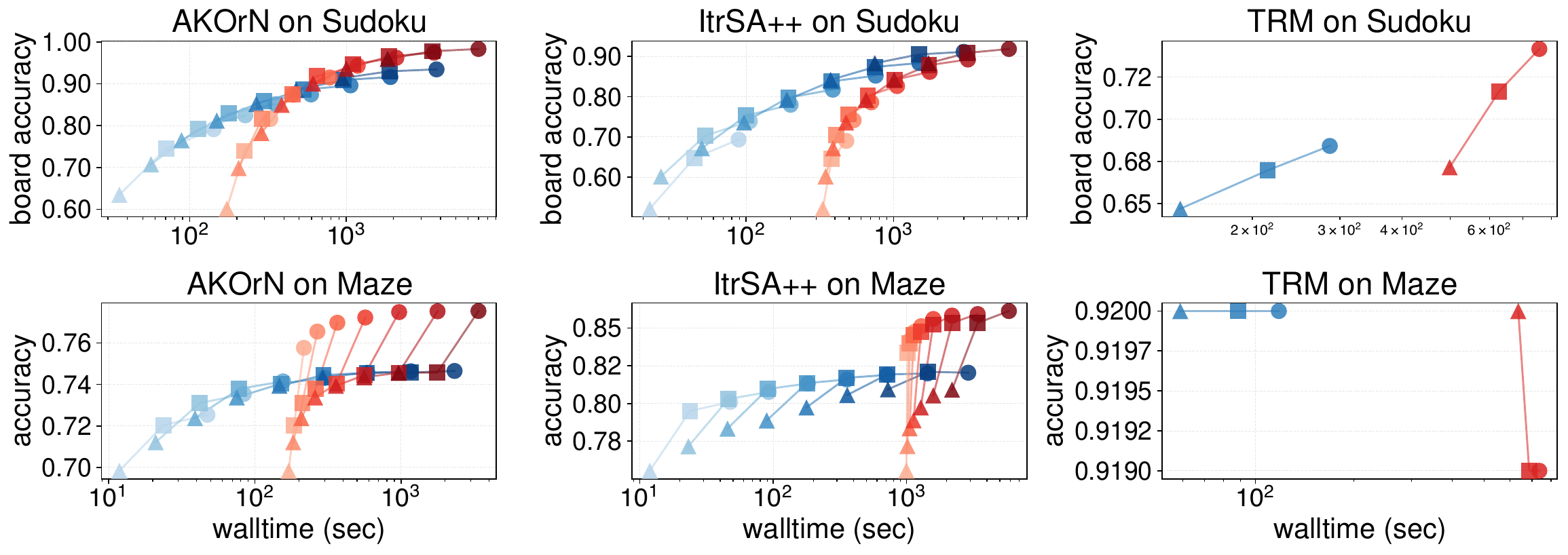}
  \caption{Same as Fig.~\ref{fig:appendix_compute_walltime_a} but with the
  hyperparameter-sweep wall added to the RoFB $x$-coordinate (RoFB only).}
  \label{fig:appendix_compute_walltime_b}
\end{figure}

\paragraph{Wall-clock axis (appendix figure).}
Figs.~\ref{fig:appendix_compute_walltime_a}--\ref{fig:appendix_compute_walltime_b} report the same operating points on a wall-clock $x$-axis instead of FLOPs. Wall and FLOPs are not interchangeable across cells: AKOrN and ItrSA++ run their forward in fp32, while TRM uses bf16, yielding a per-step wall ratio of approximately one order of magnitude between the two dtype families. Within each cell, however, baseline and RoFB share the same forward dtype (Tables \ref{tab:hp_akorn}--\ref{tab:hp_trm}), so the within-cell baseline-vs-RoFB comparison shown in each panel is dtype-matched and therefore fair.

\clearpage

\section{Full Operating-Point Grid and Inference-Seed Robustness}
\label{app:full_grid}
\begin{figure}[ht!]
  \centering
  \includegraphics[width=1.0\textwidth]{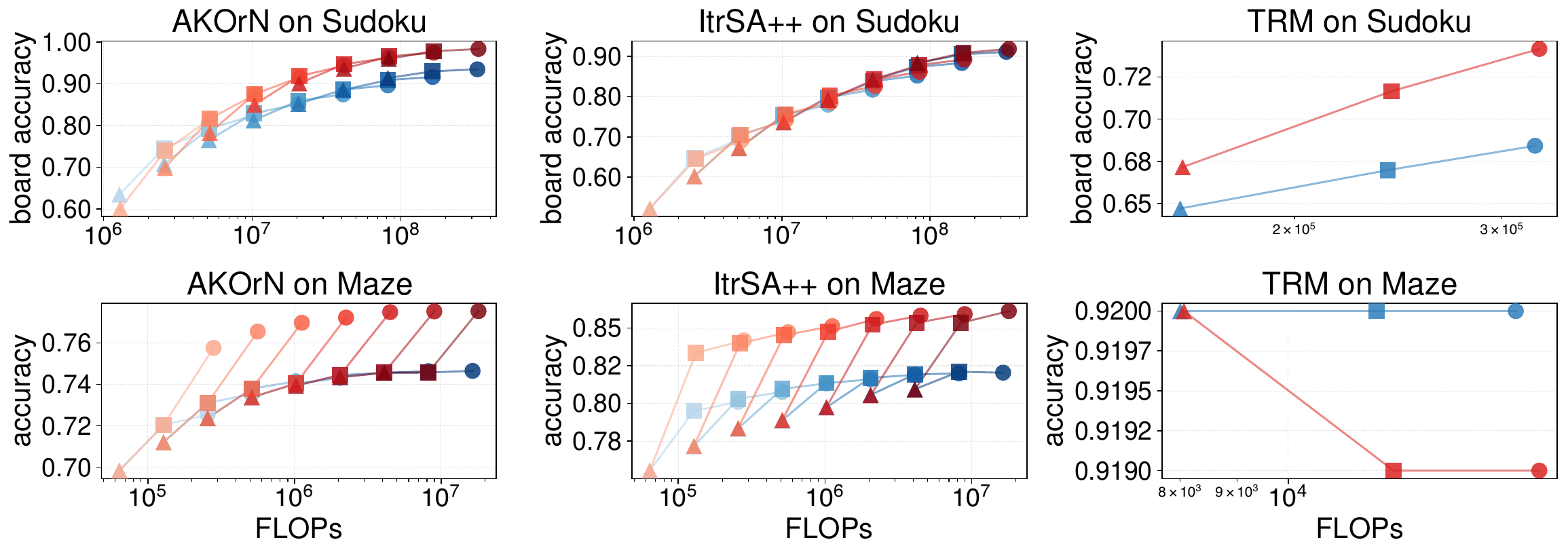}
  \caption{Complete compute-accuracy grid with all $K_{\text{vote}}\in\{1,2,4,8,16,32,64\}$. As in Fig.~\ref{fig:model_task_compute_performance}, each AKOrN and ItrSA++ point is the mean accuracy over 8 inference seeds (42--49) with the same checkpoint, RoFB hyperparameters and test boards; the seed-to-seed SD is at most $0.7$\,pp at every point, so error bars are omitted. TRM has a deterministic initial state and is a single run. The FLOPs axis is analytic and does not depend on the seed. Rows: Sudoku board accuracy, Maze shortest-path accuracy. Encoding as in Fig.~\ref{fig:model_task_compute_performance}.}
  \label{fig:appendix_full_grid}
\end{figure}

Fig.~\ref{fig:appendix_full_grid} shows the complete version of Fig.~\ref{fig:model_task_compute_performance}, including every $K_{\text{vote}}\in\{1,2,4,8,16,32,64\}$. Tables~\ref{tab:full_grid_sudoku} and~\ref{tab:full_grid_maze} list every plotted operating point numerically (Maze: shortest-path accuracy). The tables report a single evaluation run per operating point (inference seed 42; Sudoku test set of 20{,}000 boards, Maze test set of 1{,}000 boards), so their binomial 95\% half-widths are about $\pm0.7$\,pp and $\pm3.1$\,pp, respectively; for AKOrN and ItrSA++ these seed-42 values differ from the 8-seed means plotted in Fig.~\ref{fig:appendix_full_grid} by at most $0.8$\,pp. TRM uses $N_{\text{block}}$ instead of $T_{\text{eval}}$ and has no voting.

\paragraph{Inference-seed replication.}
For the two architectures whose initial latent state is random (AKOrN and ItrSA++), we additionally repeated the $K_{\text{vote}}{=}4$ evaluation with inference seeds $s\in\{42,\dots,49\}$, keeping the checkpoint, RoFB hyperparameters, test boards and board order fixed. The random seed of board $b$ under inference seed $s$ is $10^{6}s+b$, so baseline and RoFB start from the identical initial state for every $(b,s)$ pair, and the per-seed gain
$\Delta_s=\frac{1}{N}\sum_{b}\big(y^{\text{RoFB}}_{b,s}-y^{\text{base}}_{b,s}\big)$
is a paired quantity ($y\in\{0,1\}$ is board-level correctness). Table~\ref{tab:inference_seed} reports the mean and sample standard deviation over the 8 seeds at each $T_{\text{eval}}$, together with 95\% intervals of $\Delta$ at $T_{\text{eval}}{=}256$; these are hierarchical paired bootstraps ($10^{5}$ draws) that resample boards and, within each board, the 8 seed-specific effects. TRM's evaluation-time initial state is a learned constant, so it has no inference-seed axis; Table~\ref{tab:inference_seed} lists its single-run accuracies at $N_{\text{block}}{=}16$ with a paired bootstrap interval over test boards. Note that $K_{\text{vote}}{=}4$ trajectories drawn from one seed are reduced to a single output by confidence-based voting and are therefore not independent replicates; the replicate unit is the seed. For ItrSA++, $T_{\text{eval}}$ counts self-attention steps (4 per outer block). At $T_{\text{eval}}{=}64$ and $128$, RoFB may not yet be active on some boards ($t_{\min}$ in Table~\ref{tab:hypara_best_values}), so the gain there is smaller by construction.

\clearpage

\begin{table}[h]
\centering \footnotesize
\caption{\textbf{Full operating-point grid on Sudoku (board-exact accuracy).}}
\label{tab:full_grid_sudoku}
\setlength{\tabcolsep}{3.5pt}
\begin{tabular}{lrrrrrrrrrr}
\toprule
\textbf{Model} & $K_{\text{vote}}$ & \multicolumn{3}{c}{$T_{\text{eval}}{=}64$ ($N_{\text{block}}{=}8$)} & \multicolumn{3}{c}{$T_{\text{eval}}{=}128$ ($N_{\text{block}}{=}12$)} & \multicolumn{3}{c}{$T_{\text{eval}}{=}256$ ($N_{\text{block}}{=}16$)} \\
\cmidrule(lr){3-5} \cmidrule(lr){6-8} \cmidrule(lr){9-11}
 & & Base & RoFB & $\Delta$ & Base & RoFB & $\Delta$ & Base & RoFB & $\Delta$ \\
\midrule
AKOrN & 1 & 63.3 & 59.8 & $-3.6$ & 74.4 & 73.9 & $-0.5$ & 79.1 & 81.7 & $+2.6$ \\
 & 2 & 70.6 & 69.7 & $-0.9$ & 79.2 & 81.8 & $+2.7$ & 82.4 & 87.5 & $+5.0$ \\
 & 4 & 76.3 & 78.3 & $+2.0$ & 82.9 & 87.9 & $+5.0$ & 85.1 & 91.5 & $+6.4$ \\
 & 8 & 81.2 & 85.1 & $+3.9$ & 85.9 & 91.8 & $+5.9$ & 87.5 & 94.2 & $+6.7$ \\
 & 16 & 85.3 & 90.4 & $+5.1$ & 88.7 & 94.8 & $+6.1$ & 89.6 & 96.2 & $+6.6$ \\
 & 32 & 88.6 & 93.8 & $+5.2$ & 91.0 & 96.6 & $+5.5$ & 91.6 & 97.5 & $+5.8$ \\
 & 64 & 91.4 & 96.0 & $+4.6$ & 93.0 & 97.9 & $+4.9$ & 93.5 & 98.4 & $+4.9$ \\
\midrule
ItrSA++ & 1 & 52.2 & 52.2 & $+0.0$ & 64.7 & 64.6 & $-0.1$ & 69.2 & 69.1 & $-0.1$ \\
 & 2 & 60.2 & 60.2 & $-0.0$ & 70.4 & 70.5 & $+0.2$ & 73.8 & 74.1 & $+0.4$ \\
 & 4 & 67.0 & 67.0 & $+0.0$ & 75.2 & 75.4 & $+0.2$ & 77.7 & 78.3 & $+0.6$ \\
 & 8 & 73.7 & 73.7 & $-0.0$ & 79.9 & 80.1 & $+0.3$ & 81.8 & 82.6 & $+0.8$ \\
 & 16 & 79.2 & 79.2 & $+0.0$ & 83.8 & 84.1 & $+0.3$ & 85.2 & 86.1 & $+0.9$ \\
 & 32 & 84.2 & 84.2 & $+0.0$ & 87.4 & 87.7 & $+0.3$ & 88.4 & 89.1 & $+0.6$ \\
 & 64 & 88.3 & 88.3 & $-0.0$ & 90.5 & 90.8 & $+0.3$ & 91.2 & 91.8 & $+0.7$ \\
\midrule
TRM & 1 & 64.7 & 67.2 & $+2.5$ & 67.0 & 71.7 & $+4.7$ & 68.4 & 74.2 & $+5.7$ \\
\bottomrule
\end{tabular}
\resulttabnote{Accuracies are reported in \%; $\Delta=\mathrm{RoFB}-\mathrm{Baseline}$ is in percentage points (pp). Entries are point estimates from a single evaluation run.}
\end{table}

\begin{table}[h]
\centering \footnotesize
\caption{\textbf{Full operating-point grid on Maze (shortest-path accuracy).}}
\label{tab:full_grid_maze}
\setlength{\tabcolsep}{3.5pt}
\begin{tabular}{lrrrrrrrrrr}
\toprule
\textbf{Model} & $K_{\text{vote}}$ & \multicolumn{3}{c}{$T_{\text{eval}}{=}64$ ($N_{\text{block}}{=}8$)} & \multicolumn{3}{c}{$T_{\text{eval}}{=}128$ ($N_{\text{block}}{=}12$)} & \multicolumn{3}{c}{$T_{\text{eval}}{=}256$ ($N_{\text{block}}{=}16$)} \\
\cmidrule(lr){3-5} \cmidrule(lr){6-8} \cmidrule(lr){9-11}
 & & Base & RoFB & $\Delta$ & Base & RoFB & $\Delta$ & Base & RoFB & $\Delta$ \\
\midrule
AKOrN & 1 & 69.8 & 69.8 & $+0.0$ & 72.2 & 72.2 & $+0.0$ & 72.6 & 75.8 & $+3.2$ \\
 & 2 & 71.6 & 71.6 & $+0.0$ & 73.4 & 73.4 & $+0.0$ & 73.7 & 76.5 & $+2.8$ \\
 & 4 & 73.1 & 73.1 & $+0.0$ & 74.0 & 74.0 & $+0.0$ & 74.1 & 77.2 & $+3.1$ \\
 & 8 & 73.8 & 73.8 & $+0.0$ & 74.0 & 74.0 & $+0.0$ & 73.9 & 76.9 & $+3.0$ \\
 & 16 & 73.8 & 73.8 & $+0.0$ & 74.3 & 74.3 & $+0.0$ & 74.3 & 77.2 & $+2.9$ \\
 & 32 & 74.1 & 74.1 & $+0.0$ & 74.5 & 74.5 & $+0.0$ & 74.6 & 77.4 & $+2.8$ \\
 & 64 & 74.4 & 74.4 & $+0.0$ & 75.0 & 75.0 & $+0.0$ & 75.1 & 77.8 & $+2.7$ \\
\midrule
ItrSA++ & 1 & 75.8 & 75.8 & $+0.0$ & 79.8 & 83.6 & $+3.8$ & 80.0 & 85.0 & $+5.0$ \\
 & 2 & 76.9 & 76.9 & $+0.0$ & 80.1 & 84.0 & $+3.9$ & 80.6 & 85.1 & $+4.5$ \\
 & 4 & 77.9 & 77.9 & $+0.0$ & 80.5 & 84.5 & $+4.0$ & 81.1 & 85.3 & $+4.2$ \\
 & 8 & 78.9 & 78.9 & $+0.0$ & 81.5 & 85.0 & $+3.5$ & 81.7 & 85.6 & $+3.9$ \\
 & 16 & 79.3 & 79.3 & $+0.0$ & 81.9 & 85.6 & $+3.7$ & 82.1 & 85.9 & $+3.8$ \\
 & 32 & 80.5 & 80.5 & $+0.0$ & 81.7 & 85.4 & $+3.7$ & 81.6 & 85.9 & $+4.3$ \\
 & 64 & 80.6 & 80.6 & $+0.0$ & 81.8 & 85.6 & $+3.8$ & 81.7 & 85.9 & $+4.2$ \\
\midrule
TRM & 1 & 92.0 & 92.0 & $+0.0$ & 92.0 & 91.9 & $-0.1$ & 92.0 & 91.9 & $-0.1$ \\
\bottomrule
\end{tabular}
\resulttabnote{Accuracies are reported in \%; $\Delta=\mathrm{RoFB}-\mathrm{Baseline}$ is in percentage points (pp). Entries are point estimates from a single evaluation run.}
\end{table}

\begin{table}[h]
\centering \footnotesize
\caption{\textbf{Paired gains and inference-seed robustness at $K_{\text{vote}}{=}4$.}}
\label{tab:inference_seed}
\setlength{\tabcolsep}{4pt}
\resizebox{\textwidth}{!}{\begin{tabular}{lrrrrlrr}
\toprule
\textbf{Model on Task} & $T_{\text{eval}}$ & Baseline & RoFB & $\Delta$ & 95\% CI of $\Delta$ & $\Delta$ min\,--\,max & Seeds with $\Delta>0$ \\
\midrule
AKOrN on Sudoku & 64 & 76.5 $\pm$ 0.2 & 78.2 $\pm$ 0.1 & $+1.71$ $\pm$ 0.23 & -- & $+1.34$\,--\,$+1.98$ & 8/8 \\
 & 128 & 82.9 $\pm$ 0.1 & 87.5 $\pm$ 0.1 & $+4.61$ $\pm$ 0.17 & -- & $+4.41$\,--\,$+4.96$ & 8/8 \\
 & 256 & 85.1 $\pm$ 0.1 & 91.5 $\pm$ 0.1 & $+6.38$ $\pm$ 0.19 & $[6.05,\, 6.70]$ & $+6.18$\,--\,$+6.66$ & 8/8 \\
\midrule
AKOrN on Maze & 64 & 72.4 $\pm$ 0.5 & 72.4 $\pm$ 0.5 & $+0.00$ $\pm$ 0.00 & -- & $+0.00$\,--\,$+0.00$ & 0/8 \\
 & 128 & 73.8 $\pm$ 0.5 & 73.8 $\pm$ 0.5 & $+0.00$ $\pm$ 0.00 & -- & $+0.00$\,--\,$+0.00$ & 0/8 \\
 & 256 & 74.2 $\pm$ 0.4 & 77.0 $\pm$ 0.3 & $+2.83$ $\pm$ 0.25 & $[1.49,\, 4.21]$ & $+2.50$\,--\,$+3.20$ & 8/8 \\
\midrule
ItrSA++ on Sudoku & 64 & 67.2 $\pm$ 0.2 & 67.2 $\pm$ 0.2 & $-0.01$ $\pm$ 0.03 & -- & $-0.06$\,--\,$+0.02$ & 3/8 \\
 & 128 & 75.3 $\pm$ 0.1 & 75.6 $\pm$ 0.2 & $+0.29$ $\pm$ 0.07 & -- & $+0.22$\,--\,$+0.42$ & 8/8 \\
 & 256 & 78.0 $\pm$ 0.2 & 78.6 $\pm$ 0.2 & $+0.61$ $\pm$ 0.07 & $[0.46,\, 0.76]$ & $+0.54$\,--\,$+0.71$ & 8/8 \\
\midrule
ItrSA++ on Maze & 64 & 78.3 $\pm$ 0.4 & 78.4 $\pm$ 0.4 & $+0.01$ $\pm$ 0.04 & -- & $+0.00$\,--\,$+0.10$ & 1/8 \\
 & 128 & 81.0 $\pm$ 0.3 & 84.5 $\pm$ 0.1 & $+3.56$ $\pm$ 0.27 & -- & $+3.20$\,--\,$+4.00$ & 8/8 \\
 & 256 & 81.3 $\pm$ 0.2 & 85.1 $\pm$ 0.3 & $+3.80$ $\pm$ 0.40 & $[2.65,\, 5.01]$ & $+3.20$\,--\,$+4.30$ & 8/8 \\
\midrule
TRM on Sudoku & 16$^{\dagger}$ & 68.4 & 74.2 & $+5.75$ & $[5.32,\, 6.20]$ & n/a & n/a \\
TRM on Maze & 16$^{\dagger}$ & 92.0 & 91.9 & $-0.10$ & $[-0.30,\, 0.00]$ & n/a & n/a \\
\bottomrule
\end{tabular}}
\resulttabnote{Accuracies are reported in \%; paired gains, their intervals and their ranges are in percentage points (pp). For AKOrN and ItrSA++, values are means $\pm$ sample SD across eight inference seeds (42--49); for each seed, baseline and RoFB start from the identical initial latent state. The 95\% CI at $T_{\text{eval}}{=}256$ is a hierarchical paired bootstrap ($10^{5}$ draws) over boards and, within each board, the eight seed-specific effects. The last two columns report the minimum--maximum paired gain and the number of seeds with positive gain. TRM has a learned, deterministic initial state and no inference-seed axis; it is a single run without voting, and its CI is a paired bootstrap over test boards. $^{\dagger}$For TRM, the entry is $N_{\text{block}}$.}
\end{table}

\clearpage

\section{Hyperparameter Sensitivity}

\label{app:hp_sensitivity}

In this section, we examine whether the gains from RoFB persist when its hyperparameters are changed. The results depend on the model, task, and parameter:

\begin{itemize}
\item Positive gains were robust to the tested hyperparameter changes in AKOrN on Maze, ItrSA++ on Maze, and TRM on Sudoku, remaining positive at every tested endpoint.
\item Some model--task pairs show hyperparameter sensitivity: AKOrN on Sudoku can fall below its baseline, and ItrSA++ on Maze can lose most of its gain when the entropy threshold is increased.
\item No improvement was confirmed at the reference settings for ItrSA++ on Sudoku or TRM on Maze.
\end{itemize}

\begin{figure}[!tbp]
\centering
\includegraphics[width=0.96\textwidth,trim=0 6 0 25,clip]{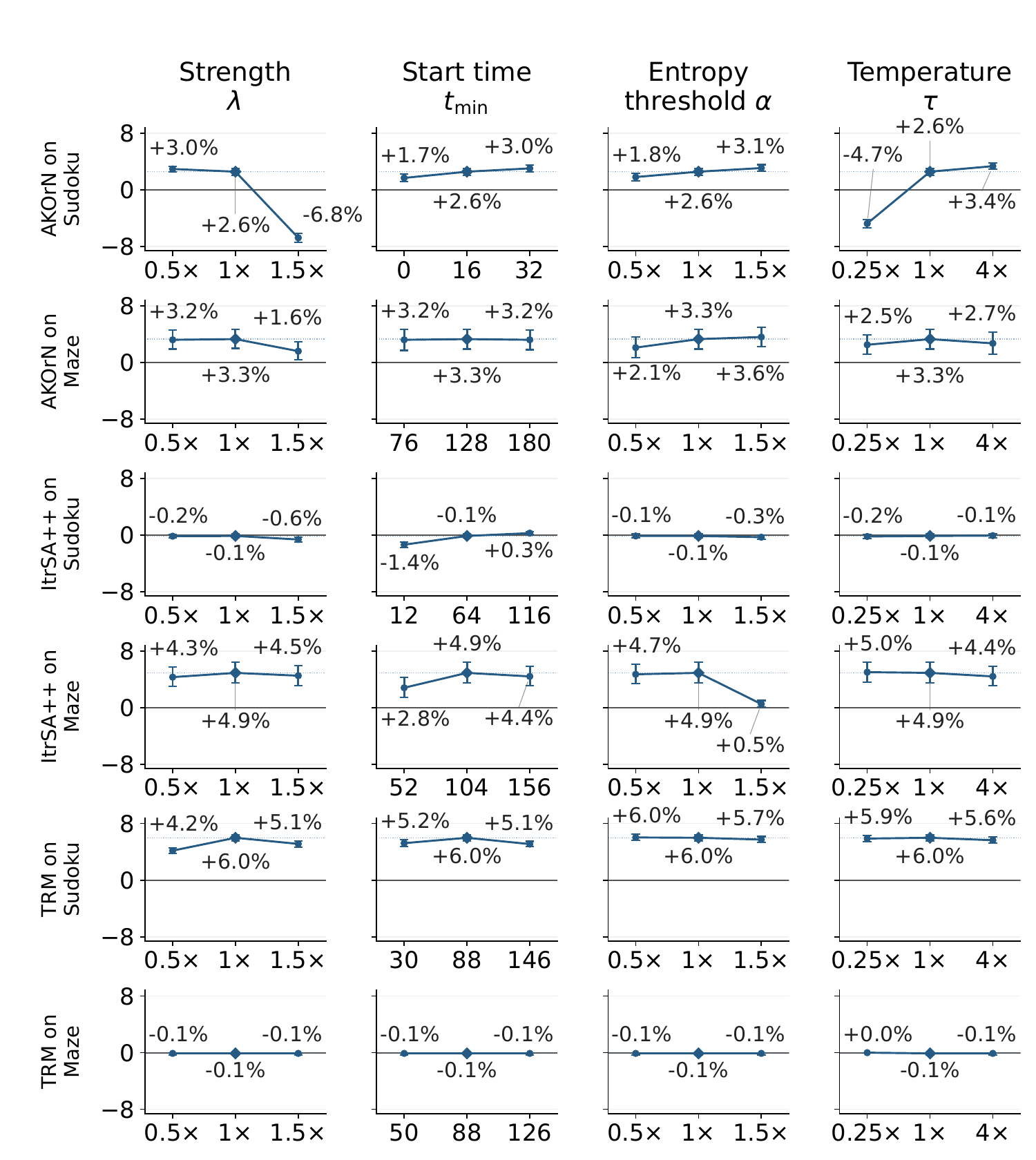}
\caption{Test-set gains when each hyperparameter is varied separately.  Rows identify the model and task; columns identify the varied parameter. The vertical axis shows accuracy gains in percentage points, on a common scale across all panels and using the baselines described above. Bars show pointwise 95\% confidence intervals; diamonds and dotted lines mark the reference setting and its gain.}
\label{fig:hp_sensitivity}
\end{figure}

We describe the evaluation settings and summarize the sensitivity of each model--task pair in Fig.~\ref{fig:hp_sensitivity}.

\paragraph{Evaluation setting.}
We vary the feedback strength $\lambda$, intervention start time $t_{\min}$, entropy threshold $\alpha$, and gate temperature $\tau$ one at a time around a fixed reference tuple for each model--task pair. All evaluations use repulsive, entropy-gated RoFB; ItrSA++ uses projected feedback.

Each model--task pair uses a fixed checkpoint, one trajectory per board, and paired inputs across conditions. The inference seed is 42 for the stochastic models; TRM uses its learned deterministic initialization. AKOrN runs for 256 recurrent updates, ItrSA++ for 64 outer blocks containing 256 self-attention updates, and TRM for 16 outer iterations.

We evaluate five values per parameter on 500 validation boards per pair, and the prespecified lowest, reference, and highest values on 20,000 Sudoku or 1,000 Maze test boards. For each model--task pair, we perform four one-at-a-time sweeps, one for each hyperparameter, while keeping the other three fixed. The reference-setting evaluation is shared across these sweeps. The validation boards use training-derived augmentations. Table~\ref{tab:hp_sensitivity_ranges} gives the perturbation ranges: a star denotes the reference value, and $d$ is the integer spacing between adjacent start times. The other three parameters are held fixed, except that AKOrN on Sudoku uses rounded center values and full-precision fixed coordinates at the endpoints.

\begin{table}[!b]
\centering\small
\caption{Hyperparameter ranges for validation and test.}
\label{tab:hp_sensitivity_ranges}
\begin{tabular}{@{}p{0.25\linewidth}p{0.40\linewidth}p{0.28\linewidth}@{}}
\toprule
Parameter & Five validation values relative to the selected value & Test values \\
\midrule
Feedback strength $\lambda$ & $\{0.5,0.75,1,1.25,1.5\}\lambda^*$ & $\{0.5,1,1.5\}\lambda^*$ \\
Start time $t_{\min}$ & $t_{\min}^*+\{-2,-1,0,1,2\}d$ & $t_{\min}^*+\{-2,0,2\}d$ \\
Entropy threshold $\alpha$ & $\{0.5,0.75,1,1.25,1.5\}\alpha^*$ & $\{0.5,1,1.5\}\alpha^*$ \\
Gate temperature $\tau$ & $\{0.25,0.5,1,2,4\}\tau^*$ & $\{0.25,1,4\}\tau^*$ \\
\bottomrule
\end{tabular}
\end{table}

For the start-time grid, $(t_{\min}^*,d)$ is $(16,8)$ and $(128,26)$ for AKOrN on Sudoku and Maze, $(64,26)$ and $(104,26)$ for ItrSA++, and $(88,29)$ and $(88,19)$ for TRM. ItrSA++ permits feedback at one-based update $t>t_{\min}$ through update 256. The spacing $d$ is set to 10\% of the total update horizon, rounded to the nearest integer, and reduced if needed so that two earlier and two later settings fit between zero and the last possible start time. For AKOrN on Sudoku, the reference start at update 16 restricts the spacing to 8; the other pairs use the rounded 10\% spacing.

\paragraph{Controls.}
We compare each setting with a no-feedback baseline on the same boards. For ItrSA++, the test comparisons in Fig.~\ref{fig:hp_sensitivity} use the same update procedure with feedback strength set to zero and the reference start time fixed; its validation comparisons use the ordinary model. We additionally check a no-feedback control matched to each intervention start time and the ordinary model, because the extra normalization in the feedback procedure can slightly change the numerical trajectory even when feedback strength is zero.

\paragraph{Metrics and uncertainty.}
We report the difference in accuracy between RoFB and its corresponding no-feedback baseline in percentage points, using exact-solution accuracy for Sudoku and shortest-path accuracy for Maze. Pointwise 95\% confidence intervals use 100,000 paired bootstrap draws over the same test problems. These intervals describe uncertainty for a fixed checkpoint and inference condition and are not corrected for multiple comparisons.

For each hyperparameter, we check whether the confidence intervals at the lower, reference, and upper settings are all above zero.

\paragraph{Improvements persist across multiple settings.}

AKOrN on Maze, ItrSA++ on Maze, and TRM on Sudoku retain positive gains at all eight tested endpoints, with pointwise 95\% confidence intervals entirely above zero. Fig.~\ref{fig:hp_sensitivity} shows the gains at the lower, reference, and upper settings for each hyperparameter and all six model--task pairs.

The results distinguish persistence of a benefit from preservation of its magnitude. For AKOrN on Maze, ItrSA++ on Maze, and TRM on Sudoku, changing any one parameter to either tested endpoint preserves a positive gain. Thus, their improvements do not depend on reproducing one exact value of any single parameter. However, a positive gain can still become much smaller, so this observation alone does not justify treating parameter selection as unimportant.

\clearpage

\section{Mechanistic Analysis of RoFB (1): Inter-class Separation}
\label{app:mechanistic_analysis_inter_class_separation}
In this subsection, we provide a detailed mechanistic analysis of how RoFB influences the latent space and model trajectories. 

\subsection{RoFB promotes inter-class separation.}

We quantify inter-class separation using the same metrics, cosine gap $\Delta_{\cos}=C_{\mathrm{same}}-C_{\mathrm{diff}}$ and the cosine-distance silhouette score, described in Appendix~\ref{app:clustering_in_correct_solutions}. $C_{\mathrm{same}}$ and $C_{\mathrm{diff}}$ are the mean cosine similarities between distinct cells with the same and different reference labels, respectively. We use reference digits in Sudoku and reference path versus non-path labels in Maze, restricting the analysis to blank Sudoku cells and Maze passage cells excluding the start and goal. Larger values of either index indicate stronger separation. For silhouette, we average only over cells with defined comparisons, excluding singleton classes, and use a denominator floor of $10^{-12}$; Appendix~\ref{app:clustering_in_correct_solutions} instead assigns singleton cells zero contribution and retains them in the averaging denominator.

We compare terminal latent states under RoFB and the no-force reference on 2,000 Sudoku instances and 900 Maze instances per model, with one candidate per instance and matched initial states. The standard inference horizon is 256 updates for AKOrN, 256 self-attention updates (64 blocks) for ItrSA++, and 16 outer iterations for TRM. RoFB replays the gate sequence recorded from the no-force reference. Reference labels are used for geometry analysis, not by standard RoFB. Reported changes are RoFB minus the no-force reference; geometry values are dimensionless, and 95\% confidence intervals (CIs) use 20,000 paired bootstrap resamples of instances.

Both indices increased with RoFB in all six model--task pairs when averaged over all instances (Table~\ref{tab:inter_class_separation}). These averages include instances that were correct under both conditions, incorrect under both conditions, or changed correctness between conditions.

\begin{table}[!bp]
\centering\small
\caption{\textbf{RoFB increases terminal inter-class separation across all instances.} Values are dimensionless mean differences (RoFB minus the no-force reference). Brackets show 95\% paired-bootstrap confidence intervals (CIs) over instances.}
\label{tab:inter_class_separation}
\setlength{\tabcolsep}{5pt}
\begin{tabular}{lrrr}
\toprule
\textbf{Geometry measure} & \textbf{AKOrN} & \textbf{ItrSA++} & \textbf{TRM} \\
\midrule
\multicolumn{4}{l}{\textit{Sudoku: all instances}} \\[2pt]
Change in $\Delta_{\cos}$ & \estimateci{+0.0412}{0.0360}{0.0464} & \estimateci{+0.0086}{0.0075}{0.0097} & \estimateci{+0.1580}{0.1532}{0.1628} \\[3pt]
Change in silhouette & \estimateci{+0.0508}{0.0361}{0.0654} & \estimateci{+0.0185}{0.0124}{0.0248} & \estimateci{+0.1487}{0.1411}{0.1565} \\[3pt]
\midrule
\multicolumn{4}{l}{\textit{Maze: all instances}} \\[2pt]
Change in $\Delta_{\cos}$ & \estimateci{+0.1805}{0.1692}{0.1921} & \estimateci{+0.1171}{0.1124}{0.1220} & \estimateci{+0.0803}{0.0797}{0.0808} \\[3pt]
Change in silhouette & \estimateci{+0.0410}{0.0381}{0.0439} & \estimateci{+0.1000}{0.0964}{0.1038} & \estimateci{+0.0325}{0.0323}{0.0327} \\[3pt]
\bottomrule
\end{tabular}
\end{table}

To examine whether the aggregate increase was explained solely by a larger fraction of correct instances, we restricted the same runs and evaluation settings used in Table~\ref{tab:inter_class_separation} to instances that were incorrect under both RoFB and the no-force reference. Correctness requires agreement with the reference solution in Sudoku and a valid shortest path in Maze, including alternative shortest paths. For each model--task pair, we recomputed the mean changes and conditional 95\% CIs by resampling paired instances within this subset; the subset sizes are reported in Table~\ref{tab:inter_class_separation_both_incorrect}. Both indices increased in all six pairs, with positive 95\% CIs throughout. Table~\ref{tab:inter_class_separation_both_incorrect} reports these subset averages, whereas Table~\ref{tab:inter_class_separation} reports averages over all instances.

\begin{table}[!bp]
\centering\small
\caption{\textbf{RoFB increases separation among instances incorrect under both conditions.}}
\label{tab:inter_class_separation_both_incorrect}
\setlength{\tabcolsep}{5pt}
\begin{tabular}{lrrr}
\toprule
\textbf{Geometry measure} & \textbf{AKOrN} & \textbf{ItrSA++} & \textbf{TRM} \\
\midrule
\multicolumn{4}{l}{\textit{Sudoku: incorrect under both conditions}} \\[2pt]
Number of instances & 271 & 566 & 477 \\[3pt]
Change in $\Delta_{\cos}$ & \estimateci{+0.0082}{0.0070}{0.0094} & \estimateci{+0.0026}{0.0016}{0.0035} & \estimateci{+0.0191}{0.0161}{0.0221} \\[3pt]
Change in silhouette & \estimateci{+0.0095}{0.0031}{0.0159} & \estimateci{+0.0055}{0.0025}{0.0085} & \estimateci{+0.0270}{0.0235}{0.0307} \\[3pt]
\midrule
\multicolumn{4}{l}{\textit{Maze: incorrect under both conditions}} \\[2pt]
Number of instances & 209 & 140 & 72 \\[3pt]
Change in $\Delta_{\cos}$ & \estimateci{+0.2291}{0.2043}{0.2538} & \estimateci{+0.1219}{0.1075}{0.1370} & \estimateci{+0.0604}{0.0581}{0.0627} \\[3pt]
Change in silhouette & \estimateci{+0.0518}{0.0460}{0.0578} & \estimateci{+0.1084}{0.0973}{0.1200} & \estimateci{+0.0271}{0.0264}{0.0278} \\[3pt]
\bottomrule
\end{tabular}
\resulttabnote{Values are dimensionless mean differences (RoFB minus the no-force reference). Brackets show conditional 95\% paired-bootstrap CIs obtained by resampling instances within the subset incorrect under both conditions.}
\end{table}

The increases among instances that remained incorrect show that improved separation was not confined to instances that became correct. These correctness groups are defined after intervention and provide descriptive stratification, rather than causal adjustment for correctness. Separation also does not guarantee a correct solution: TRM on Maze showed increases in both geometry indices while retaining the same instance-level correctness as the no-force reference.

\paragraph{Ablating inter-class separation component hinders performance.}
To see the effect of inter-class separation on performance gain, we next performed inference with RoFB while removing the component of the gradient of inter-class similarity. Specifically, we removed the projection of the feedback force onto the direction that changes $C_{\mathrm{diff}}$ to first order.

We compute the direction in external-force space along which inter-class similarity $C_{\mathrm{diff}}$ increases most rapidly around zero force:

\begin{equation}
\mathbf d_t=\left.\nabla_{\mathbf w}C_{\mathrm{diff}}\bigl(F_t(\mathbf w)\bigr)\right|_{\mathbf w=\mathbf 0}.
\end{equation}

Here $\mathbf w$ is an external force, and $F_t(\mathbf w)$ is the next latent state obtained by applying it at update $t$:

\begin{equation}
F_t(\mathbf w)=\Pi_{\mathcal M}\!\left(R(\mathbf z_t)+L_t(\mathbf w)\right).
\end{equation}

$R(\mathbf z_t)$ is the ordinary model update before feedback and final normalization, also denoted by $\mathbf r_t$ below. $L_t$ projects the external force onto the tangent space at the current state, including the native step size for AKOrN; $\Pi_{\mathcal M}$ is the model's final normalization. The current state and other inputs to this one-update map are held fixed when computing the gradient.

Let $\mathbf v_t$ be the external force supplied by RoFB, including its strength and gate value. We remove its component along $\mathbf d_t$ and perform inference with the remaining force:

\begin{equation}
\mathbf v_t^{\mathrm{sep}}=
\frac{\langle\mathbf d_t,\mathbf v_t\rangle}{\|\mathbf d_t\|^2}\mathbf d_t,
\qquad
\mathbf v_t^{\mathrm{rem}}=\mathbf v_t-\mathbf v_t^{\mathrm{sep}},
\qquad
\mathbf z_{t+1}=F_t(\mathbf v_t^{\mathrm{rem}}).
\end{equation}

The inner product sums over all blank cells and latent coordinates of one candidate. The remaining force is orthogonal to $\mathbf d_t$, so it induces no first-order change in $C_{\mathrm{diff}}$ relative to the zero-force update; finite forces may still induce higher-order changes. The removed component can point in either direction along $\mathbf d_t$. Reference labels define $C_{\mathrm{diff}}$ for this diagnostic intervention and are not used by standard RoFB.

Removing a component also changes the force magnitude. We therefore compare the remaining force with the original RoFB direction scaled down to the same magnitude, and separately compare the remaining direction scaled back up to the original magnitude with unmodified RoFB. Magnitudes are measured over all blank cells of one candidate, after model-specific linear projection and step scaling and before nonlinear normalization. In the notation above, the two magnitude controls are

\begin{align}
n_t(\mathbf w)
&= \|L_t(\mathbf w)\|_F, \\[2pt]
\mathbf v_t^{\mathrm{scaled}}
&= \frac{n_t(\mathbf v_t^{\mathrm{rem}})}{n_t(\mathbf v_t)}\mathbf v_t,
\qquad n_t(\mathbf v_t^{\mathrm{scaled}}) = n_t(\mathbf v_t^{\mathrm{rem}}), \\[2pt]
\mathbf v_t^{\mathrm{restored}}
&= \frac{n_t(\mathbf v_t)}{n_t(\mathbf v_t^{\mathrm{rem}})}\mathbf v_t^{\mathrm{rem}},
\qquad n_t(\mathbf v_t^{\mathrm{restored}}) = n_t(\mathbf v_t).
\end{align}

The Frobenius norm treats the stacked vector as a blank-cell-by-feature array and is equal to its Euclidean norm. The two comparisons in Table~\ref{tab:separation_component_removal} are therefore $\mathbf v_t^{\mathrm{rem}}$ versus $\mathbf v_t^{\mathrm{scaled}}$, and $\mathbf v_t^{\mathrm{restored}}$ versus $\mathbf v_t$. These equalities hold at each condition's own state, not between the accumulated displacements of two diverging trajectories. We additionally remove random directions with matched removal and remaining magnitudes. The gate sequence is recorded from the no-force trajectory and replayed across conditions; directions and gradients are recomputed at each condition's current state throughout the intervention window. Numerically undefined component operations contribute zero force without dropping the instance. Online-gated RoFB is retained as a separate reference.

All pairs use the same sample sizes, one-candidate protocol, and standard horizons as Table~\ref{tab:inter_class_separation}.

\begin{table}[!bp]
\centering\small
\caption{\textbf{Accuracy differences after removing the separation-related component.}}
\label{tab:separation_component_removal}
\setlength{\tabcolsep}{5pt}
\begin{tabular}{lrrr}
\toprule
\textbf{Intervention} & \textbf{AKOrN} & \textbf{ItrSA++} & \textbf{TRM} \\
\midrule
\multicolumn{4}{l}{\textit{Sudoku}} \\[2pt]
\shortstack[l]{Removal $-$\\ magnitude-matched RoFB} & \estimateci{-2.300}{-3.400}{-1.200} & \estimateci{-2.700}{-3.550}{-1.900} & \estimateci{-3.250}{-4.400}{-2.100} \\[3pt]
\shortstack[l]{Magnitude-restored\\ removal $-$ RoFB} & \estimateci{-18.250}{-20.000}{-16.500} & \estimateci{-7.150}{-8.300}{-6.050} & \estimateci{-6.950}{-8.300}{-5.550} \\[3pt]
\midrule
\multicolumn{4}{l}{\textit{Maze}} \\[2pt]
\shortstack[l]{Removal $-$\\ magnitude-matched RoFB} & \estimateci{-8.889}{-10.889}{-7.000} & \estimateci{-11.000}{-13.111}{-9.000} & \estimateci{+0.111}{0.000}{0.333} \\[3pt]
\shortstack[l]{Magnitude-restored\\ removal $-$ RoFB} & \estimateci{-12.222}{-14.444}{-10.111} & \estimateci{-22.444}{-25.222}{-19.778} & \estimateci{-0.111}{-0.333}{0.000} \\[3pt]
\bottomrule
\end{tabular}
\resulttabnote{Values are paired accuracy differences in percentage points (pp), using the comparator specified in each row. Brackets show 95\% paired-bootstrap CIs over instances.}
\end{table}

Both contrasts decrease accuracy for AKOrN on both tasks, TRM on Sudoku, and ItrSA++ on both tasks. These differences survive Holm correction within their prespecified families: eight comparisons for AKOrN/TRM and four for ItrSA++. The two TRM on Maze differences remain unresolved. These are the original separate families, not a post hoc combined twelve-comparison test. For ItrSA++, among updates with a defined reference change of absolute size greater than $10^{-6}$, the median local ratio of the absolute inter-class similarity change after removal to that with the original force was 0.0004 on Sudoku and 0.0089 on Maze; restoring the magnitude gave ratios of 0.0072 and 0.0205. These diagnostics evaluate a local update map at the states visited during inference and support suppression of the intended change. They do not assert bitwise agreement with the native update, whose floating-point operation order differs.

Together with the positive RoFB gains under this protocol, the component-removal results support a contribution of separation-related directions on AKOrN on Sudoku, AKOrN on Maze, TRM on Sudoku, and ItrSA++ on Maze. ItrSA++ on Sudoku requires a different interpretation: RoFB's gain over the no-force reference was only 0.45 pp $[-0.45, 1.35]$, so the removal-induced decrease does not establish that the component was necessary for an existing gain. TRM on Maze likewise had no RoFB accuracy gain. The experiment tests one local gradient direction, not whether inter-class separation is the sole mediator of improvement.

\subsection{Comparison to other class separation interventions}

We compare standard RoFB with three readout-based alternatives for introducing repulsion: margin repulsion, a probability-weighted contrastive loss, and a loss combining within-class variance with a between-centroid margin. We additionally consider a K-means centroid-attraction direction that does not use readout predictions. None uses reference labels to construct its force. All four alternatives retain standard RoFB's online entropy gate, warmup, and model-specific feedback adapter. Each condition computes the gate from its own current readouts.

At each update, we construct a candidate direction $\mathbf u_t$ from the current states and, where used, readouts. For the definitions below, we omit $t$, retain $\hat{\mathbf z}^{(i)}$ for the unit-normalized state, and write $\mathbf p^{(i)}$ for the readout probabilities and $\hat\ell_i=\arg\max_c p_c^{(i)}$ for the predicted class. All cell indices lie in $\mathcal B$. Let $s_{ij}=\hat{\mathbf z}^{(i)\top}\hat{\mathbf z}^{(j)}$, $q_{ij}=\mathbf p^{(i)\top}\mathbf p^{(j)}$, and $r_{ij}=\|\hat{\mathbf z}^{(i)}-\hat{\mathbf z}^{(j)}\|$. For each loss $\mathcal L$ below, the candidate direction is $\mathbf u^{(i)}=-\nabla_{\mathbf z^{(i)}}\mathcal L$: we differentiate through state normalization while holding predictions and discrete assignments fixed. K-means instead uses the stated displacement directly.

\paragraph{Margin repulsion.} We penalize pairs that are closer than a margin $m>0$, weighted by their predicted dissimilarity:
\begin{equation}
\mathcal L_{\mathrm{rep}}
=\frac{1}{\binom{|\mathcal B|}{2}}
\sum_{\substack{i,j\in\mathcal B\\i<j}}
(1-q_{ij})[m-r_{ij}]_+^2,\qquad [x]_+=\max(x,0).
\end{equation}
The resulting force separates nearby cells with dissimilar predictions; pairs already beyond the margin contribute no force.

\paragraph{Probability-weighted contrastive loss.} We use agreement between the readout distributions as positive-pair weights:
\begin{equation}
\mathcal L_{\mathrm{prob}}
=-\frac{1}{|\mathcal I_q|}\sum_{i\in\mathcal I_q}
\sum_{\substack{j\in\mathcal B\\j\ne i}}
\frac{q_{ij}}{\sum_{k\in\mathcal B\setminus\{i\}}q_{ik}}
\log\frac{\exp(s_{ij}/\kappa)}{\sum_{k\in\mathcal B\setminus\{i\}}\exp(s_{ik}/\kappa)}.
\end{equation}
Here $\kappa>0$ is the temperature and $\mathcal I_q=\{i\in\mathcal B:\sum_{j\ne i}q_{ij}>0\}$ contains cells with a positive-weight partner. Minimizing this loss favors latent similarity to cells with similar predictions relative to the other cells; self-pairs are excluded. Uncertain predictions contribute continuously through their probability distributions.

\paragraph{Within-class variance plus centroid margin.} Let $\widehat{\mathcal B}_c=\{i\in\mathcal B:\hat\ell_i=c\}$ and $\boldsymbol\mu_c=|\widehat{\mathcal B}_c|^{-1}\sum_{i\in\widehat{\mathcal B}_c}\hat{\mathbf z}^{(i)}$ be the cells and centroid of a nonempty predicted class. We use
\begin{equation}
\mathcal L_{\mathrm{var+margin}}
=\sum_{c:\,|\widehat{\mathcal B}_c|>0}\ \sum_{i\in\widehat{\mathcal B}_c}
\|\hat{\mathbf z}^{(i)}-\boldsymbol\mu_c\|^2
+\sum_{\substack{c<c'\\|\widehat{\mathcal B}_c|,\,|\widehat{\mathcal B}_{c'}|>0}}
[m-\|\boldsymbol\mu_c-\boldsymbol\mu_{c'}\|]_+^2.
\end{equation}
The first term contracts each predicted class and the second separates nearby class centers. Both terms are sums with equal coefficients; their gradient includes the dependence of the centroids on the states.

\paragraph{K-means centroid attraction.} To test a clustering direction that does not use readout predictions, we apply K-means to the normalized states, with $G$ clusters and returned centers $\boldsymbol\mu_1,\ldots,\boldsymbol\mu_{G}$, and use
\begin{equation}
a_i^{\mathrm{km}}=\arg\min_{1\le k\le G}\|\hat{\mathbf z}^{(i)}-\boldsymbol\mu_k\|^2,
\qquad
\mathbf u^{(i)}=\boldsymbol\mu_{a_i^{\mathrm{km}}}-\hat{\mathbf z}^{(i)}.
\end{equation}
The direction moves each cell toward its assigned center without using readout predictions; its entropy gate still uses the current readout. Clustering is recomputed at each update with a fixed initialization seed and at most 20 Lloyd iterations; $G$ is capped by the number of mutable cells.

All directions are recomputed online from the current state. Empty comparisons or numerically negligible denominators contribute zero force. To make the strength selection insensitive to a constant rescaling of a loss, we normalize each alternative direction before applying the gate. Let $\mathbf z_{t,\mathcal B}$ contain the mutable-cell states, $\lambda$ be standard RoFB's fixed feedback strength, and $\beta$ be an additional, validation-selected multiplier for the alternative direction. We use
\begin{equation}
\widetilde{\mathbf u}_t=\frac{\|\mathbf z_{t,\mathcal B}\|_F}{\|\mathbf u_t\|_F}\mathbf u_t,
\qquad
\mathbf w_t^{\mathrm{alt}}=\lambda\beta g_t\widetilde{\mathbf u}_t,
\end{equation}
with zero force when $\|\mathbf u_t\|_F\le10^{-10}$. The norm is over mutable cells and features within each candidate. The multiplier $\beta$ scales the external force outside the sigmoid; it does not change the gate's entropy threshold or temperature. Standard RoFB remains an unchanged anchor and does not use this alternative-direction normalization or the additional multiplier. Thus $\beta=1$ does not imply equal force norms between an alternative and standard RoFB.

Using $\widetilde{\mathbf u}_t$ in the one-update map defined above, one inference step can be written as
\begin{equation}
\mathbf z_{t+1}=\Pi_{\mathcal M}\!\left(R(\mathbf z_t)+\lambda\beta g_t L_t(\widetilde{\mathbf u}_t)\right).
\end{equation}
Here $R(\mathbf z_t)$ is the ordinary update before feedback and final normalization, $L_t$ is the model-specific linear tangent projection including the native step size for AKOrN, and $\Pi_{\mathcal M}$ is the final state normalization. The force is inserted into AKOrN's oscillator update, ItrSA++'s self-attention update, or TRM's low-level update. These operations and their implementation order follow the standard RoFB adapter. The force is not rescaled to the natural update or renormalized after gating.

For each model--task pair, all four standard RoFB hyperparameters ($t_{\min}$, $\alpha$, $\tau$, and $\lambda$) are fixed to their saved values and are not retuned for any alternative. The tangent projection, feedback location, normalization, checkpoint, inference horizon, and number of candidates are also held fixed within that pair. Fixing the gate parameters does not fix the gate values: each condition computes $g_t$ from its own evolving readouts.

Only the additional force multiplier $\beta\in\{0.01,0.1,1\}$ and one method-specific parameter are selected for each alternative and each model--task pair. Their Cartesian product gives nine validation configurations per method:

\begin{table}[!bp]
\centering\small
\caption{Validation search: three force multipliers crossed with three values of the method-specific parameter. $G$ counts state clusters, not inference candidates. Candidate counts remain four for AKOrN and ItrSA++ and one for TRM.}
\label{tab:alternative_intervention_search}
\setlength{\tabcolsep}{5pt}
\renewcommand{\arraystretch}{1.15}
\begin{tabular}{p{0.30\linewidth}p{0.62\linewidth}}
\toprule
\textbf{Alternative} & \textbf{Method-specific parameter searched jointly with $\beta$} \\
\midrule
Margin repulsion & Distance margin $m\in\{0.5,1,1.5\}$ \\
Probability-weighted contrastive loss & Contrastive temperature $\kappa\in\{0.05,0.1,0.2\}$; separate from the fixed gate temperature $\tau$ \\
Within-class variance plus centroid margin & Centroid-distance margin $m\in\{0.5,1,1.5\}$; both loss coefficients remain one \\
K-means centroid attraction & Cluster count $G\in\{8,9,10\}$ on Sudoku or $\{1,2,3\}$ on Maze \\
\bottomrule
\end{tabular}
\end{table}

We use the same 500 validation boards per task across all models. For each of the four alternatives in each model--task pair, we select the highest exact accuracy, breaking ties by fewer newly incorrect boards relative to no feedback, smaller $\beta$, and then configuration order. This selects 24 configurations in total, each locked before test evaluation. Standard RoFB uses its saved configuration without a new hyperparameter search in this experiment; the equal nine-configuration budget applies to the alternatives, not to the historical tuning of RoFB.

We evaluate the same previously used 2,000 Sudoku and 900 Maze test boards across all three models. AKOrN and ItrSA++ use four candidates with minimum-entropy selection, and TRM uses one. All conditions use the standard horizon: 256 AKOrN updates, 64 ItrSA++ blocks (256 self-attention updates), or 16 TRM outer iterations. Checkpoints, initialization, and the inference seed are fixed within each model.

\begin{figure}[!t]
\centering
\includegraphics[width=\linewidth]{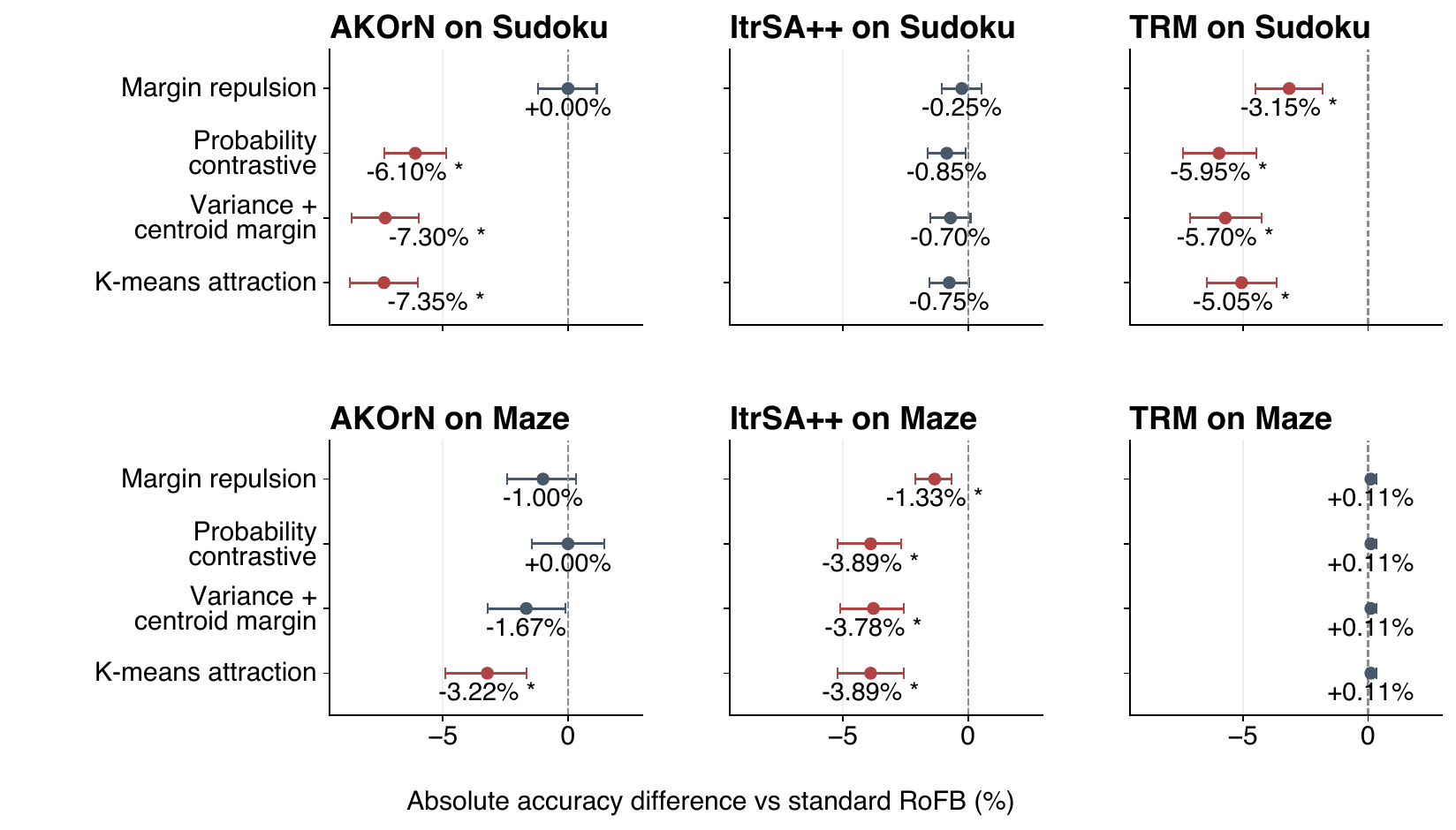}
\caption{Absolute accuracy differences from standard RoFB (\%; paired-bootstrap 95\% confidence intervals). Columns: AKOrN, ItrSA++, TRM; rows: Sudoku, Maze. The dashed line is standard RoFB with its native entropy gate and saved strength. The alternatives retain that gate and adapter and use the validation-selected pre-gate direction scaling above. Intervals use 10,000 paired bootstrap resamples and are unadjusted; stars indicate exact McNemar $p<0.05$ after Holm correction across the prespecified 24 comparisons (four alternatives across six model--task pairs).}
\label{fig:alternative_separation_interventions}
\end{figure}

None of the four alternatives had a supported accuracy improvement over standard RoFB after correction across the 24 comparisons. Lower accuracy was supported in 12 comparisons; the remaining 12 comparisons did not resolve a difference. Unadjusted intervals and corrected p-values answer different uncertainty questions, and a non-significant result is not evidence of equivalence.

Relative to the no-feedback reference in the same runs, several alternatives also improved accuracy. These are descriptive comparisons outside the 24-comparison family, with unadjusted 95\% paired-bootstrap CIs above zero: margin repulsion on AKOrN on Sudoku (+7.55\% [6.15, 9.05]), TRM on Sudoku (+2.30\% [1.20, 3.40]), AKOrN on Maze (+2.22\% [0.89, 3.56]), and ItrSA++ on Maze (+2.44\% [1.33, 3.56]); the probability-weighted contrastive loss on AKOrN on Sudoku (+1.45\% [0.65, 2.30]) and AKOrN on Maze (+3.22\% [1.89, 4.67]); and within-class variance plus centroid margin on AKOrN on Maze (+1.56\% [0.67, 2.56]). K-means centroid attraction did not improve on the reference in any pair. In the same runs, standard RoFB changed accuracy by +7.55, +0.95, and +5.45\% for AKOrN, ItrSA++, and TRM on Sudoku, and by +3.22, +3.78, and $-0.11$\% on Maze.

On AKOrN on Sudoku, margin repulsion reached 92.20\% compared with 92.20\% for standard RoFB, rescuing 70 boards while newly failing on 70 others; its paired difference was +0.00\% [-1.20, 1.15]. Thus equal point accuracy does not imply identical outcomes or establish equivalence. All four alternatives had supported accuracy losses on TRM on Sudoku and ItrSA++ on Maze.

All paired analyses use the reevaluated reference runs from this experiment. The reported intervals do not include variation between repeated executions, and the earlier saved reference outcomes were not bit-identical in every model--task condition.

These results apply to the tested checkpoints, tasks, and finite search. The four interventions shared the gate and validation budget, but their parameterizations and pre-gate force scales differed from the unchanged RoFB anchor. The selected multiplier was at a search boundary in 12 of 24 method--task conditions. The experiment therefore does not establish the best attainable performance of each direction, the general superiority of RoFB, or compute-matched efficiency. Evaluation on unused boards and across inference and training seeds would be needed to establish broader robustness.

\FloatBarrier
\subsection{Why Inter-class Separation Induces Solution}
Why does inter-class separation induce a solution? We speculate that the learned recurrent dynamics already contain task-specific information that supports solution finding, and that RoFB can help these dynamics by reinforcing separation between cells representing different classes. RoFB leaves the learned update rule and its parameters unchanged while adding feedback to the evolving state. In this view, the feedback facilitates the use of structure acquired during training, rather than supplying a solution independently of the learned dynamics.

\section{Mechanistic Analysis of RoFB (2): Ablation Study on RoFB}
\label{subsec:ablation_rofb}
We next performed an ablation study to understand the contribution of different components of RoFB intervention: connectivity $J$, intervention direction $\sum_{j} \tilde{J}^{(ij)}\mathbf{z}^{(j)}$, and gate $g$, all of which contribute to the overall performance gain of RoFB in both Sudoku and Maze tasks.

\paragraph{Connectivity: $J$.}
To check the effect of coupling, we examined the dependency of the coupling $J^{(ij)}=h(d(\mathbf{p}^{(i)}, \mathbf{p}^{(j)}))$ as follows: (i) a purely repulsive coupling $h(d)=-d$ (the original RoFB), (ii) a purely attractive coupling $h(d)=1-d$, where tokens of close readout probability vectors attract each other, (iii) a mixed coupling $h(d)=1-2d$, (iv) the constant repulsion $h(d)=-1$. Within each evaluation, we used the same blank-cell mask, feedback strength, and gate. As for row normalization, we kept it the same as in the original RoFB setup in (i), (ii), and (iv), and used a different way of normalization $\tilde J_{ij}=\frac{J_{ij}}{N_{\text{blank}}}$ in (iii) because the original implementation suppresses rows containing mixed signs. Table~\ref{tab:ablation_coupling} compares the four kernels across all six model--task pairs at the standard inference horizon. Because Mixed uses a different normalization, this is not an ablation of the coupling function alone.
Constant repulsion removes prediction dependence from the pairwise coupling, but still uses predictions in the entropy gate.

The standard horizon is 256 Kuramoto updates for AKOrN, 256 self-attention updates (64 outer blocks) for ItrSA++, and 16 outer iterations for TRM. AKOrN and TRM on Sudoku use the earlier 20{,}000-instance evaluations, and ItrSA++ on Maze uses the earlier 1{,}000-instance evaluation; the other three pairs use 500 test instances each. Within each model--task pair, all four conditions share the checkpoint, instances, initialization, gate, and feedback strength, with four candidates for AKOrN and ItrSA++ and one for TRM. Mixed uses the original strength ($1\times$) and $J_{ij}/N_{\mathrm{blank}}$; the other kernels use absolute-row-sum normalization. Maze uses valid paths of reference-matched length. ItrSA++ uses tangent-projected feedback.

\begin{table}[!bp]
\centering \small
\caption{\textbf{Coupling ablations at the standard horizon.}}
\label{tab:ablation_coupling}
\setlength{\tabcolsep}{4pt}
\begin{tabular}{lrrr}
\toprule
\textbf{Intervention} & \textbf{AKOrN} & \textbf{ItrSA++} & \textbf{TRM} \\
\midrule
\multicolumn{4}{l}{\textit{Sudoku}} \\
Original RoFB ($h(d)=-d$) & \estimateci{91.52}{91.12}{91.89} & \estimateci{77.00}{73.11}{80.47} & \estimateci{74.17}{73.55}{74.77} \\[4pt]
Attractive ($h(d)=1-d$) & \estimateci{72.35}{71.72}{72.96} & \estimateci{75.40}{71.44}{78.97} & \estimateci{56.88}{56.19}{57.56} \\[4pt]
Mixed ($h(d)=1-2d$) & \estimateci{90.82}{90.41}{91.21} & \estimateci{77.20}{73.32}{80.66} & \estimateci{73.33}{72.71}{73.94} \\[4pt]
Const ($h(d)=-1$) & \estimateci{91.62}{91.23}{92.00} & \estimateci{77.00}{73.11}{80.47} & \estimateci{74.72}{74.11}{75.32} \\[4pt]
\midrule
\multicolumn{4}{l}{\textit{Maze}} \\
Original RoFB ($h(d)=-d$) & \estimateci{78.80}{75.00}{82.16} & \estimateci{85.30}{82.97}{87.36} & \estimateci{92.80}{90.19}{94.75} \\[4pt]
Attractive ($h(d)=1-d$) & \estimateci{75.40}{71.44}{78.97} & \estimateci{70.60}{67.70}{73.34} & \estimateci{93.00}{90.42}{94.92} \\[4pt]
Mixed ($h(d)=1-2d$) & \estimateci{79.60}{75.85}{82.90} & \estimateci{84.40}{82.02}{86.52} & \estimateci{92.80}{90.19}{94.75} \\[4pt]
Const ($h(d)=-1$) & \estimateci{73.80}{69.77}{77.46} & \estimateci{84.30}{81.91}{86.42} & \estimateci{93.00}{90.42}{94.92} \\[4pt]
\bottomrule
\end{tabular}
\resulttabnote{Values are candidate-selected accuracies in \%; brackets show 95\% Wilson score CIs over instances for each accuracy, not for differences between conditions. Intervals are unadjusted and conditional on the evaluated checkpoint and inference configuration.}
\end{table}

In separate evaluations at four times the standard horizon, using 500 shared instances per task with four candidates for AKOrN and ItrSA++ and one for TRM, constant repulsion retained much of the observed benefit on AKOrN and TRM Sudoku, with no resolved difference from readout-dependent coupling. On Maze, however, trajectories revealed a difference in retaining correct solutions. For AKOrN, readout-dependent and constant coupling reached a correct solution in 1,581 and 1,590 of 2,000 trajectories, respectively, but 49 versus 181 of these trajectories were incorrect at the end. For ItrSA++, the corresponding numbers were 1,688 versus 1,694 reaching a solution and 3 versus 34 ending incorrectly after doing so. These are descriptive comparisons of each condition's trajectories, not continuations branched from identical already-correct states.

The coupling provides one possible explanation. If all readouts are uniform over $C$ classes, then $1-\mathbf p^{(i)\top}\mathbf p^{(j)}=1-1/C$, which becomes the constant coupling after row normalization, up to the implementation's stabilizing term. In contrast, for confident predictions the coupling approaches zero between cells assigned the same class and remains repulsive between different predicted classes. It can therefore suppress direct repulsion within an emerging class without removing repulsion between classes.

This distinction may matter more when many cell pairs share a predicted class. At the standard horizon, the observed same-predicted-class pair fraction was approximately 0.10 for Sudoku and 0.68 for Maze. The cosine between the readout-dependent and constant force vectors before tangent projection and gating was 0.997–0.999 on Sudoku, compared with 0.353, 0.867, and 0.209 for AKOrN, ItrSA++, and TRM on Maze. These diagnostics motivate a relationship between class composition and the effect of the coupling, but do not establish that class imbalance causes the accuracy difference. TRM–Maze, in particular, had a large force-direction difference without an accuracy gain. Suppression of same-class pair weights also does not imply that the total feedback force vanishes at a correct solution.

\FloatBarrier
\paragraph{Intervention direction: $\sum_{j} \tilde{J}^{(ij)}\mathbf{z}^{(j)}$.}
To see the effect of the intervention direction, we compare it to norm-matched random perturbations, where $\sum_{j\ne i}\tilde J_t^{(ij)}\mathbf z_t^{(j)}$ in the RoFB update is replaced by a Gaussian random vector. The random force is restricted to blank cells and matched to the RoFB force after tangent projection and linear step scaling, using one norm over all blank cells of each candidate instance. It uses the same gate rule and start time. Matching is performed at each condition's own state, so the forces need not have identical magnitudes across their diverging trajectories.

In the same 500-instance evaluations at four times the standard horizon, RoFB exceeded the mean accuracy of three random-direction seeds by 8.33 percentage points [5.87, 10.93] on AKOrN–Sudoku, 11.13 [8.13, 14.13] on TRM–Sudoku, and 4.07 [2.47, 5.80] on ItrSA++–Maze. All three differences remained significant after Holm correction across the eighteen primary contrasts. The other three pairs did not show a resolved advantage after correction. These results support a contribution of the structured direction in the three identified settings, rather than a benefit from arbitrary perturbations of the same magnitude. They do not identify a particular unstable dynamical subspace.

\paragraph{Gate $g$.}
The gate function $g$, Eq.~\eqref{eq:rofb_gate}, consists of (i) a warmup gate $1[t \geqslant t_{\min}]$, which lets the dynamics evolve freely for some steps before applying the feedback, and (ii) an entropy gate, which modulates the strength of the feedback based on the confidence of the current predictions. To see the effect of each component, we ablate them individually by setting each gate to $\equiv 1$. 

We report candidate-selected accuracy at the standard inference horizon on 500 shared test instances per task, with matched initial states across gate conditions ($K_{\text{vote}}{=}4$ for AKOrN and ItrSA++, one candidate for TRM). Each ablated factor of the gate $g$ in Eq.~\eqref{eq:rofb_gate} is replaced exactly by one; the feedback strength and $t_{\min}$ are kept at their tuned values without retuning. We compute 95\% CIs for accuracy differences using 10,000 paired bootstrap resamples of instances. Significance is assessed with two-sided exact McNemar tests and Holm correction across the twelve prespecified single-factor contrasts; joint ablations are exploratory.

\begin{table}[!bp]
\centering \small
\caption{\textbf{Gate-factor ablations.}
}
\label{tab:ablation_gates}
\setlength{\tabcolsep}{6pt}
\begin{tabular}{lrrr}
\toprule
\textbf{Intervention} & \textbf{AKOrN} & \textbf{ItrSA++} & \textbf{TRM} \\
\midrule
\multicolumn{4}{l}{\textit{(a) Candidate-selected accuracy (\%)}} \\[2pt]
\multicolumn{4}{l}{\textit{Sudoku}} \\
Original RoFB & 93.2 & 77.0 & 71.0 \\
Warmup $\equiv 1$ & 93.8 & 78.0 & 72.0 \\
Entropy gate $\equiv 1$ & 89.0$^{\dagger}$ & 77.0 & 69.4 \\
Both $\equiv 1$ & 82.0 & 76.2 & 63.0 \\
\midrule
\multicolumn{4}{l}{\textit{Maze}} \\
Original RoFB & 78.8 & 85.0 & 92.8 \\
Warmup $\equiv 1$ & 63.8$^{\dagger}$ & 82.2 & 92.8 \\
Entropy gate $\equiv 1$ & 75.0$^{\dagger}$ & 84.8 & 92.8 \\
Both $\equiv 1$ & 30.0 & 77.4 & 93.4 \\
\midrule
\multicolumn{4}{l}{\textit{(b) Difference from original RoFB (pp)}} \\[2pt]
\multicolumn{4}{l}{\textit{Sudoku}} \\[2pt]
Warmup $\equiv 1$ & \estimateci{+0.6}{-1.6}{2.8} & \estimateci{+1.0}{-1.2}{3.4} & \estimateci{+1.0}{-2.2}{4.2} \\[4pt]
Entropy gate $\equiv 1$ & \estimateci{-4.2}{-6.8}{-1.6} & \estimateci{0.0}{-1.0}{1.0} & \estimateci{-1.6}{-4.6}{1.2} \\[4pt]
Both $\equiv 1$ & \estimateci{-11.2}{-14.8}{-7.8} & \estimateci{-0.8}{-3.0}{1.4} & \estimateci{-8.0}{-11.8}{-4.2} \\[4pt]
\midrule
\multicolumn{4}{l}{\textit{Maze}} \\[2pt]
Warmup $\equiv 1$ & \estimateci{-15.0}{-18.6}{-11.4} & \estimateci{-2.8}{-5.6}{0.0} & \estimateci{0.0}{-1.0}{1.0} \\[4pt]
Entropy gate $\equiv 1$ & \estimateci{-3.8}{-6.0}{-1.8} & \estimateci{-0.2}{-1.2}{0.8} & \estimateci{0.0}{0.0}{0.0} \\[4pt]
Both $\equiv 1$ & \estimateci{-48.8}{-53.4}{-44.2} & \estimateci{-7.6}{-11.0}{-4.2} & \estimateci{+0.6}{-0.6}{2.0} \\[4pt]
\bottomrule
\end{tabular}
\resulttabnote{Panel (a) reports accuracy in \%; panel (b) reports differences from original RoFB in percentage points (pp), with unadjusted 95\% paired-bootstrap CIs in brackets. $^{\dagger}$ in panel (a) marks a decrease supported by exact McNemar tests with Holm correction across the twelve prespecified single-factor contrasts. Joint ablations are exploratory and excluded from this correction.}
\end{table}

The results show that removing the warmup gate reduced AKOrN–Maze accuracy significantly by 15.0 percentage points, while removing the entropy gate reduced AKOrN–Sudoku and AKOrN–Maze accuracy by 4.2 and 3.8 percentage points, respectively. 
The remaining single-factor effects were inconclusive. Removing both gates reduced AKOrN–Maze accuracy from 78.8\% to 30.0\%, illustrating that the combined ablation can be much more disruptive than either single ablation.

On AKOrN–Maze, removing warmup reduced the number of trajectories that ever reached a correct solution from 1,581 to 1,097 out of 2,000. Removing the entropy gate instead left the arrival count close to the original value, at 1,574, but increased the number that subsequently ended incorrectly from 47 to 98. This is consistent with different effects on reaching and retaining a solution. 

\FloatBarrier
\section{Mechanistic Analysis of RoFB (3): RoFB and Confidence}
\label{app:rofb_confidence}
\paragraph{RoFB increases confidence.}

To examine whether RoFB increases predictive confidence, we measure the entropy of the intermediate readout distributions. For the set of mutable cells $\mathcal B$ and the readout vocabulary size $C$, we define the normalized mean entropy
\begin{equation}
\bar H_t=\frac{1}{|\mathcal B|\log C}\sum_{i\in\mathcal B}H(\mathbf p_t^{(i)}),\qquad
H(\mathbf p)=-\sum_{c=1}^{C}p_c\log p_c.
\label{eq:rofb_confidence_1}
\end{equation}
Lower entropy corresponds to a more concentrated prediction, but does not imply a more accurate one. We compare online RoFB with the no-force reference through the same intervention path, using paired instances and initial states. The primary summary averages over all fixed candidates within each instance before averaging across instances, avoiding selection of the most confident candidate for the confidence comparison itself. With $N$ instances and $K$ candidates per instance, the reported paired mean difference is
\begin{equation}
\widehat{\Delta H}_t=
\frac{1}{N}\sum_{b=1}^{N}\frac{1}{K}\sum_{k=1}^{K}
\left(\bar H_{b,k,t}^{\mathrm{RoFB}}-\bar H_{b,k,t}^{\mathrm{zero}}\right),
\label{eq:rofb_confidence_2}
\end{equation}
where $b$ indexes instances, $k$ indexes matched candidates, and ``zero'' denotes the no-force reference. Bootstrap resampling draws entire instances, keeping their candidates together.

\begin{table}[!bp]
\centering\small
\caption{\textbf{Mean normalized readout entropy with and without RoFB.}}
\label{tab:rofb_confidence_entropy}
\setlength{\tabcolsep}{3pt}
\begin{tabular}{@{}lrrr@{}}
\toprule
\textbf{Condition / difference} & \textbf{AKOrN} & \textbf{ItrSA++} & \textbf{TRM} \\
\midrule
\multicolumn{4}{l}{\textit{Sudoku}} \\[2pt]
No-force reference & $0.0982$ & $0.118$ & $0.00825$ \\[4pt]
RoFB & $0.0774$ & $0.108$ & $0.00591$ \\[4pt]
RoFB $-$ reference & \estimateci{-0.0209}{-0.0318}{-0.00998} & \estimateci{-0.00908}{-0.0135}{-0.00467} & \estimateci{-0.00234}{-0.00329}{-0.00142} \\[4pt]
\midrule
\multicolumn{4}{l}{\textit{Maze}} \\[2pt]
No-force reference & $0.00293$ & $0.00183$ & $1.48\times10^{-6}$ \\[4pt]
RoFB & $0.00114$ & $2.21\times10^{-4}$ & $3.23\times10^{-7}$ \\[4pt]
RoFB $-$ reference & \estimateci{-0.00179}{-0.00211}{-0.00150} & \estimateci{-0.00161}{-0.00214}{-0.00115} & \estimateci{-1.16\times10^{-6}}{-2.99\times10^{-6}}{9.13\times10^{-26}} \\[4pt]
\bottomrule
\end{tabular}
\resulttabnote{Values are dimensionless and rounded to three significant digits; differences are computed before rounding. Brackets on the RoFB-minus-reference rows show 95\% paired-bootstrap CIs over instances, with matched candidates kept together.}
\end{table}

Each pair uses 500 instances at the standard inference horizon, with four candidates for AKOrN and ItrSA++ and one for TRM. The CIs use 20,000 paired bootstrap resamples of instances. Normalization uses the actual head vocabulary size, not the number of semantic classes in the geometry analysis. Five differences are supported by sign-flip tests with Holm correction across the six primary comparisons; TRM on Maze is unresolved ($p=0.494$). Its CI includes zero despite an upper bound very close to numerical zero.

RoFB reduced mean entropy in five of the six pairs, supporting increased predictive confidence in those settings. TRM on Maze already had very low entropy without feedback and showed no resolved change. The same five pairs had negative mean-entropy differences at four times the standard horizon, a secondary endpoint. The effect was not uniform across confidence summaries: on ItrSA++ on Sudoku, the maximum-cell entropy difference at the standard horizon was \ensuremath{-}0.00082 [\ensuremath{-}0.00810, 0.00653], despite the decrease in mean entropy. Moreover, confidence need not track accuracy: on the 500-instance ItrSA++ on Sudoku cohort, candidate-selected accuracy was 77.0\% under both RoFB and the no-force reference. Among ItrSA++ on Sudoku candidates that were incorrect under both conditions, mean entropy decreased by 0.03397 [0.02891, 0.03906] (568 paired candidates from 178 instances). This is a descriptive post-intervention stratum, with uncertainty resampled by instance. These results establish a change in predictive concentration, not improved calibration or guaranteed correctness.

\paragraph{Confidence-increasing intervention.}

To test whether increasing confidence alone can reproduce the gains of RoFB, we replace its relational feedback direction with the negative gradient of mean predictive entropy. The intervention differentiates the readout entropy with respect to the feedback target while keeping model weights fixed, applies the model-specific tangent projection, and uses the same feedback location and online gate rule. In TRM, a target-matched update connects the high-level readout to the low-level feedback target. We compare a fixed-scale version with a version whose projected magnitude is matched to RoFB separately for each mutable cell at each update. The gate rule is shared, but its values evolve along each intervention's own trajectory. Let $\mathcal H_t$ denote the mean raw readout entropy as a function of the feedback target, with other model variables fixed:
\begin{equation}
\mathcal H_t(\mathbf z)=\frac{1}{|\mathcal B|}\sum_{i\in\mathcal B}H\!\left(\mathbf p_t^{(i)}(\mathbf z)\right),\qquad
\mathbf u_t^{H}=-\left.\nabla_{\mathbf z}\mathcal H_t(\mathbf z)\right|_{\mathbf z=\mathbf z_t}.
\label{eq:rofb_confidence_3}
\end{equation}
For TRM, $\mathbf p_t(\mathbf z)$ includes the target-matched high-level update needed to read out from a perturbed low-level state. Unlike the displayed confidence metric $\bar H_t$, $\mathcal H_t$ is not divided by $\log C$. The fixed-scale force is $\lambda g_t\mathbf u_t^H$. For the magnitude-matched version, each mutable cell is scaled separately:
\begin{equation}
\mathbf w_t^{H,(i)}=
\frac{\|[L_t(\mathbf v_t)]^{(i)}\|_2}
{\|[L_t(\mathbf u_t^H)]^{(i)}\|_2}\,
\mathbf u_t^{H,(i)},\qquad i\in\mathcal B.
\label{eq:rofb_confidence_4}
\end{equation}
Thus matching uses the current RoFB force including its gate and strength, before the final nonlinear normalization. Ratios are applied only where numerically defined; zero or unconstructible directions are handled by the recorded zero-force rule.

\begin{table}[!bp]
\centering\small
\caption{\textbf{Accuracy gains over paired no-feedback baselines.}}
\label{tab:rofb_confidence_interventions}
\setlength{\tabcolsep}{3pt}
\begin{tabular}{@{}lrrr@{}}
\toprule
\textbf{Intervention} & \textbf{AKOrN} & \textbf{ItrSA++} & \textbf{TRM} \\
\midrule
\multicolumn{4}{l}{\textit{Sudoku}} \\[2pt]
\shortstack[l]{Fixed-scale entropy-gradient \\ intervention} & \estimateci{+0.700}{0.000}{1.400} & \estimateci{+0.050}{-0.650}{0.750} & \estimateci{+0.450}{-0.350}{1.250} \\[4pt]
\shortstack[l]{Magnitude-matched \\ entropy-gradient intervention} & \estimateci{-0.100}{-1.250}{1.100} & \estimateci{-1.200}{-2.150}{-0.250} & \estimateci{-1.200}{-2.400}{-0.050} \\[4pt]
Online RoFB & \estimateci{+7.150}{5.800}{8.500} & \estimateci{+0.800}{-0.050}{1.650} & \estimateci{+5.600}{4.250}{7.000} \\[4pt]
\midrule
\multicolumn{4}{l}{\textit{Maze}} \\[2pt]
\shortstack[l]{Fixed-scale entropy-gradient \\ intervention} & \estimateci{-0.111}{-0.333}{0.000} & \estimateci{-0.111}{-0.333}{0.000} & \estimateci{+0.000}{0.000}{0.000} \\[4pt]
\shortstack[l]{Magnitude-matched \\ entropy-gradient intervention} & \estimateci{+0.111}{-1.111}{1.333} & \estimateci{-24.778}{-27.778}{-21.778} & \estimateci{+0.000}{0.000}{0.000} \\[4pt]
Online RoFB & \estimateci{+3.222}{1.667}{4.778} & \estimateci{+3.889}{2.667}{5.222} & \estimateci{-0.111}{-0.333}{0.000} \\[4pt]
\bottomrule
\end{tabular}
\resulttabnote{Values are accuracy differences from the paired no-feedback baseline in percentage points (pp). Brackets show 95\% paired-bootstrap CIs over instances; these intervals are not adjusted for multiple comparisons.}
\end{table}

All three models use the same 2,000 Sudoku instances and 900 Maze instances. AKOrN and ItrSA++ use four candidates and TRM one, at their standard horizons. Within each model--task pair, all conditions share the instances and inference seed. The 95\% CIs use 20,000 paired instance-bootstrap resamples, without multiple-comparison correction; they describe uncertainty over instances for the fixed checkpoint and inference conditions, not across training seeds. AKOrN/TRM lack immediate entropy-change checks, so these comparisons test gradient-direction replacement rather than verified confidence increases.

For ItrSA++ on Sudoku, the magnitude-matched replacement passed the entropy-reduction checks but achieved 76.70\% accuracy versus 78.70\% for RoFB (\ensuremath{-}2.00 pp [\ensuremath{-}3.00, \ensuremath{-}1.00], a paired comparison supported after correction across the four primary entropy-intervention comparisons). Thus, this verified confidence-increasing replacement did not reproduce RoFB's performance. On ItrSA++ on Maze, the entropy-reduction checks failed, and the AKOrN/TRM comparisons lack the corresponding checks; these comparisons do not establish the insufficiency of increasing confidence. None of these results establishes that confidence changes are irrelevant to RoFB or identifies their causal mediation of its gains.

\FloatBarrier

\newpage

\section{Natural Frequencies and the Original AKOrN Architecture}
\label{app:akorn_natural_frequencies}
RoFB does not require removing AKOrN's natural-frequency term. We evaluated models trained with and without this term in both our modified architecture (Eq.~\eqref{eq:modified_akorn}) and a reproduction of the original Sudoku architecture~\citep{Miyato2024-zl}. The latter retains the original four-dimensional oscillators, geometric-transform attention, and rotation-invariant readout. RoFB was applied only at inference, with three independently trained checkpoints per condition.

For the modified architecture, we retained the four-channel spherical grouping and evaluated one fixed RoFB configuration on 20,000 Sudoku Extreme test boards at 256 recurrent steps. For the original architecture, we trained on 9,000 SATNet boards, selected one shared RoFB configuration across the two natural-frequency conditions on 1,000 RRN validation boards, and evaluated on 18,000 disjoint RRN test boards at 128 steps. Both experiments used one trajectory per board and compared RoFB with the corresponding baseline at the same recurrent horizon. Training recipes and evaluation sets differ across architectures; the relevant comparison is therefore within each architecture.

With natural frequencies enabled, all three gains were positive in both architectures. Without them, gains were mixed in the modified architecture, while the original-architecture reproduction showed no improvement: all three within-checkpoint paired 95\% intervals included zero. Thus, our results do not support attributing RoFB's benefit to the removal of natural frequencies, but they also do not establish robustness across all training conditions. The improvement on the original architecture required validation-based adjustment; directly transferring the modified model's configuration reduced accuracy by 31.7--48.2 percentage points on average across training seeds. These results are limited to Sudoku and do not establish the same behavior on Maze.

\begin{table}[!hb]
\centering
\caption{Effect of natural frequencies on RoFB gains in Sudoku.}
\label{tab:akorn_natural_frequencies}
\footnotesize
\setlength{\tabcolsep}{3pt}
\begin{tabular*}{\linewidth}{@{\extracolsep{\fill}}lclcc@{}}
\toprule
Architecture & \shortstack{Natural\\frequencies} & Gain for each training seed & Mean $\pm$ SD & \shortstack{Positive\\seeds} \\
\midrule
Modified & Off & $+0.715 / -4.170 / +0.070$ & $-1.128 \pm 2.654$ & 2/3 \\
Modified & On & $+4.280 / +4.245 / +3.280$ & $+3.935 \pm 0.568$ & 3/3 \\
Original (reproduced) & Off & $-0.067 / -0.022 / -0.272$ & $-0.120 \pm 0.133$ & 0/3 \\
Original (reproduced) & On & $+1.511 / +0.233 / +0.744$ & $+0.830 \pm 0.643$ & 3/3 \\
\bottomrule
\end{tabular*}
\par\smallskip
\begin{minipage}{\linewidth}
\footnotesize
Gains are absolute differences in board accuracy (percentage points), ordered by training seeds 42, 43, and 44. SD is the sample standard deviation across these checkpoints, not a confidence interval.
\end{minipage}
\end{table}

\clearpage
\section{RoFB on TRM for ARC-AGI-1}
\label{app:arc}
We evaluated test-time RoFB on ARC-AGI-1 using the public TRM checkpoint \citep{ARCPrize2025-trm-ckpt}, without retraining. The intervention uses repulsive coupling and a warmup followed by a soft maximum-cell-entropy gate, as in Eq.~\eqref{eq:rofb_gate}; for this experiment, the gate is recomputed once per outer iteration. Of the 400 evaluation puzzles, 100 were used to select $(\lambda,\alpha,t_{\min},\tau)$ by validation pass@1 using Optuna's TPE sampler, and the remaining 300 were used for evaluation. The selected configuration, $(0.0533,0.0407,13,0.0114)$, was reused across augmentation settings, with feedback active during outer iterations 14--16. This historical sweep completed 56 trials and used 20 augmentations for validation; its split and search protocol differ from the main experiments.

Table~\ref{tab:arc_performance} shows small, mixed changes across augmentation settings. At 100 augmentations, pass@1, pass@5, and pass@100 are unchanged. These point estimates from one checkpoint do not establish a consistent accuracy gain.

A separate diagnostic subset of 40 puzzles, balanced between baseline successes and failures, suggests a limitation of prediction-based coupling. With four augmented inputs per puzzle and no feedback, 45 of the 97 terminal predictions that were confident (maximum input-cell entropy at most $10^{-3}$) were incorrect, and 36 of these 45 did not change during the final eight high-level updates. Without feedback, cell errors concentrate where the output should differ from the input ($35.0\%$, versus $1.8\%$ on cells that should be copied, for matching input/output shapes). An erroneous cell's predicted-color group nevertheless contains mostly correct cells (median $72\%$). The coupling therefore does not explicitly distinguish incorrect assignments from correct peers sharing the same confident prediction. This is consistent with limited benefit, but does not establish the cause of the small gains or imply that such errors cannot be corrected.

\begin{table}[!hb]
\centering
\caption{ARC-AGI-1 results on 300 evaluation puzzles with validation-selected RoFB. Scores are percentages.}
\label{tab:arc_performance}
\small
\setlength{\tabcolsep}{4pt}
\begin{tabular*}{\linewidth}{@{\extracolsep{\fill}}r rr rr rr@{}}
\toprule
 & \multicolumn{2}{c}{pass@1} & \multicolumn{2}{c}{pass@5} & \multicolumn{2}{c}{pass@100} \\
\cmidrule(lr){2-3}\cmidrule(lr){4-5}\cmidrule(lr){6-7}
Aug. & Baseline & RoFB & Baseline & RoFB & Baseline & RoFB \\
\midrule
1 & 35.333 & 35.500 & 37.833 & 37.500 & 37.833 & 37.500 \\
3 & 38.833 & 39.000 & 43.500 & 43.167 & 43.500 & 43.167 \\
7 & 39.333 & 39.000 & 46.667 & 46.667 & 47.333 & 47.000 \\
20 & 41.000 & 40.667 & 49.500 & 49.500 & 51.333 & 51.333 \\
100 & 41.667 & 41.667 & 50.167 & 50.167 & 55.000 & 55.000 \\
\bottomrule
\end{tabular*}
\par\smallskip
\begin{minipage}{\linewidth}
\footnotesize
Aug. denotes the augmentation setting. Pass@$K$ tests whether the correct grid appears among the top $K$ distinct predictions ranked by votes, averaging test pairs within each puzzle and then puzzles. Increasing $K$ changes the rank cutoff within a fixed candidate pool, not the inference budget. The baseline uses the same evaluation pipeline with $\lambda=10^{-6}$.
\end{minipage}
\end{table}

\clearpage
\newpage

\section{Detailed Training Setups}
\label{app:hparams}
\begin{table}[ht]
\centering
\caption{AKOrN: full architecture and training hyperparameters.}
\label{tab:hp_akorn}
\small
\begin{tabular*}{\linewidth}{@{\extracolsep{\fill}}lcc@{}}
\toprule
                                  & Sudoku Extreme        & Maze Hard       \\
\midrule
\multicolumn{3}{l}{\textit{Architecture}} \\
oscillator dim $m$                & 4                     & 4                     \\
embed width $D$           & 512                   & 512                   \\
\# blocks $L$                      & 1                     & 1                     \\
\# heads                           & 8                     & 8                     \\
$\gamma$ (step size)              & 1.0                   & 1.0                   \\
coupling $J$                      & self-attn             & self-attn             \\
positional enc.                   & GTA  \citep{Miyato2023-op}     & GTA                   \\
input injection                   & \texttt{add}          & \texttt{add}          \\
\midrule
\multicolumn{3}{l}{\textit{Training}} \\
$\Ttrain$                         & 64                    & 64                    \\
$\Tgrad$ (TBPTT)                  & 8                     & 8                     \\
optimizer                         & AdamW                 & AdamW                 \\
learning rate                     & $1{\times}10^{-3}$    & $3{\times}10^{-4}$    \\
weight decay                      & $1{\times}10^{-2}$    & $1{\times}10^{-4}$ \\
batch size (per GPU)              & 100                   & 32                    \\
epochs                            & 5                     & 200                   \\
clip grad norm                    & 1.0                   & 1.0                   \\
EMA $\beta$                       & 0.995                 & 0.995                 \\
EMA \texttt{update\_every}        & 10                    & 1                     \\
loss                              & token-wise CE         & masked CE on cells    \\
\bottomrule
\end{tabular*}
\\[2pt]
\end{table}

\clearpage

\begin{table}[!h]
\centering
\caption{ItrSA++: full architecture and training hyperparameters. We use the affine-free variant ($\mathtt{rmsnorm\_affine}{=}\mathrm{False}$) as the canonical baseline. Inference horizons follow the convention $T = R \cdot n_{\mathrm{block}}$ with $R{=}\mathrm{num\_rep\_attn}{=}4$.}
\label{tab:hp_itrsa}
\small
\begin{tabular*}{\linewidth}{@{\extracolsep{\fill}}lcc@{}}
\toprule
                                  & Sudoku Extreme        & Maze Hard       \\
\midrule
\multicolumn{3}{l}{\textit{Architecture}} \\
\texttt{embed\_dim}               & 384                   & 384                   \\
\texttt{num\_heads}               & 12                    & 12                    \\
\texttt{num\_layers}              & 1                     & 1                     \\
\texttt{num\_iter} ($=n_{\mathrm{block}}^{\mathrm{train}}$) & 16 & 32                    \\
\texttt{num\_rep\_attn} ($R$)     & 4                     & 4                     \\
\texttt{ffn\_dim\_multiplier}     & 4                     & 4                     \\
\texttt{use\_cross\_attn}         & True                  & True                  \\
\texttt{rmsnorm\_affine}          & False                 & False                 \\
positional enc.                   & GTA & GTA  \\
\texttt{confidence\_type}         & \texttt{log\_prob}    & \texttt{log\_prob}    \\
$L_{\mathrm{sqrt}}$ / vocab / classes & 9 / 10 / 9        & 30 / 4 / 5            \\
\midrule
\multicolumn{3}{l}{\textit{Training}} \\
optimizer                         & AdamW                 & AdamW                 \\
$\beta_1, \beta_2$                & 0.9, 0.95             & 0.9, 0.95             \\
learning rate                     & $5{\times}10^{-4}$    & $5{\times}10^{-4}$    \\
weight decay                      & 0.01                  & 0.01                  \\
batch size (per device)           & 64                    & 64                    \\
\texttt{accumulate\_grad\_batches} & 1                    & 1                     \\
\texttt{num\_nodes}               & 8                     & 8                     \\
global batch size                 & 512                   & 512                   \\
\texttt{gradient\_clip\_val}      & 1.0                   & 1.0                   \\
\texttt{precision}                & bf16-mixed            & bf16-mixed            \\
\texttt{max\_epochs}              & 10                    & 6000                  \\
\texttt{num\_remain\_grad} (TBPTT) & 2                    & 4                     \\
EMA $\beta$                       & 0.995                 & 0.995                 \\
\texttt{update\_after\_step}      & 100                   & 10                    \\
\texttt{update\_every}            & 10                    & 1                     \\
\texttt{use\_compile}             & True                  & True                  \\
\bottomrule
\end{tabular*}
\end{table}

\pagebreak
\newpage

\begin{table}[!h]
\centering
\caption{TRM: full architecture and training hyperparameters.}
\label{tab:hp_trm}
\small
\begin{tabular*}{\linewidth}{@{\extracolsep{\fill}}lcc@{}}
\toprule
                                  & Sudoku-Extreme        & Maze-30x30-Hard       \\
\midrule
\multicolumn{3}{l}{\textit{Architecture}} \\
$H_{\mathrm{cycles}}$             & 3                     & 3                     \\
$L_{\mathrm{cycles}}$             & 6                     & 4                     \\
$H_{\mathrm{layers}}$             & 0 (shared with $L$)   & 0 (shared with $L$)   \\
$L_{\mathrm{layers}}$             & 2                     & 2                     \\
hidden\_size                      & 512                   & 512                   \\
num\_heads                        & 8                     & 8                     \\
expansion (SwiGLU)                & 4                     & 4                     \\
positional enc.                   & RoPE                  & RoPE                  \\
\texttt{halt\_max\_steps}         & 16                    & 16                    \\
\texttt{halt\_exploration\_prob}  & 0.1                   & 0.1                   \\
\texttt{no\_ACT\_continue}        & True                  & True                  \\
\texttt{mlp\_t}                   & False                 & False                 \\
\texttt{puzzle\_emb\_len}         & 16                    & 16                    \\
\texttt{puzzle\_emb\_ndim}        & 512                   & 512                   \\
forward dtype                     & bfloat16              & bfloat16              \\
\midrule
\multicolumn{3}{l}{\textit{Training}} \\
optimizer                         & AdamW                 & AdamW                 \\
$\beta_1, \beta_2$                & 0.9, 0.95             & 0.9, 0.95             \\
learning rate                     & $1{\times}10^{-4}$    & $1{\times}10^{-4}$    \\
\texttt{lr\_warmup\_steps}        & 2000                  & 2000                  \\
\texttt{lr\_min\_ratio}           & 1.0 (constant after warmup) & 1.0 (constant after warmup) \\
weight\_decay                     & 1.0                   & 1.0                   \\
\texttt{puzzle\_emb\_lr}          & $1{\times}10^{-4}$    & $1{\times}10^{-3\,\S}$ \\
\texttt{puzzle\_emb\_weight\_decay} & 1.0                 & 1.0                   \\
global batch size                 & 768 (8 GPU)           & 768 (4 GPU)           \\
epochs                            & $5{\times}10^{4}$     & $5{\times}10^{4}$     \\
\texttt{eval\_interval}           & 5000                  & 5000                  \\
loss                              & stablemax CE + ACT BCE & stablemax CE + ACT BCE \\
EMA                               & on, rate $0.999$      & on, rate $0.999$      \\
seed                              & 0                     & 0                     \\
\bottomrule
\end{tabular*}
\\[2pt]
{\footnotesize $^{\S}$The Maze launcher \texttt{scripts/run\_miyabi.sh} contains a double assignment \texttt{puzzle\_emb\_lr=1e-4 puzzle\_emb\_lr=1e-3}; the Hydra later-wins rule yields $1{\times}10^{-3}$.\par}
\end{table}

\end{document}